%% file: main.tex
\documentclass[11pt, a4paper, logo, copyright]{googledeepmind}

\usepackage{dm-colors}
\usepackage{amsmath}
\usepackage{pstricks, pst-node}
\usepackage{verbatim}
\usepackage{multirow}
\usepackage{scalerel}
\usepackage{booktabs}
\usepackage{enumitem}
\usepackage{xspace}
\usepackage{bm}
\usepackage{bbm}
\usepackage{mathtools}
\usepackage{soul}
\usepackage{epsfig}
\usepackage{graphicx}
\usepackage{tcolorbox}
\usepackage{subcaption}
\usepackage{tcolorbox}
\usepackage{amssymb}
\usepackage{colortbl}
\usepackage{csquotes}
\usepackage{etex}
\usepackage{setspace}
\usepackage{colortbl}
\usepackage{tabularx,ragged2e}
\usepackage{svg}
\usepackage{placeins}
\usepackage{float}
\usepackage[symbol]{footmisc}
\usepackage[bibstyle=nature,citestyle=numeric-comp,%
            natbib=true,backend=biber,maxbibnames=99,%
            giveninits=false,sorting=none]{biblatex}
\usepackage{nameref}
\usepackage{varioref}
\hypersetup{
    pagebackref=false,
    breaklinks=false, 
    colorlinks=true,
    bookmarks=true,   
    citecolor=ourdarkblue,
    urlcolor=ourdarkblue,
    linkcolor=ourdarkblue
}
\usepackage[noabbrev,capitalize]{cleveref}

\newtcolorbox{promptbox}[1]{ 
    enhanced, 
    text width=0.97\textwidth, %
    colback=gray!10, 
    colframe=gray!75!black, 
    coltitle=white, 
    colbacktitle=gray!50!black, 
    fonttitle=\bfseries, 
    title=#1, 
    boxrule=0.5mm, 
    arc=2mm, 
    attach boxed title to top left={yshift=-0.1in, xshift=0.15in}, 
    boxed title style={sharp corners, boxrule=0.5mm} 
}

\bibliography{main}

\title{Advancing Conversational Diagnostic AI with Multimodal Reasoning}
\correspondingauthor{$\{$ksaab$,\, $janfreyberg$,\, $pcj$,\,$alankarthi$,\,$rtanno$ \}$@google.com}

\reportnumber{} 

\author[$\ast$,1]{Khaled Saab}
\author[$\ast$,2]{Jan Freyberg}
\author[$\ast$,1]{Chunjong Park}
\author[1]{Tim Strother}
\author[1]{Yong Cheng}
\author[1]{Wei-Hung Weng}
\author[1]{\\David G.T. Barrett}
\author[1]{David Stutz}
\author[1]{Nenad Tomasev}
\author[2]{Anil Palepu}
\author[1]{Valentin Liévin}
\author[2]{Yash Sharma}
\author[2]{\\Roma Ruparel}
\author[2]{Abdullah Ahmed}
\author[1]{Elahe Vedadi}
\author[2]{Kimberly Kanada}
\author[2]{Cian Hughes}
\author[2]{Yun Liu}
\author[1]{Geoff Brown}
\author[1]{Yang Gao}
\author[1]{Sean Li}
\author[1]{S. Sara Mahdavi}
\author[2]{James Manyika}
\author[2]{Katherine Chou} 
\author[2]{Yossi Matias} 
\author[1]{Avinatan Hassidim}
\author[2]{\\Dale R. Webster}
\author[1]{Pushmeet Kohli}
\author[1]{S.M. Ali Eslami}
\author[1]{Joëlle Barral}
\author[2]{Adam Rodman}
\author[2]{Vivek Natarajan}
\author[2]{\\Mike Schaekermann}
\author[1]{Tao Tu}
\author[$\dagger$,2]{Alan Karthikesalingam}
\author[$\dagger$,1]{Ryutaro Tanno}
\affil[*]{Equal contributions}
\affil[$\dagger$]{Equal leadership}

\affil[1]{Google DeepMind}
\affil[2]{Google Research}

\begin{abstract}
AI systems based on Large Language Models (LLMs) have demonstrated great potential for conducting diagnostic conversations but evaluation has been largely limited to language-only interactions, deviating from the real-world requirements of remote care delivery. Instant messaging platforms can permit clinicians and patients to upload and discuss multimodal medical artifacts seamlessly in conversation, but the ability of LLMs to reason over such data while preserving other attributes of competent diagnostic conversation remains unknown. Here we advance the conversational diagnosis and management performance of the Articulate Medical Intelligence Explorer (AMIE) through a new capability to gather and interpret multimodal medical data, and reason about this precisely during consultations. 
Leveraging Gemini 2.0 Flash, our system implements a state-aware dialogue framework, where conversation flow is dynamically controlled by intermediate model outputs reflecting patient states and evolving diagnoses. Follow-up questions are strategically directed by uncertainty in such patient states, leading to a more structured multimodal history-taking process which emulates that of experienced clinicians. 
We compared AMIE to primary care physicians (PCPs) in a randomized, double-blind study of chat based consultations with 25 patient actors in the style of an Objective Structured Clinical Examination (OSCE). We constructed 105 evaluation scenarios, using common artifacts such as smartphone photos of skin conditions, ECG tracings, and PDFs of clinical documents across a diversity of conditions and demographics. We designed a rubric to assess multiple aspects of the multimodal capability of AMIE and PCPs, in addition to other clinically meaningful axes such as history-taking, diagnostic accuracy, management reasoning, communication skills, and empathy. Evaluation by 18 specialists revealed superior performance of AMIE relative to PCPs in handling and reasoning about multimodal data on 7 out of 9 axes while displaying similar superior performance in the non-multimodal metrics, including diagnostic accuracy, on 29 out of 32 axes. These results demonstrate clear progress in multimodal conversational diagnostic AI, although real-world translation necessitates further research.
\end{abstract}

\begin{document}

\maketitle
\input{introduction}
\input{methods}
\input{study}
\input{results}
\input{related}
\input{discussion}
\input{conclusion}

\section{Acknowledgements}
This project was an extensive collaboration between many teams at Google Research and Google DeepMind. We thank David Racz, Shakir Mohamed, Christian Wright, Rachelle Sico, Joelle Wilson, Brian Gabriel and Jenn Sturgeon for their comprehensive review and detailed feedback on the manuscript. We also thank Josh Grondie, Kenya Moore and John Guilyard for their contributions to the animations and visuals. We would like to thank SiWai Man, Brett Hatfield and Gordon Turner for supporting the OSCE study, and GoodLabs Studio Inc, Intel Medical Inc and Chris Smith for their partnership in conducting the OSCE study in North America, and JSS Academy of Higher Education and Research and Vikram Patil for their partnership in conducting the OSCE study in India. Finally, we are grateful to Ewa Dominowska, Kavita Kulkarni, Shruti Garg, Renee Wong, Amy Wang, Roma Ruparel, Ellery Wulczyn for their support during the course of this project. Lastly, we would like to thank Raia Hadsell, Zoubin Ghahramani, Oriol Vinyals, Koray Kavukcuoglu for their support of this work.

\section{Data Availability}
Some of the datasets used in the development of multimodal AMIE are open-source (SCIN, PAD-UFES-20, PTB-XL and ECG-QA).

\section{Code Availability}
Our system utilizes Gemini 2.0 Flash as its base foundation model. Base Gemini models, including Gemini 2.0 Flash, are generally available via Google Cloud APIs. The core techniques, particularly the state-aware dialogue phase transition framework described in Section \ref{sec:reasoning_method} provide details for implementing our approach. However, the specific implementation relies on internal Google infrastructure and tooling. Due to this, and more importantly, the safety implications associated with the unmonitored deployment of AI systems in medical contexts, we are not open-sourcing the codebase and the specific prompts employed in our work at this time. In the interest of responsible innovation, we will be working with research partners, regulators, and healthcare providers to further validate and explore safe onward uses of our medical models.

\section{Competing Interests}
This study was funded by Alphabet Inc and/or a subsidiary thereof (`Alphabet'). All authors are (or were) employees of Alphabet and may own stock as part of the standard compensation package.

\setlength\bibitemsep{3pt}
\printbibliography
\clearpage
\appendix
\input{appendix}

\end{document}

%% file: introduction.tex
\section{Introduction}


Healthcare delivery faces significant challenges globally from ageing populations and increasing care fragmentation through to clinician burnout and discontinuity between clinical and financial incentives~\cite{Lawson2023-jd}. This drives increased wait times for providers, delayed care, and potentially worsened morbidity and mortality~\cite{Russo2023-vq}. Even in high-income countries, there are many regions where access to primary care is limited and challenging ~\cite{Lee2024-gg}. 

Recently, we have witnessed the remarkable promise of AI systems in healthcare, especially those based on large language models (LLMs), as potential tools to augment healthcare delivery and address some of these challenges. For instance, the Articulate Medical Intelligence Explorer (AMIE) -- an LLM-based conversational diagnostic AI system -- demonstrated physician-like capabilities to steer text-based clinical conversations, attaining non-inferior or superior performance compared to primary care physicians (PCPs) across a broad array of measurements for consultation quality in studies involving patient actors ~\cite{tu2025towards,Palepu2025-de}. Real-world deployments of conversational AI systems for triage and care navigation have also shown promising performance in certain domains~\cite{Zeltzer2023-iu, Zeltzer2025-ks} including patient acceptability~\cite{Lizee2024-uy}.

Despite the early promise of this technology, LLM-based medical AI systems have predominantly been studied and implemented as text-only chatbots. This represents a significant deviation from how multimodal information is routinely used in care. As demonstrated during the COVID-19 pandemic~\cite{Shaver2022-yb}, to augment in-person care through visits at lower costs and increased patient convenience~\cite{Liu2024-ud}, it is integral for remote care participants to be able to converse about, and interpret multimodal medical information. This can take the form of images like patient-provided skin pictures, ECG tracings, or lab reports, which can be shared via ubiquitous messaging platforms~\cite{wong2018real, giordano2017whatsapp} that combine text and multimodal chat; or even more commonly can occur via real-time video consultations.

Multimodal medical tests are essential in effective care and can significantly guide the course of a consultation. However, evidence validating the use of LLMs for diagnostic conversations involving such multimodal data is scarce, revealing an important discrepancy between clinical needs and current technology. Patients may struggle to precisely articulate the results of an investigation using text alone, potentially omitting crucial details such as precise laboratory test values, while essential clinical information commonly resides only in non-textual formats \cite{wong2018real}. For example, visual data like photographs are paramount for remote assessment of dermatological conditions, while clinical documents such as blood reports or specialist letters contain objective findings that can be otherwise impractical to relay accurately via transcription to text-chat. A text-only approach prevents AI from leveraging these rich, supplementary objective sources of information, hindering its ability to form a complete clinical picture.

An over-reliance on text-only input does not only increase the risk of diagnostic errors but also risks exacerbating disparities in access to telehealth~\cite{Campanozzi2023-xl}, presenting barriers for individuals with lower digital literacy and language proficiency. In contrast, instant messaging apps that allow users to send text, voice, and video messages (or share photo, audio and video messages) are commonplace with billions of users. Although real-time video-call is more common in telemedicine, such multimedia instant messaging platforms enable considerably richer communication than text-only conversation and have seen reported use as tools for remote clinical consultations in which doctors and patients can exchange multimodal medical data (lab results, videos, photos, and more) \cite{giordano2017whatsapp,john2022smartphone}

\begin{figure}[htp]
    \centering
    \includegraphics[width=1\textwidth,height=1\textheight,keepaspectratio]{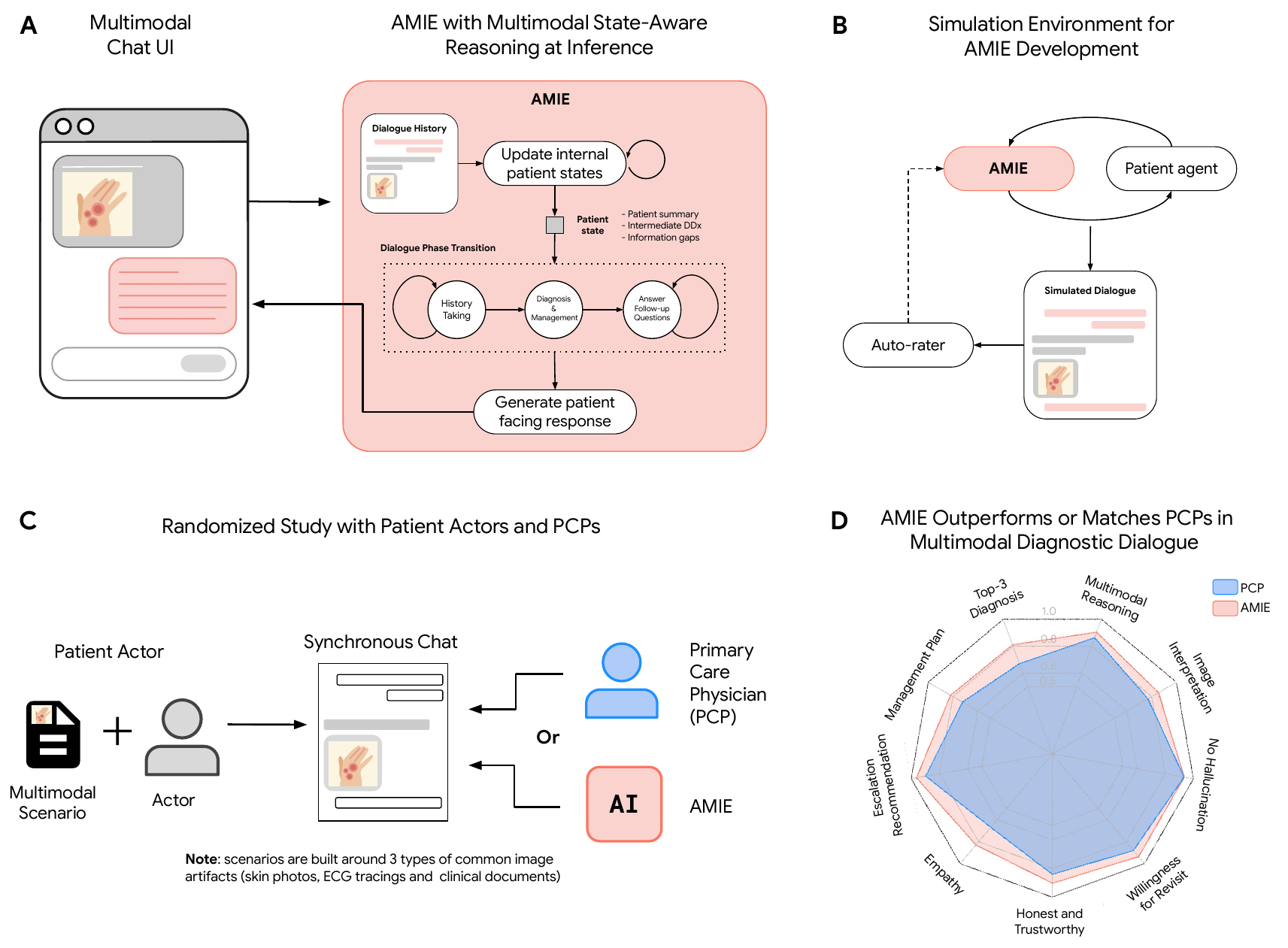}
    \caption{\footnotesize \textbf{Overview of key contributions.} This figure provides a schematic overview of the key components enabling and evaluating multimodal diagnostic conversations within the Articulate Medical Intelligence Explorer (AMIE) system, facilitated through a multimodal chat interface. The main aspects include \textbf{A. Multimodal state-aware reasoning:} AMIE employs a novel state-aware dialogue phase transition framework built on the publicly available Gemini 2.0 Flash model. This dynamically controls the conversation flow through history taking, diagnosis, and management phases, guided by intermediate outputs reflecting patient state and diagnostic uncertainty, allowing strategic requests and interpretation of multimodal artifacts. \textbf{B. Simulation environment:} A comprehensive simulation framework enables rapid development and automated evaluation. It involves generating realistic patient scenarios grounded in real images and metadata, using Gemini 2.0 Flash, simulating turn-by-turn multimodal dialogues between AMIE and patient agents, and utilizing an auto-rater agent for assessment against clinical criteria. \textbf{C. Randomized comparative study:} AMIE's performance was rigorously evaluated against primary care physicians (PCPs) in a randomized, double-blind, OSCE-style study. Trained patient actors conducted synchronous chat consultations based on 105 diverse multimodal scenarios involving artifacts such as skin photos, ECGs, and clinical documents. \textbf{D. Evaluation results:} Specialist evaluations demonstrated that AMIE attains comparable or superior performance to PCPs in handling and reasoning about multimodal data over multiple evaluation axes alongside strong performance in diagnostic accuracy and overall consultation quality.}
    \label{fig:overview_figure}
\end{figure}

The ability of LLMs to request, interpret and reason about multimodal medical tests during clinical conversations has not been investigated or evaluated. To address this, we introduce multimodal AMIE, advancing the conversational diagnostic capabilities of AMIE~\cite{tu2025towards} by integrating multimodal medical perception. Leveraging Gemini 2.0 Flash, our system implements a novel state-aware framework to dynamically gather, interpret, and reason about multimodal medical data within the flow of a clinical conversation, aiming to emulate the diagnostic process of experienced clinicians in a telehealth setting. Through a randomized, blinded human evaluation that emulates an Objective Structured Clinical Examination (OSCE)  -- a standardised assessment method used in both medical schools and physician licensing exams worldwide -- we compared AMIE to PCPs in multimodal text-based chats and evaluated their behaviours on a number of carefully designed multimodal patient scenarios. Figure \ref{fig:overview_figure} provides a schematic overview of our system's key components, our evaluation methodology and findings.

\clearpage
Our contributions include:
\begin{itemize}
    \item\textbf{Multimodal state-aware reasoning at inference}: We introduce a novel state-aware dialogue phase transition framework that orchestrates the conversation flow (Figure \ref{fig:cor}). Leveraging Gemini 2.0 Flash, this framework dynamically adapts AMIE's responses based on intermediate model outputs reflecting the evolving patient state, diagnostic hypotheses and uncertainty. This enables AMIE to strategically request relevant multimodal artifacts (in our case skin images, ECGs, and clinical documents) when needed, interpret the findings accurately, integrate this information seamlessly into the ongoing dialogue, and use it to refine diagnoses and guide further questioning, emulating the structured, adaptive reasoning process of experienced clinicians during telemedicine consultations.
    \item \textbf{Simulation environment for dialogue evaluation}: To enable rapid iteration and robust automated assessment, we developed a comprehensive simulation framework (Figure \ref{fig:simulation}). This involves: (1) Generating realistic patient scenarios, including detailed profiles and multimodal artifacts derived from existing medical image datasets, augmented with plausible clinical context using Gemini with web-search. (2) Simulating turn-by-turn multimodal dialogues between an AMIE agent (instructed to be empathetic and clinically accurate) and a patient agent (adhering to the scenario). (3) Employing an auto-rater agent to evaluate these simulated dialogues against predefined clinical criteria, including diagnostic accuracy, information gathering effectiveness, management plan appropriateness and safety (e.g., hallucination detection). 
    \item \textbf{Dedicated multimodal evaluation OSCE rubric}: We developed and applied a specific Multimodal Understanding \& Handling (MUH) rubric within our OSCE framework (detailed in Table \ref{tab:mm_osce_rubric}). This rubric was designed to rigorously assess and compare the competence of both AMIE and PCPs in handling and interpreting multimodal artifacts (e.g., understanding image quality, engaging with the artifact, requesting additional relevant artifacts, interpreting findings and communicating these findings) in the context of clinical consultations.
    \item \textbf{Effective expert evaluation through multimodal OSCE}: We designed and conducted a remote OSCE study with 105 case scenarios where validated patient actors engaged in synchronous conversations with AMIE or PCPs in a blinded fashion. The sessions were performed through a chat interface where patient actors could upload multimodal artifacts (e.g., skin photos) if requested by the doctor or AMIE. 
    \item \textbf{Comparing AMIE and PCP performance}: Our evaluation harness showed that the ability of AMIE to handle and interpret multimodal data in clinical conversations was superior or non-inferior to PCPs whilst scoring higher in other key indicators of consultation quality such as diagnostic accuracy, management reasoning and empathy.
\end{itemize}

%% file: methods.tex
\section{Approach}

This work advances the diagnostic capabilities of the AMIE system~\cite{tu2025towards,Palepu2025-de} by integrating multimodal perception, enabling it to conduct diagnostic conversations that are more clinically realistic and that incorporate various forms of medical data beyond text.

Our approach centers on a state-aware dialogue phase transition framework, detailed in Section \ref{sec:reasoning_method}, which leverages the multimodal reasoning capabilities of Gemini 2.0 Flash \cite{googleGemini20FlashGenerative}. This framework guides AMIE through structured phases of history taking, diagnosis \& management and follow-up, dynamically adapting the conversation based on intermediate model outputs that reflect the evolving patient state and diagnostic hypotheses. To rigorously evaluate the system's performance during model development and in comparison to human clinicians, we employed a two-pronged evaluation strategy. Firstly, we developed an automated evaluation pipeline (Section \ref{sec:methods_automated_evals}) involving perception tests on isolated medical artifacts and simulated dialogues assessed by an auto-rater across key clinical dimensions like diagnostic accuracy and information gathering. Secondly, as described further in Section \ref{sec:osce_study}, we conducted an expert evaluation using an OSCE-style methodology to assess AMIE's capabilities in realistic, simulated patient encounters involving multimodal data.

\subsection{Multimodal State-Aware Reasoning} \label{sec:reasoning_method}

\begin{figure} 
    \centering
    \includegraphics[width=1\textwidth,keepaspectratio]{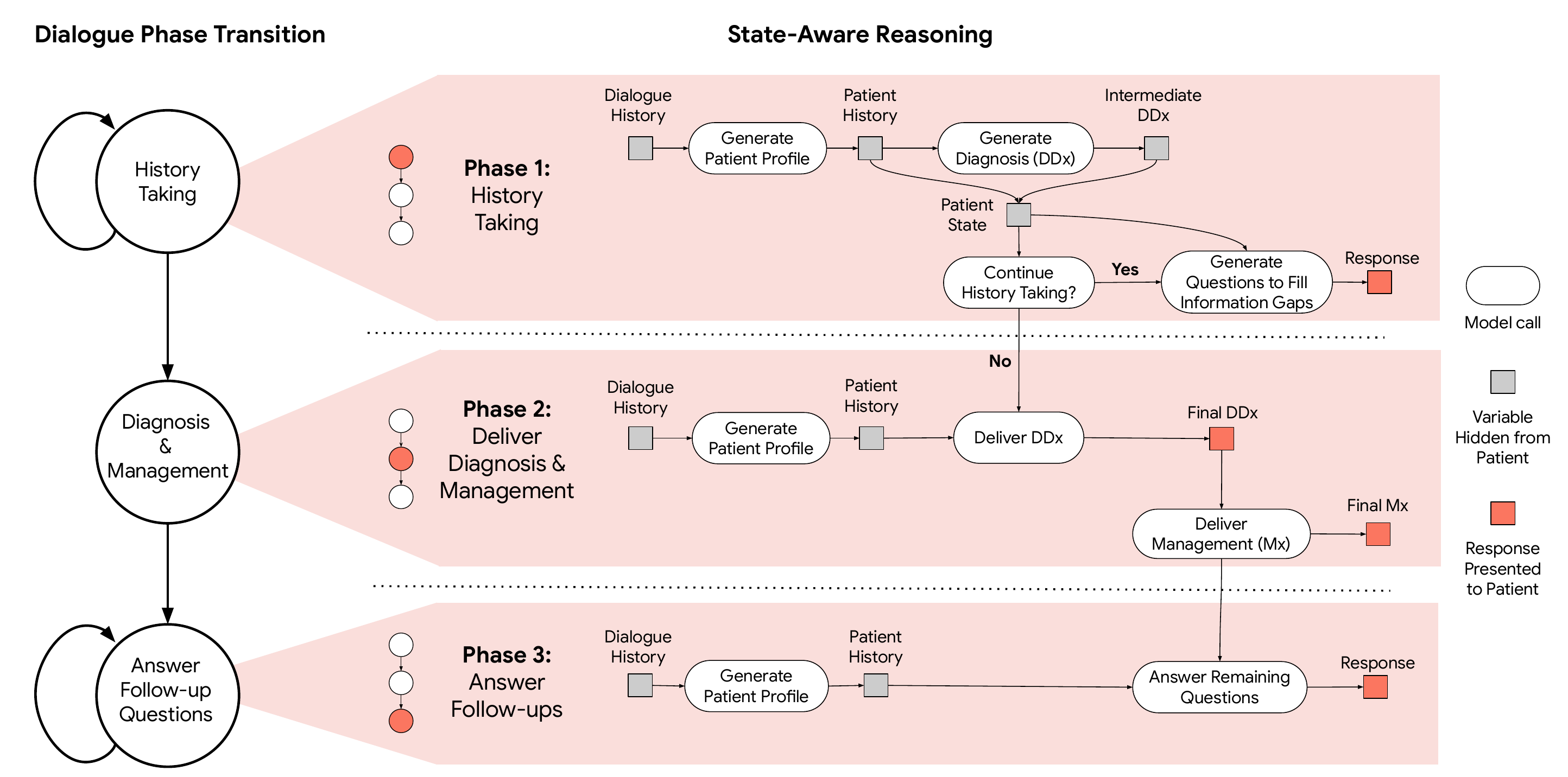}
    \caption{\footnotesize \textbf{Multimodal state-aware reasoning at inference.} AMIE's state-aware dialogue phase transition framework, which structures the diagnostic conversation is illustrated here. The system progresses through three distinct phases, each with a specific goal: (1) History Taking (gathering comprehensive patient information), (2) (Differential) Diagnosis \& Management (formulating and presenting DDx and Mx Plan), and (3) Answer Follow-up Questions (addressing remaining concerns). Within each phase, AMIE maintains an internal state – its dynamic understanding of the patient's situation, evolving diagnoses (DDx), and knowledge gaps – derived from the dialogue history and any multimodal inputs. This state guides specific actions, such as asking targeted questions, requesting multimodal data (if needed), generating internal summaries/DDx, or providing explanations. Transitions between phases are triggered automatically when the system assesses that the objectives of the current phase (e.g., sufficient information gathered, DDx presented) have been met, based on its internal state evaluation. This mechanism enforces a structured yet flexible dialogue flow, inspired by the methodical approach of experienced clinicians.}
    \label{fig:cor}
\end{figure}

Real clinical diagnostic dialogues follow a structured yet flexible path. Clinicians methodically gather information, form potential diagnoses, strategically request and interpret further details (including multimodal data like skin photos or ECGs), continually update their assessment based on new evidence, and eventually formulate a management plan. This process requires a clinician to adapt the line of questioning based on evolving hypotheses and uncertainty, while ensuring that all critical information is considered~\citep{gruppen2017clinical}.

Given the rapid advancements in LLM capabilities and their increasing proficiency in following complex instructions, one might achieve considerable progress towards emulating this process using a sophisticated system prompt alone. However, we hypothesize that for a safety-critical and highly dynamic task like multimodal diagnosis, building an explicit state-aware reasoning system layered on top of the LLM offers critical advantages. Such a system provides greater control over the dialogue flow, enables more reliable tracking of the diagnostic state and uncertainty, facilitates more deliberate integration of multimodal inputs, and ultimately leads to higher-quality, more dependable clinical reasoning rather than relying solely on complex prompting (validated in Section \ref{sec:results_automated_evals}).

Therefore, the multimodal AMIE system implements this state-aware dialogue phase transition framework to manage the diagnostic process. 
This framework dynamically controls AMIE's progression through three distinct phases: (1) History Taking, (2) Diagnosis \& Management, and (3) Follow-up. 
Transitions between phases, and actions within each phase, are driven by intermediate model outputs representing the evolving patient state and diagnostic hypotheses (Figure \ref{fig:cor}). 
Crucially, each phase builds upon the accumulated dialogue history, which contextually incorporates various forms of patient data, including text, images (such as skin photos or ECG tracings), and clinical documents (like lab reports or prior consultation notes).
This state-aware approach allows AMIE to emulate the structured yet adaptive reasoning process of clinicians.

\paragraph{Phase 1: History Taking - Building a Comprehensive Picture}

\begin{enumerate}
    \item \textbf{Patient profile initialization:} A structured patient profile is initialized and dynamically updated throughout the interaction. This profile acts as a condensed, evolving record of known patient information, encompassing the chief complaint, history of present illness, demographics (age, sex, race), positive and negative symptoms, past medical, family, and social/travel histories, medications, other relevant details and a prioritized list of knowledge gaps. Initially, this profile may contain minimal information.
    
    \item \textbf{Evolving differential diagnosis (DDx) generation:} An internal, evolving differential diagnosis (DDx) is generated. This DDx is \textit{not} initially presented to the patient. DDx generation begins after an initial interaction period, allowing baseline information collection. The frequency of DDx updates is configurable.

    \item \textbf{Continuation decision:} A key decision point: whether to continue gathering history or transition to presenting a diagnosis. This decision uses a set of criteria, including assessing if sufficient information exists to formulate a reasonable differential diagnosis. A decision module, querying Gemini 2.0 Flash, determines if current information is sufficient to proceed or if more targeted questions are needed. This module considers the dialogue history and preliminary DDx.
    
    \item \textbf{Targeted question \& multimodal data request generation:} If history taking continues, the system generates focused questions to address information gaps identified in the patient profile and DDx uncertainty. Crucially, the system is designed to recognize when multimodal data is necessary and strategically requests it. For instance, based on reported symptoms like a rash, AMIE will prompt the user to upload skin photos. Its reasoning extends to requesting additional views if needed (e.g., "Could you please provide a photo from a different angle or in better lighting?" or "Do you have photos showing how the rash looked previously?"). Similarly, for reported cardiac symptoms, it might request an ECG tracing if available. Upon receiving an artifact, AMIE elicits detailed descriptions adhering to instructions for interpreting the artifacts and their key aspects for determining salient findings (e.g., for skin: lesion morphology, distribution, color; for ECGs: heart rate, rhythm, key waveforms and intervals). This explicit prompting for descriptive details ensures that the model extracts salient features from the artifact to inform the ongoing conversation and diagnostic reasoning. The generated question or request is presented to the patient.

   \item \textbf{Iterative refinement:} The process is iterative. New information from patient responses and data uploads are incorporated into the dialogue history, updating the patient profile and internal DDx. The patient summary is periodically refreshed to reflect the current understanding.

\end{enumerate}
Once the continuation decision indicates readiness, the system transitions to Phase 2.

\paragraph{Phase 2: Diagnosis \& Management - From Data to Actionable Plan}

\begin{enumerate}
    \item \textbf{Patient profile (optional update):} The patient profile may be further refined.

    \item \textbf{Differential diagnosis validation (sub-phase):} The system enters a DDx validation phase. Focused questions are generated to gather specific evidence that supports or refutes potential diagnoses within the internal DDx. A decision module determines when the DDx is sufficiently validated to present to the patient.
    
    \item \textbf{Differential diagnosis presentation:} The system presents a ranked DDx (5-10 conditions). Crucially, the explanation for each diagnosis is grounded in evidence from the entire interaction, explicitly referencing and explaining findings from the provided multimodal data (e.g., "Based on the photo you sent, the circular shape and central clearing of the rash are characteristic of...," or "The ECG shows specific changes in the ST segment which support the possibility of...").
    
    \item \textbf{Management plan formulation:} After presenting the DDx, the system formulates a management plan based on dialogue history, patient profile, and the presented DDx. The plan includes recommendations for investigations, testing, and/or treatment.
    
    \item \textbf {Management plan delivery:} The system delivers the management plan, potentially iteratively across several turns, allowing for clarification and addressing patient concerns.
\end{enumerate}

The presentation of the refined DDx and management plan signals the transition to Phase 3.

\paragraph{Phase 3: Answer Follow-up Questions - Ensuring Patient Understanding}

\begin{enumerate}
    \item \textbf{Patient profile (optional update):} The patient profile may continue to be updated with information from follow-up questions.
    
     \item \textbf{Plan communication and question answering:} The system addresses remaining patient questions, ensuring the patient understands the proposed management plan, potentially referencing the multimodal artifacts again to clarify points. Responses are guided by the dialogue history, patient profile, presented DDx, and management plan.

    \item \textbf{Dialogue termination:} Dialogue continues until patient questions are addressed, the management plan is communicated, and a natural conclusion is reached.

\end{enumerate}

Upon reaching this termination state, the system proceeds to synthesize the interaction into a structured summary.

\paragraph{After dialogue conclusion: structured post-questionnaire generation}

Once the dialogue reaches a natural conclusion (Phase 3 completion), the system automatically generates a structured post-questionnaire. This process leverages the full multimodal dialogue history and the final internal state (patient profile, final DDx, and management plan) to produce a comprehensive summary suitable for clinical review. Specifically, AMIE is prompted to:

\begin{enumerate}
    \item \textbf{Finalize differential diagnosis (DDx):} Based on the entire interaction, including all textual exchanges and interpretations of multimodal artifacts, generate a final, ranked differential diagnosis listing the most probable condition and several plausible alternatives.
    \item \textbf{Formulate management plan (Mx):} Leveraging both the established DDx and \textbf{web search} for grounding in current medical knowledge and guidelines, detail the recommended management plan. This includes proposed in-visit and ordered investigations, specific actions or recommendations for the patient, necessary escalation level (e.g., video call, in-person visit) with justification, and follow-up requirements (necessity, timeframe, reason).
    \item \textbf{Extract salient artifact findings:} Identify and list the key clinical findings observed in any provided images (skin photos, ECGs, documents) that were relevant to the diagnostic and management reasoning.
\end{enumerate}

The structured post-questionnaire serves as a standardized record of AMIE's clinical assessment, reasoning, and recommendations based on the completed multimodal consultation, and is a crucial output for evaluation and potential clinical handoff.

\subsection{Automatic Evaluations} \label{sec:methods_automated_evals}

We established an automated evaluation framework to support rapid iteration and rigorous assessment of the AMIE system. This included evaluating perception capabilities on isolated medical artifacts and utilizing a simulation environment with auto-raters to assess complete diagnostic multimodal dialogues and perform component ablations.

\subsubsection{Evaluating Base Model Perception on Medical Artifacts}

A critical precursor to effective multimodal diagnostic conversation is ensuring our underlying models possess a fundamental capability: accurate perception of diverse medical artifacts.  While LLMs have demonstrated remarkable progress in understanding and generating text, their ability to reliably "see" and interpret medical images and documents – akin to a clinician's initial visual assessment – remains less explored in the context of conversational differential diagnosis.  Without robust perceptual grounding, even the most sophisticated conversational framework would be limited in its ability to meaningfully integrate multimodal data into history-taking and diagnostic reasoning.

Therefore, we first conducted a suite of perception tests designed to isolate and evaluate the visual understanding of our base models when presented with common medical artifacts: smartphone-captured skin images, ECG tracings, and clinical documents. The objective was not to achieve state-of-the-art performance on complex diagnostic tasks per se, but rather to establish a baseline confidence in whether our models could reliably discern key visual features and clinical information from these modalities in isolation.

For skin images, we utilized the Skin Condition Image Network (SCIN) dataset~\citep{ward2024creating} and PAD-UFES-20~\citep{PACHECO2020106221}, assessing the model's ability to describe lesion morphology, color, and distribution.  For ECGs, we employed PTB-XL~\citep{wagner2020ptb} and ECG-QA benchmark~\citep{oh2023ecgqa}, prompting the model to provide probable diagnoses or answer expert-validated questions based on ECG images generated from raw signals. Finally, for clinical documents, we created the ClinicalDoc-QA dataset that consists of question answering tasks based on a large collection of de-identified clinical notes and patient records generated by physicians, to evaluate the model's comprehension of medical information from clinical documents. Additional dataset details can be found in Appendix \ref{app:perception_tests}.

The results of these perception tests, presented in Section \ref{sec:results_automated_evals} and Appendix \ref{app:perception_test_results}, provide an essential confidence check on our base LLM, Gemini 2.0 Flash, and also enable us to explore critical questions such as the impact of history taking in addition to multimodal perception. 

\subsubsection{Auto-Rating Simulated Diagnostic Conversations} \label{sec:simulation_method}

\begin{figure}[htp]
    \centering
    \includegraphics[width=1\textwidth,height=0.82\textheight,keepaspectratio]{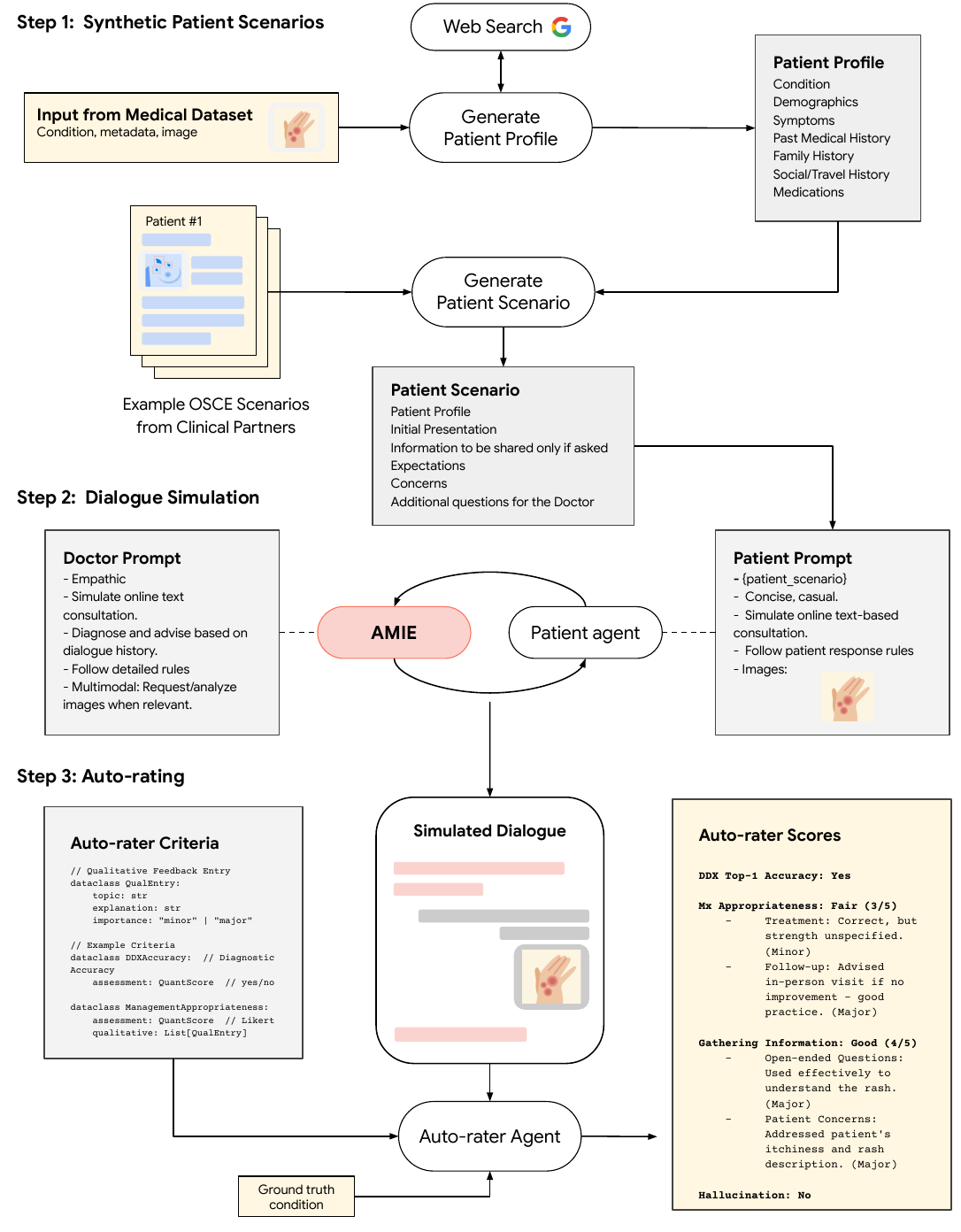}
    \vspace{1mm}
    \caption{\footnotesize\textbf{Overview of multimodal dialogue simulation and evaluation framework.} This figure illustrates the three key components of the system. Step 1 (Patient Scenario Generation): Comprehensive patient profiles are created, including condition, demographics, symptoms, and medical history, using information derived from web searches and real-world datasets e.g., PTB-XL (for cardiology) and SCIN (for dermatology). These profiles are then used to generate detailed patient scenarios, outlining the patient's presentation and expectations. Step 2 (Dialogue Simulation): A doctor agent and a patient agent engage in a text-based consultation with multimodal artifact upload. The doctor agent is instructed to provide empathetic and clinically accurate responses, while the patient agent responds truthfully based on the generated scenario. Step 3 (Auto-rating): An auto-rater agent evaluates the simulated dialogue based on pre-defined criteria, including management appropriateness, information gathering effectiveness, and the presence of hallucinations. Qualitative feedback is also provided by the auto-rater to explain the scores.}
    \label{fig:simulation}
\end{figure}

The following subsections detail our three-step approach illustrated in Figure \ref{fig:simulation} to creating a simulation environment for auto-evaluation: step (1) encompasses patient profile and scenario generation, which is then used in step (2) for the turn-by-turn multimodal dialogue generation process used to mimic real-world telemedicine interactions, and finally the simulated dialogue is used in step (3) by the auto-rater for automated assessment of AMIE's conversational abilities.

\paragraph{Step 1: Patient profile and scenario generation}

Generating realistic synthetic dialogues requires detailed patient profiles and corresponding clinical scenarios. Our methodology involves compiling comprehensive patient metadata, including symptoms, demographics, and medical history, tailored to specific medical domains.

\begin{itemize}
    \item \textbf{Dermatology:} We utilized the SCIN and PAD-UES-20 dataset, which provide rich, pre-existing patient attributes (e.g., age, sex, symptoms, skin tone, medical and social history), requiring no synthetic imputation.
    \item \textbf{Cardiology:} We started with the PTB-XL electrocardiogram dataset. As it only includes age and sex, we employed Gemini 2.0 Flash, utilizing its web search tool, to impute clinically plausible symptoms, social/family/medical history, and cardiovascular risk factors associated with the ECG's indicated condition. This grounded the synthetic profiles in the model's medical knowledge and real-world examples from the Web.
    \item \textbf{Clinical Documents:} For this domain, we collaborated with the same external clinical partners who developed our OSCE scenarios. They created realistic patient profiles, document artifacts, and corresponding metadata specifically for our simulation needs.
\end{itemize}

Once comprehensive metadata was established for all domains, we used Gemini 2.0 Flash to generate detailed clinical scenarios. These scenarios provide context for the patient agent in the simulation, outlining the initial presentation, details to share only upon questioning, patient expectations and concerns, desired outcomes, and potential questions for the doctor. We guided Gemini using few-shot examples, ensuring scenario diversity (e.g., varied ethnicities and occupations) and clinical realism. Crucially, the scenarios deliberately omit the final diagnosis, requiring the AI agent to deduce it through the simulated conversation and analysis of any provided multimodal artifacts.

\paragraph{Step 2: Turn-by-turn multimodal dialogue generation}

We simulate dialogues turn-by-turn between a doctor agent (representing AMIE) and a patient agent, mimicking a telemedical interaction.

The doctor agent (AMIE) is instructed to be empathetic and clinically accurate. It utilizes the state-aware dialogue phase transition framework (Section \ref{sec:reasoning_method}) to navigate history taking, diagnosis \& management, and follow-up phases. A key capability is its multimodal nature: it can strategically request and analyze relevant medical artifacts (like skin images or ECGs) based on patient-reported symptoms during the conversation.

The patient agent simulates a patient strictly following a predefined scenario (Step 1 in Section \ref{sec:simulation_method}), which details their profile and clinical scenario. It is prompted to respond truthfully using concise, casual language appropriate for an online consultation, while adhering to specific rules regarding politeness and pacing the release of information to avoid overwhelming the doctor agent.

In each dialogue turn, the doctor agent generates a question or statement based on the conversation history and its internal state. The patient agent then formulates a response according to its scenario instructions. If the scenario dictates and the doctor agent requests it, the patient agent can upload a relevant medical artifact (e.g., an image). AMIE then analyzes this artifact, integrating the findings into its ongoing diagnostic reasoning and subsequent dialogue turns.

This turn-by-turn exchange continues until the conversation reaches a natural conclusion (e.g., patient questions are resolved) or hits a predefined maximum turn limit. After the exchange is concluded, the doctor agent generates the structured post-questionnaire. The output is a complete simulated dialogue transcript, capturing the dynamic exchange of text and multimodal data. 

\paragraph{Step 3: Criteria and scoring with auto-raters}

Auto-rating (automated evaluation) is crucial for the iterative development and safety assessment of conversational models. While human evaluation is valuable, it is often constrained by cost, time and scalability. Auto-raters provide a mechanism for rapid, scalable and consistent performance assessment across essential characteristics.

We employ an auto-rater based on Gemini 2.0 Flash to evaluate the simulated dialogues. The auto-rater, which is given access to the ground truth condition, assesses various aspects, from quantitative metrics like diagnostic accuracy to qualitative dimensions such as information gathering and safety. The specific criteria and scoring methods used by our auto-rater are detailed below (see \cref{table:auto_rating}):

\begin{itemize}
    \item \textbf{Differential diagnosis accuracy:} Measures if the patient's ground truth condition is present within the top 1, 3, or 10 diagnoses listed by AMIE's generated post-questionnaire. Evaluation accounts for synonyms and uses binary scoring (1 if present in top-N, 0 otherwise), averaged across dialogues.
    \item \textbf{Gathering information:} Assesses the model's effectiveness in eliciting information, including its use of open-ended questions, active listening, addressing patient concerns, and summarization. Rated on a 5-point Likert scale (from 1=very poor to 5=excellent).
    \item \textbf{Management plan appropriateness:} Evaluates the suitability of the proposed management plan (including recommended actions) relative to best medical practices and the patient's specific situation. Rated on a 5-point Likert scale (from 1=very poor to 5=excellent).
    \item \textbf{Hallucination:} Detects instances of fabricated information stated by the model (e.g., mentioning details not provided by the patient, making incorrect assumptions, claiming access to non-existent data or medications). This safety-critical metric excludes assessments of the diagnostic or management reasoning itself and is evaluated with a binary score (1=hallucination present, 0=absent).
\end{itemize}

This simulation framework advances beyond previous text-only versions~\citep{tu2025towards} by generating multimodal patient scenarios grounded in real-world datasets (SCIN, PTB-XL, clinical documents) and simulating realistic patient-clinician interactions involving these multimodal medical data. As detailed in Appendix \ref{app:auto_rater_calibration}, calibration analysis confirms good alignment with human expert judgments, thus enabling automated evaluation of the system's multimodal reasoning capabilities.

%% file: study.tex
\section{Expert Evaluation: Multimodal Virtual OSCE}
\label{sec:osce_study}

To compare AMIE's capabilities in undertaking multimodal diagnostic conversations to those of primary care physicians (PCPs), we extended the remote Objective Structured Clinical Examination (OSCE) study design introduced by \citep{tu2025towards}. The quality of dialogues were assessed using a set of rubrics and metrics reflecting the perspective of patients and specialist physicians. We also introduced a new evaluation rubric specifically to assess the ability to use multimodal medical data effectively in the context of clinical consultations. 

The OSCE is a standardized practical assessment widely used in healthcare education to objectively evaluate clinical skills and competencies by simulating real-world practice \cite{sloan1995objective,newble2004techniques}. Unlike traditional knowledge-based examinations, the OSCE assesses practical skills of real-world clinical encounters, typically involving candidates rotating through a series of timed stations where they encounter a trained patient actor portraying a specific clinical scenario. Test takers perform designated tasks such as taking a medical history, conducting a physical examination, interpreting results, or counseling the patient. Examiners observe these interactions and score the test taker's performance against detailed, predefined checklists that assess crucial skills like history taking, examination technique, clinical reasoning and communication.

In this work, we designed and conducted a virtual analogue of the OSCE adapted for multimodal text chat. In this setting patient actors engaged in blinded, synchronous chat conversations with either the AMIE system or PCPs, conducted through a chat interface that allowed exchange of both text and uploaded images as is now commonplace for mobile chat applications (see Figure~\ref{fig:osce_study_flow} for an overview). Within the virtual consultations, patient actors were instructed to upload images such as skin photos, lab tests, or ECG tracings - emulating how popular text chat platforms have been reportedly used as a means for remote consultation. 

\begin{table}[ht!]
    \centering
    \includegraphics[width=1\textwidth,height=\textheight,keepaspectratio]{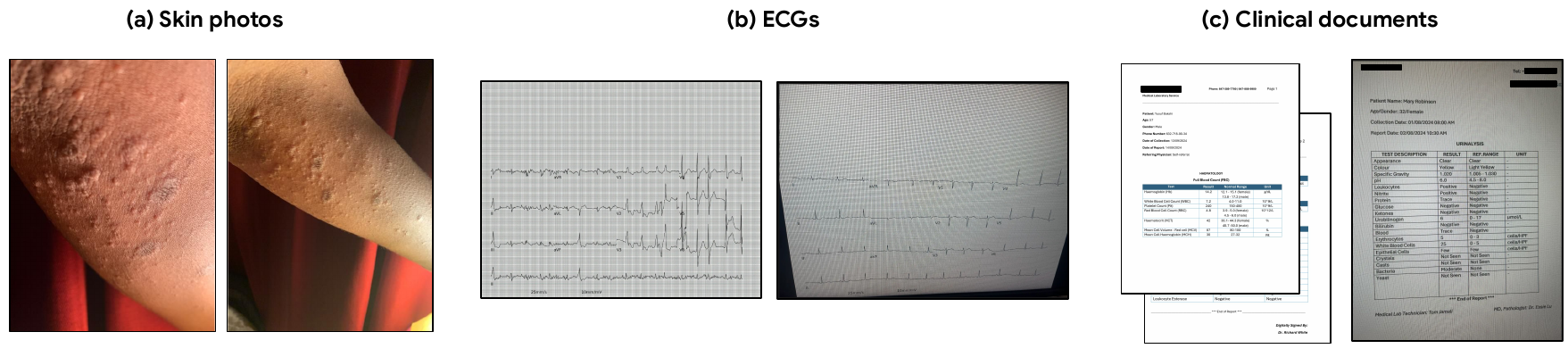}
    \centering
    \resizebox{\textwidth}{!}{
    \begin{tabular}{cccl}
    \toprule
    \textbf{Artefact Type} & \textbf{N} & \multicolumn{1}{c}{\textbf{Source}} & \multicolumn{1}{c}{\textbf{Details}} \\
    \midrule
    \begin{tabular}[c]{@{}c@{}} Phone-captured \\ skin pictures \end{tabular}
    & 35
    & \begin{tabular}[c]{@{}c@{}} SCIN (test split) \cite{ward2024creating} \end{tabular}
    & \begin{tabular}[c]{@{}l@{}} 
    All images in this dataset are captured by mobile phones. \\
    We aimed to attain high coverage of important high-risk and medium-risk dermatological conditions while \\
    ensuring the inclusion of prevalent conditions in practice; we specifically included the 11 conditions \\ 
    delineated in \cite{liu2020deep} including  BCC, Melanoma, SCC/SCCIS, Urticaria, Actinic Keratosis, \\ 
    Allergic Contact Dermatitis, Eczema, Hidradenitis, Psoriasis, Stasis Dermatitis and Tinea, and \\
    selected the remaining 39 conditions based on the prevalence in data, which contained common low-risk conditions \\
    such as skin tag, cyst, folliculitis and acnes. For each condition, we randomly selected Fitzpatrick skin type \\ 
    to achieve demographic diversity, and selected the case with the lowest average annotator confidence in differentials. \\
    Scenarios were then created by Dermatologists in our partner clinical providers based on the combination of images \\
    and other available demographic and clinical metadata. The original skin conditions were used as the ground truth \\
    primary diagnosis of the resultant scenaio packs.
    \\
     \end{tabular} 
    \\
    \midrule
    \begin{tabular}[c]{@{}c@{}} Images of 12-Lead \\ electrocardiography \end{tabular} & 35 & PTB-XL (test split) \cite{wagner2020ptb} & \begin{tabular}[c]{@{}l@{}}
    The raw ECG signals were converted into realistic image format using ECG-Image-Kit \cite{shivashankara2024ecg} to align \\
    with the OSCE context, where physicians often review ECG tracings printed on paper. \\
    We defined relevant categories of ECG findings of varying severity \\
    (Myocardial Infarction, Conduction Abnormalities, Chamber Hypertrophy/Enlargement, \\
    Arrhythmia, ST-T Segment Changes, and Normal Cases). For each category, we selected cases \\
    with lowest confidence in the findings (as estimated by the model in \cite{wagner2020ptb}) \\
    while ensuring coverage of sex and age groups. \\
    Scenarios were then created by Cardiologists based on the combination of the ECG images and demographic data. \\
    The primary diagnosis of the scenarios were defined in such a way that are consistent with the ECG finding. \\
    To test another realistic representation of the artifact, we based a half of the scenarios on the smartphone \\
    photos of the ECG displayed on a computer screen. In all cases, ECGs were provided to PCPs and AMIE as images. \\ 
    \end{tabular} \\
    \midrule
    \begin{tabular}[c]{@{}c@{}} Screenshots of \\clinical documents \end{tabular} 
    &35
    & \begin{tabular}[c]{@{}c@{}} Clinical providers \\in Canada and India \end{tabular} & \begin{tabular}[c]{@{}l@{}} 
    Realistic fictitious clinical documents of diverse types, including blood work, urine test, stool analysis \\
    and prior consultation reports) and their accompanying case scenarios were prepared by internal medicine specialists.  \\ 
    The generated clinical documents were designed to cover high-prevalence conditions in primary care, \\
    and we made sure that they are contain key clinical indicators that should be interpreted appropriately \\
    while they do not explicitly contain the diagnosis or management plans.  \\
    For a half of the scenarios, we used the smartphone photos of the original clinical documents \\
    displayed on a computer screen to simulate another realistic representation of the artifact. \\ 
    \end{tabular} \\
    \bottomrule
    \end{tabular}
    }
    \caption{\footnotesize \textbf{Source materials and breakdown of multimodal scenarios.} The table describes the breakdown of patient scenarios, and details of the steps taken to select cases in the respective categories. Multimodal source materials were either sourced from public datasets such as SCIN \cite{ward2024creating} for skin photos and PTB-XL \cite{wagner2020ptb} for ECG data, or are carefully crafted by our clinical partners for this study (clinical documents). Note that the contents of clinical documents are fictitious, including patient names while they are designed to be realistic. We also simulated real-world perturbations by taking smartphone photos of artifacts displayed on a screen. }
    \label{tab:scenario_details}
\end{table}

\subsection{Scenario Packs}

In collaboration with two organisations that routinely perform OSCE assessments in Canada and India, we developed 105 case scenarios. The scenarios were centered around three types of image artifacts that are commonly available to patients in telemedical primary care: (1) smartphone captured skin images, (2) ECG tracings, and (3) clinical documents. We select these modalities as the most likely to occur in primary tele-care: patients can readily photograph their skin, scan documents they received, or report with ECGs collected via consumer tools \cite{li2022using}.
Table~\ref{tab:scenario_details} provides an overview and examples of image artifacts. Scenarios contained different representations of these image artifacts to represent varying levels of image quality, including photos of a given skin concern from different angles, as well as screenshots and smartphone photos of ECG tracings and clinical documents.

We selected cases for dermatology from the SCIN (Skin Condition Image Network) dataset \cite{ward2024creating}. SCIN contains representative skin images along with rich meta data crowd-sourced from real internet users with skin concerns. ECG tracings were taken from PTB-XL \cite{wagner2020ptb}, the largest publicly available dataset for this modality. Clinical documents were crafted by OSCE laboratories for the purpose of this study. Importantly, we included normal cases, too. Scenarios were designed such that both requesting and interpreting images, as well as taking the patient's history were required---while either alone were insufficient---in order to form a confident diagnosis. To this end, for both skin photo and ECG-based scenarios, we selected challenging images with high annotation ambiguity in their diagnosis labels as detailed in Table~\ref{tab:scenario_details}. Furthermore, dermatologists, cardiologists and internists (medical specialists in internal medicine) ensured that the accompanying text-based component of scenarios (e.g., family/medical history and symptoms) was complementary in the sense that both image and textual information were required to arrive at an accurate diagnosis. Nevertheless, we note that while the scenarios are consistent with the artifacts and the diagnosis, there is no guarantee that they reflect the true case history, as they were created post-hoc. Scenarios were crafted to match case metadata provided in the SCIN and PTB-XL datasets (such as age, sex and ethnicity) when available. For clinical documents, we selected conditions for which diagnosis formation can be guided based on the results of lab reports. 

Lastly, in order to simulate variability in image quality in real-world care settings, for half of the ECG and clinical document cases, we used smartphone photographs of a computer screen showing the artifacts (see Table~\ref{tab:scenario_details}), as in previous work, e.g. \cite{phillips2020chexphoto10000photostransformations}).

\subsection{Study Design}

\begin{figure}[t!]
    \centering
    \includegraphics[width=1\textwidth,height=\textheight,keepaspectratio]{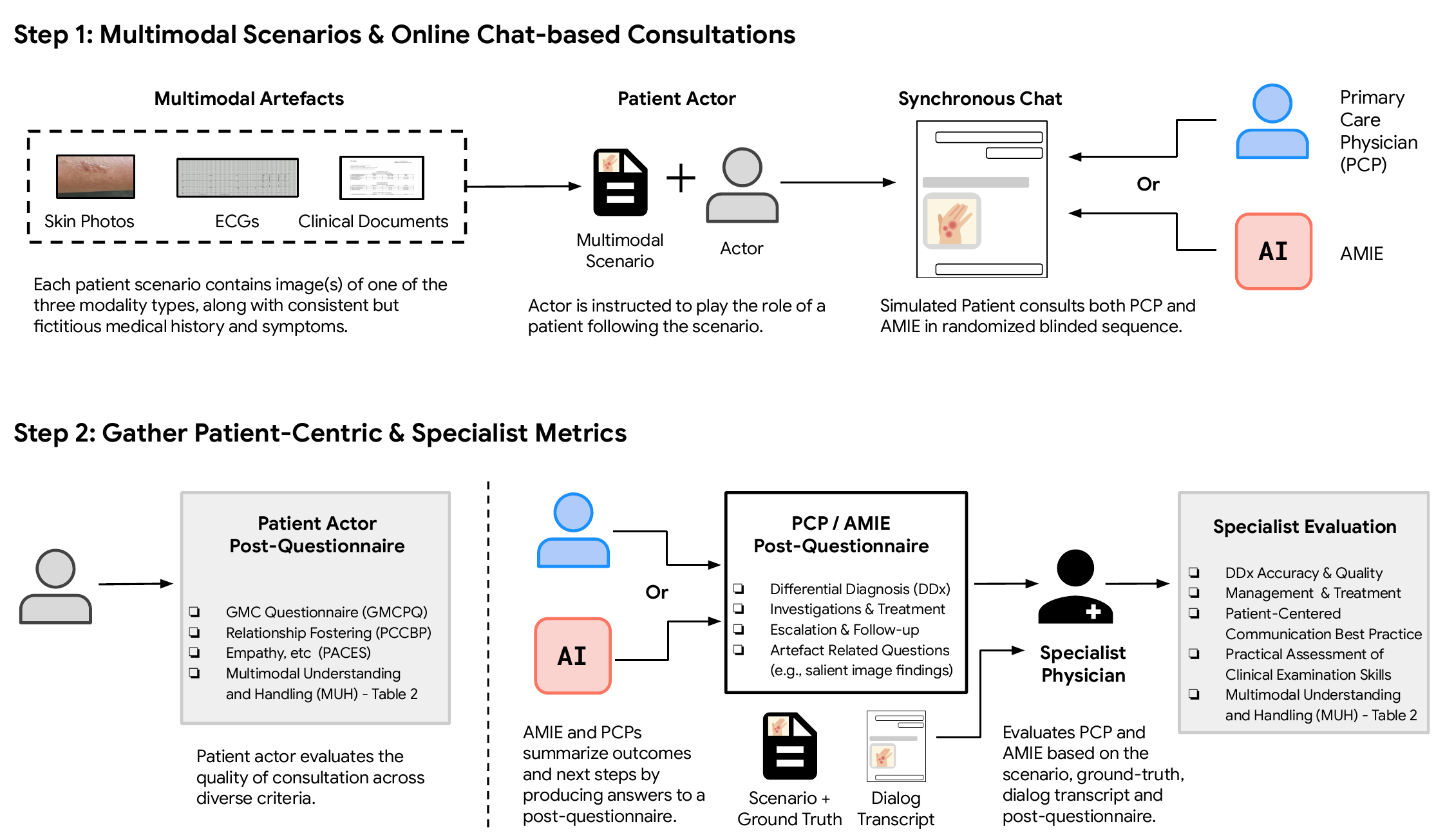}
    \caption{\footnotesize \textbf{Overview of expert evaluation (OSCE)}. Step 1: Patient actors simulate cases based on multimodal scenarios containing history, symptoms, and visual artefacts (skin photos, ECGs, or clinical documents) and conduct synchronous chat consultations with both a PCP and AMIE in a randomized, blinded order. Step 2 (Evaluation):  After consultations, patient actors assess interaction quality by filling out a questionnaire, providing patient-centric quality measures. In parallel, both PCPs and AMIE document their findings by generating answers to a separate post-questionnaire detailing differential diagnosis, management plans (investigation, treatment and escalation needs) and also salient image findings. Subsequently, specialist physicians evaluate the performance of PCPs and AMIE based on the dialogue transcript, post-questionnaire answers, and scenario ground truth across multiple criteria.}
    \label{fig:osce_study_flow}
\end{figure}

Figure~\ref{fig:osce_study_flow} provides an overview of our OSCE study design. For each case scenario, the same patient actor performed one conversation with AMIE and a qualified PCP each via a synchronous chat interface, in a blinded and randomized order. The chat interface supported text-based communication and sharing of images from the scenarios, while displaying the scenario pack information to patient actors throughout the conversation (see Figure~\ref{fig:patient_interface} in Appendix). 

Following each consultation, patient actors completed a questionnaire to rate their experience to represent the patient actor perspective. Separately, a different questionnaire was completed by both AMIE (using offline generation) and PCPs to summarise key clinical findings and next steps. In particular, the questionnaire asked for a differential diagnosis list (at least 3, up to 10 plausible items, ordered by likelihood), a management plan, and a description of salient image findings. Lastly, a group of specialist physicians assessed the performance of AMIE and PCPs in a blinded fashion, based on their consultation transcripts and responses to the post-questionnaires using various rubrics representing the specialist physician perspective.

A large part of the collected patient-centric as well as the specialist metrics are derived from the evaluation rubric introduced in the previous work \cite{tu2025towards} for their direct applicability in the assessment of multimodal diagnostic dialogues. These criteria are designed to primarily to evaluate the consultation quality, the appropriateness of diagnostic and management decisions, the accuracy of clinical reasoning and various aspects of effective communication skills (such as the ability to elicit information and manage patient concerns). The questions in this rubric were derived from consideration of authoritative assessment schemes of clinical interactions such as the Practical Assessment of Clinical Examination Skills (PACES) used by the Royal College of Physicians in the UK for examining history-taking skills \cite{dacre2003mrcp}, the General Medical Council Patient Questionnaire (GMCPQ)\footnote{\scriptsize \href{https://edwebcontent.ed.ac.uk/sites/default/files/imports/fileManager/patient_questionnaire\%20pdf_48210488.pdf}{edwebcontent.ed.ac.uk/.../patient\_questionnaire.pdf}}
and approaches to Patient-Centered Communication Best Practices (PCCBP)\cite{king2013best}.

The Multimodal Understanding \& Handling (MUH) rubric was designed to assess competence in handling and interpreting multimodal artifacts in the context of clinical consultations, including the ability to understand medical image artifacts, to use that understanding to guide the conversation and inform a clinically accurate assessment, and to communicate the salient findings and address the patients' questions in an appropriate manner. Details are provided in Table~\ref{tab:mm_osce_rubric} in Appendix. 

\subsection{Participants}
This study involved 19 board-certified PCPs and 20 validated patient actors, with participants split between India and Canada (10/9 for PCPs, 10/10 for actors, for India/Canada respectively). The PCPs had a median post-residency experience of 6 years with inter-quartile range of 3.5-11.5 years. For quality assessment of the AMIE/PCP consultations and their post-questionnaire responses, we recruited 18 specialist physicians from India and North America over three medical specialities (Dermatology, Cardiology and Internal Medicine) to ensure diverse clinical viewpoints as well as requisite expertise. These specialists, who had a median post-residency experience of 5 years (interquartile range 4-8 years), were assigned evaluation tasks matching the medical specialty of the scenarios (for example, dermatology scenarios were evaluated by dermatologists). For each of 105 scenarios, each assigned patient actor performed two sessions, one with a PCP and another with AMIE in a randomised order, yielding a total of 210 consultations. Each of these conversations was evaluated by 3 independent specialists.

\subsection{Statistical Analysis} \label{sec:stat_analysis_method}
We analysed the results from the remote OSCE study using a mixed-effect approach to account for differences in the scenarios using random effects.
Most rating choice opions for patient actors and specialist physicians were ordinal (e.g. ranging from "Very unfavourable" to "Very favourable"), and we model them using a cumulative ordinal model with logit link function. For binary ratings (e.g. "Yes / No"), we use a Bernoulli model with logit link. For diagnostic accuracy, we again use a Bernoulli model for "Correct / Incorrect".
In all models, the scenario is used as a random intercept to model differences in difficulty or quality, and experimental arm (PCP/AMIE) as a fixed effect. For diagnostic accuracy, we also model the fixed effect of the differential size (k) using monotonic regression.
In some analyses, we also estimate preference by comparing ordinal or binary metrics, e.g. in Figure \ref{fig:main_result_figure_2}.C. Here, we model the proportion of preference for AMIE and PCPs using the one-sided $\chi^2$ test.
For ablation results (e.g. in Figure \ref{fig:ablation_autoeval}), we test differences in metrics using Mann-Whitney U tests.

%% file: results.tex
\section{Results}

This section details the performance of the multimodal AMIE system, evaluating its diagnostic capabilities, conversational quality, and multimodal reasoning. We first present the results from our rigorous comparison against primary care physicians (PCPs) in randomised, blinded OSCE-style assessments involving multimodal text-chat consultations (Section \ref{sec:results_osce}). These findings demonstrate that multimodal AMIE generally matches or exceeds PCP performance across a diverse array of clinically important axes, including history-taking, diagnostic accuracy, management reasoning, communication skills, empathy, and crucially, the understanding and handling of multimodal data. Subsequently, we report findings from our automated evaluation framework (Section \ref{sec:results_automated_evals}), encompassing medical visual question-answering perception tests and ablation studies, which further validate AMIE's core capabilities and the effectiveness of our system design choices.

\begin{figure}
    \centering
    \includegraphics[width=\textwidth]{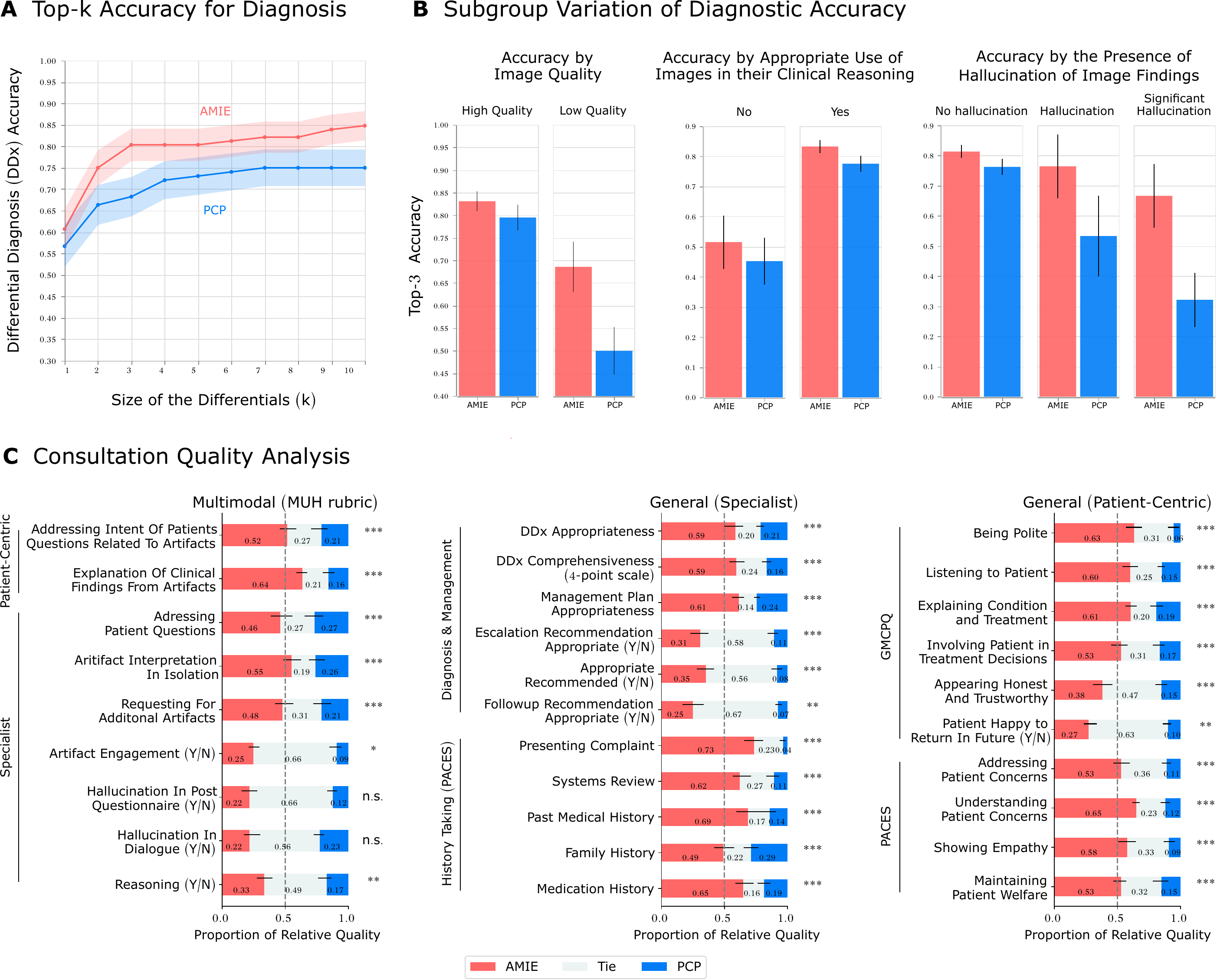}
    \caption{\footnotesize \textbf{Comparison of OSCE metrics between PCPs and AMIE}. Across all three panels, the error bars represent $95\%$ confidence intervals estimated based on $10^4$ bootstrap samples. \textbf{A. Top-k differential diagnosis (DDx) accuracy}. Both AMIE and PCPs have the opportunity to submit a differential diagnosis list (at least 3, up to 10 plausible items, ordered by likelihood) for each dialogue. We compute the top-k accuracy using an auto-rater that compares each of the 10 diagnoses with the ground truth. AMIE outperforms PCPs in diagnostic accuracy on average ($p < 0.001$).
    \textbf{B. Subgroup analysis of DDx accuracy.} We compare the top-3 accuracy across subgroups defined by three axes defined in the MUH rubric: (1) image quality, (2) image grounded reasoning and (3) hallucination of image findings in consultation (see Table.~\ref{tab:mm_osce_rubric} for details). These findings provide some insight into why AMIE may be more accurate.
    \textbf{C. Distributions of relative performance across patient scenarios.} In particular, we present the proportion of scenarios where PCPs are rated more favourably than AMIE and vice versa for three categories of OSCE rubric, namely multimodal understanding and handling rubric (right), non-multimodal specialist metrics (middle) and non-multimodal patient-centric metrics (right). Each specialist and patient actor separately rated two dialogues (one with PCP and the other with AMIE) for the same scenario, and their relative ranking is derived from their scores. For the specialist metrics, each scenario was evaluated by three different expert physicians, yielding three pairwise rankings which we aggregate by taking the majority vote. We assigned the `Tie' label to cases where two or more specialists assigned the same ratings to both PCP and AMIE consultations or the cases where the opinions on the relative performance were equally divided. Significant differences between AMIE and PCP segments calculated using a one-sided chi-squared test are indicated by asterisks ($*:p<0.05$, $**:p<0.01$, $***:p<0.001$, $n.s.: $ not significant). Figure~\ref{fig:osce_summary_results_distribution} in the Appendix shows the underlying distributions of ratings for the same set of metrics.
    }
    \label{fig:main_result_figure_1}
\end{figure}

\subsection{AMIE Matches or Exceeds PCPs on OSCE Assessment} \label{sec:results_osce}

In our remote OSCE-style evaluation of simulated consultations using multimodal text-chat, we find that on every axis of comparison, AMIE at least matches, and often outperforms, PCPs. Below we report performance on an objective measure of diagnostic accuracy, and subjective but blinded measures of patient experience and other key attributes of effective clinical consultations rated by specialist physicians.

\subsubsection{AMIE Demonstrates Strong Diagnostic Accuracy Compared to PCPs}

AMIE demonstrates superior diagnostic accuracy compared to PCPs, a critical outcome of the OSCE-style simulated clinical consultations that both AMIE and PCPs performed with patient-actors using multimodal text-chat.
As illustrated in Figure~\ref{fig:main_result_figure_1}.A, which plots the top-k diagnostic accuracy, AMIE's differential diagnosis was both more accurate and comprehensive. 

AMIE accuracy consistently outperformed PCPs across the full range of k (from 1 to 10 diagnoses). This indicates not only that AMIE's leading diagnosis (Top-1) was more frequently correct than that of PCPs following the consultation with a patient-actor, but also that AMIE's broader differential diagnosis list (Top-2 to Top-10) was more likely to contain the ground truth condition compared to the lists generated by PCPs. This overall difference in diagnostic accuracy was statistically significant ($p < 0.001$, based on a mixed-effects model accounting for scenario difficulty, see Section \ref{sec:stat_analysis_method}). We note that although we analyze accuracy up to top 10, neither AMIE nor PCPs always report 10 differential diagnoses. These findings mirror the results from previous evaluations in the text-only setting~\cite{tu2025towards}, and confirm AMIE's ability to effectively combine conversational (verbal) history with the interpretation of multimodal (visual) artifacts. 

This work advances from prior comparison of AMIE in simulated consultations by incorporating the ability for patients to upload multimodal artifacts into text-chat. To understand the extent to which our results specifically reflected this new capability, and explore the extent to which the comprehension and utilization of multimodal data contributed to AMIE's superior overall diagnostic accuracy, we performed specific subgroup analyses using specialist physicians' blinded consultation ratings (Figure \ref{fig:main_result_figure_1}.B): 
\begin{itemize}
    \item \textbf{Impact of image quality} (Figure~\ref{fig:main_result_figure_1}.B, left): When independent specialists rated the provided artifact quality as low, as anticipated, the top-3 DDx accuracy decreased for both AMIE and PCPs. However, AMIE demonstrated greater robustness; its performance drop was significantly smaller than that of PCPs, indicating the potential for better handling of suboptimal visual data common in telehealth.
    
    \item \textbf{Impact of artifact use in reasoning} (Figure~\ref{fig:main_result_figure_1}.B, middle): When specialists judged that the consulting agent (AMIE or PCP) appropriately used the visual artifact in their reasoning, diagnostic accuracy improved similarly for both. This validates the scenarios as effective tests of multimodal reasoning, underscoring that meaningful integration of visual data is crucial for diagnostic success by both AI and human clinicians. It also confirms that AMIE's superior performance stemmed from utilizing image-based diagnostic information, not solely from information obtained purely through text analysis.
    
    \item \textbf{Impact of misreporting/hallucination} (Figure~\ref{fig:main_result_figure_1}.B, right): We examined consultations where specialists identified instances of the clinician or AMIE reporting findings not actually present in the artifact (which we refer to as `hallucination' here). Such misreporting negatively impacted diagnostic accuracy for both, but significantly more so for PCPs compared to AMIE. 
    This finding might suggest that AMIE, perhaps through its structured reasoning or iterative checks, is better able to overcome intermediate misinterpretations to reach an accurate final diagnosis.
\end{itemize}

\begin{figure}
    \centering
     \includegraphics[width=0.95\textwidth]{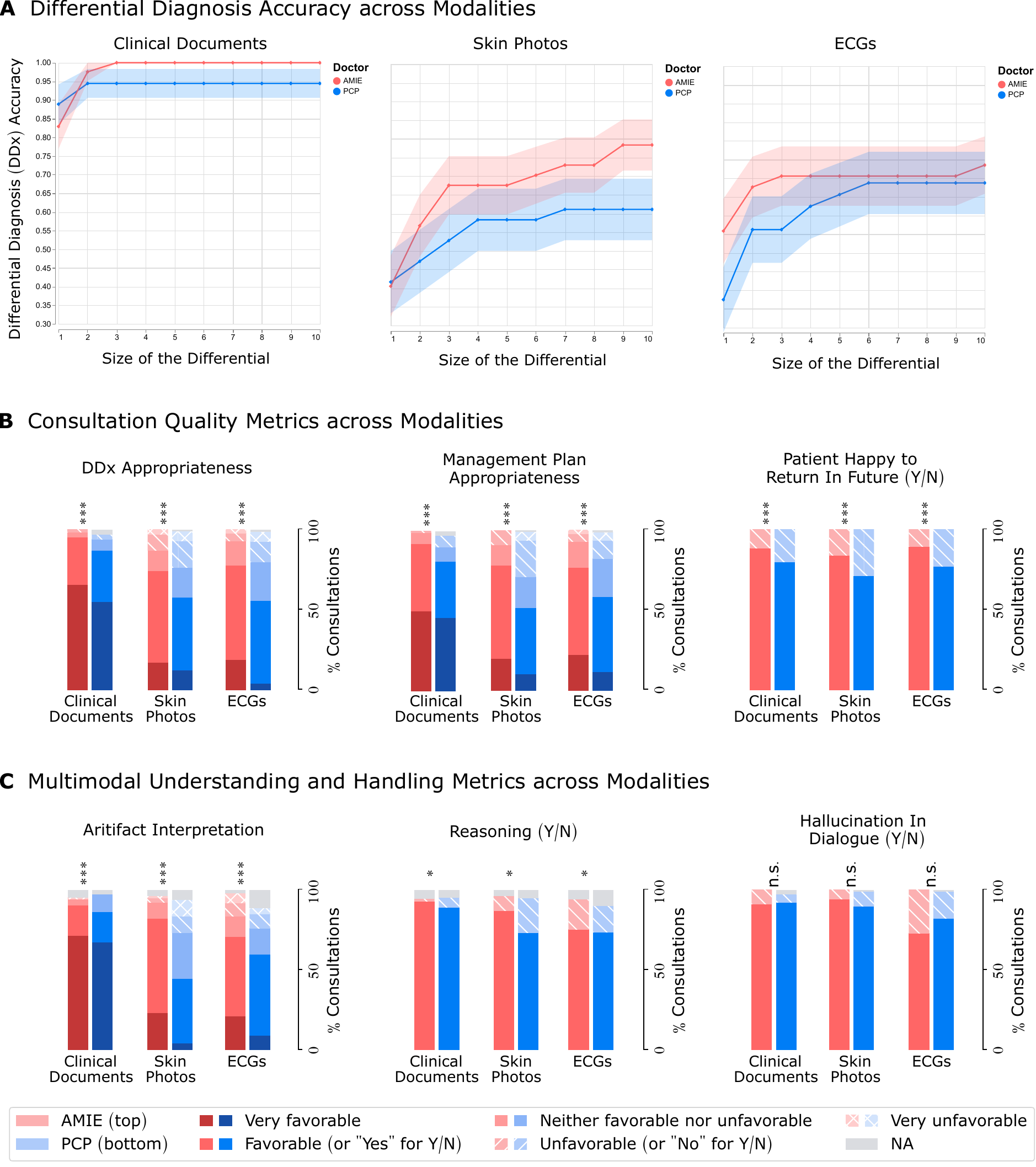}
    \caption{
        \footnotesize
        \textbf{Variation across modality types.} \textbf{A}. The average top-k differential diagnosis (DDx) accuracy is shown as a function of the size of the differentials (k) for three artifact types. \textbf{B.} The panel shows the distributions of specialist physicians' ratings (DDx appropriateness and Management Plan Appropriateness) and patient actors' ratings (Patient Happy to Return in Future), which attempts to capture the overall user satisfaction of the consultation. The distributions of ratings are displayed separately for the sessions with PCPs and AMIE, and for modality types. \textbf{C.} The distributions of scores for three MUH (multimodal understanding and handling) metrics are displayed. For illustration purposes, all responses from five-point rating scales were mapped to a generic five-point scale ranging from `Very favorable' to `Very unfavorable'. For Yes/No questions as indicated by `(Y/N)', a (positive) `Yes' response was mapped to the same color as `Favorable' and a (negative) `No' response to the same color as `Unfavorable'.
        The McNemar test was used for each evaluation axis, and p-values were adjusted using the false discovery rate (FDR) correction. Asterisks represent statistical significance ($*:p<0.05$, $**:p<0.01$, $***:p<0.001$, $n.s.: $ not significant). 
    }
    \label{fig:main_result_figure_2}
\end{figure}

We also analysed performance variation across the different types of multimodal data used in our OSCE scenarios: skin photos (SCIN), ECG tracings (PTB-XL), and clinical documents (Figure \ref{fig:main_result_figure_2}.A). Here we used a monotonic regression model on $k$ (position of diagnosis in the differential used to compute accuracy), with additional effects estimated for AMIE versus PCPs, and modalities. Both AMIE and PCPs were more accurate for scenarios based on clinical documents ($p < 0.001$). While the overall higher accuracy of AMIE compared to PCPs was maintained in the scenarios based on clinical documents, the differences were not statistically significant for those based on ECGs and skin photos. 

\subsubsection{High Conversation Quality as Rated by Patient Actors}
\label{sec:patient-results}
We compared AMIE and PCPs on the patient-centric quality metrics that were collected upon conclusion of the consultations. Patient actors often rated the interactions with AMIE as non-inferior or superior to those with board-certified PCPs across nearly all dimensions (see Figure~\ref{fig:main_result_figure_1}.C right). This included aspects often considered uniquely human strengths, such as being polite, listening, explaining conditions, involving patients in decisions, appearing honest/trustworthy, building rapport, showing empathy, and managing patient concerns (GMCPQ and PACES criteria, $p < 0.01$ for all axes). This shows that AMIE attains high conversation quality from the patient actors' perspective in the new multimodal scenarios beyond the text-only setting considered in the prior work \cite{tu2025towards,Palepu2025-de}. 

We also asked patient actors two questions specifically about the nature of multimodal interaction (see Table \ref{tab:mm_osce_rubric} in Appendix). The two questions, rated from 1 to 5, asked how well the doctor addressed the patients' questions about image artifacts, and how well the doctor explained the findings from the artifacts. Figure~\ref{fig:main_result_figure_1}.C (left) shows that AMIE was rated more highly than clinicians on both questions ($p < 0.01$).

\subsubsection{Specialist Evaluation Confirms AMIE's Strong Performance and Robustness}
\label{sec:specialist-results}

A group of 18 specialists in dermatology, cardiology, and general practice (internal medicine) evaluated AMIE and PCPs in terms of numerous attributes of effective diagnostic conversations including multimodal reasoning, history-taking, diagnostic accuracy, management reasoning, communication skills, and empathy. Aggregated results are presented in Figure \ref{fig:main_result_figure_1}.C. To account for potential inter-disciplinary variations, we also present results by specialty (Figure \ref{fig:main_result_figure_2}). The conversations held by AMIE were consistently rated more highly by specialists overall and across three disciplines. This is reflected in all items in our assessment, across Diagnosis and Management and quality of history taking. The positive assessments of the diagnostic dialogues AMIE conducted were also supported by results in the MUH component of our assessment: specialists assigned higher ratings on average to AMIE's interpretation of the multimodal artifacts, its reasoning about these artifacts, and the way it handled patient questions and concerns about the multimodal artifacts.

This result is also reflected in all three disciplines studied. AMIE provided more appropriate diagnoses and management plans, and patients were more happy to return for a second visit with AMIE in our three fields (Figure \ref{fig:main_result_figure_2}.B; all $p < 0.001$). Similarly, in all three disciplines, AMIE performed more accurate interpretation of artifacts and reasoning about those artifacts, with no increase in hallucination or mis-reporting (Figure \ref{fig:main_result_figure_2}.C; all $p < 0.001$, $p < 0.05$, and $p > 0.05$ respectively). This consistent superiority in all disciplines was broadly mirrored in the top-k accuracy (Figure \ref{fig:main_result_figure_2}.A), although the superiority was only statistically significant for clinical documents, potentially due to the smaller sample size. In Appendix \ref{sec:appendix-qualitative-examples} we provide qualitative examples from the three disciplines, of dialogues between AMIE and a patient actor and between a PCP and patient actor where AMIE has correctly identified the most probable diagnosis. For each discipline, we use the same patient scenario for the dialogue with AMIE and with the PCP, to facilitate a qualitative comparison.

Robust performance in multimodal handling, confirmed by specialist evaluations, provides critical validation for AMIE's system design. In the eyes of specialists across three disciplines, AMIE demonstrated significantly higher proficiency than PCPs not just in basic artifact engagement, but crucially, in the accurate interpretation and clear communication of findings derived from images and documents in conversation. This capability to effectively process, reason, and discuss multimodal information is fundamental for AI systems aiming to function reliably in real-world clinical workflows, particularly in telehealth where such data exchange is integral to diagnosis and patient management. 

\subsection{Automated Evaluations Validate AMIE Capabilities and Design} \label{sec:results_automated_evals}

Beyond the OSCE study, this section details automated evaluations validating multimodal AMIE's core capabilities and design. We first assess the base model's fundamental perception of medical artifacts (Section \ref{sec:results_perception}). We then use simulated dialogues and auto-rating to quantify the significant diagnostic improvements gained from our state-aware reasoning framework and conversational context, confirming system robustness (Section \ref{sec:results_ablations}). These results provide strong quantitative support for our system design choices.

\subsubsection{Base Model Provides Necessary Medical Multimodal Grounding} \label{sec:results_perception}

To ensure accurate processing of artifacts in multimodal diagnostic conversations, we first verified the perceptual capabilities of our base model, Gemini 2.0 Flash, chosen for its balanced multimodal understanding, reasoning, and conversational fluency. We specifically verified its perceptual competence using OSCE-relevant benchmarks – SCIN (skin images), PTB-XL (ECG tracings), ECG-QA (ECG tracings), and ClinicalDoc-QA (clinical documents) – comparing its performance against previous Gemini versions, including Gemini 1.5 Flash and Gemini 1.5 Pro (see Appendix \ref{app:perception_test_results}, Figure \ref{fig:perception_test}). This assessment confirmed robust capabilities for Gemini 2.0 Flash in interpreting dermatology images (SCIN) and clinical documents (ClinicalDoc-QA), demonstrating reliable visual feature extraction and information processing. Performance on the ECG-related benchmarks (PTB-XL, ECG-QA), while sufficient for a baseline and showing incremental gains over prior versions, reflected the inherent complexities of ECG interpretation from static images. Overall, this verification demonstrated that Gemini 2.0 Flash possessed the necessary foundational perceptual grounding across our target modalities, consistent with an observed trend of improving capabilities in newer model generations. This provided the confidence required to build AMIE's advanced reasoning and dialogue functions upon this base model.

\subsubsection{State-Aware Reasoning Boosts Performance and Demonstrates Robustness} \label{sec:results_ablations}

To further understand the contributions of different system components to overall performance, we conducted several ablation studies using our automated evaluation framework. This section details experiments comparing the full AMIE system against simpler baselines and evaluating the specific impact of test-time reasoning, history taking, and robustness to variations in patient presentation on conversation quality, as assessed by the auto-rater.

\begin{figure}[htbp]
    \centering
    \includegraphics[width=1\textwidth,height=\textheight,keepaspectratio]{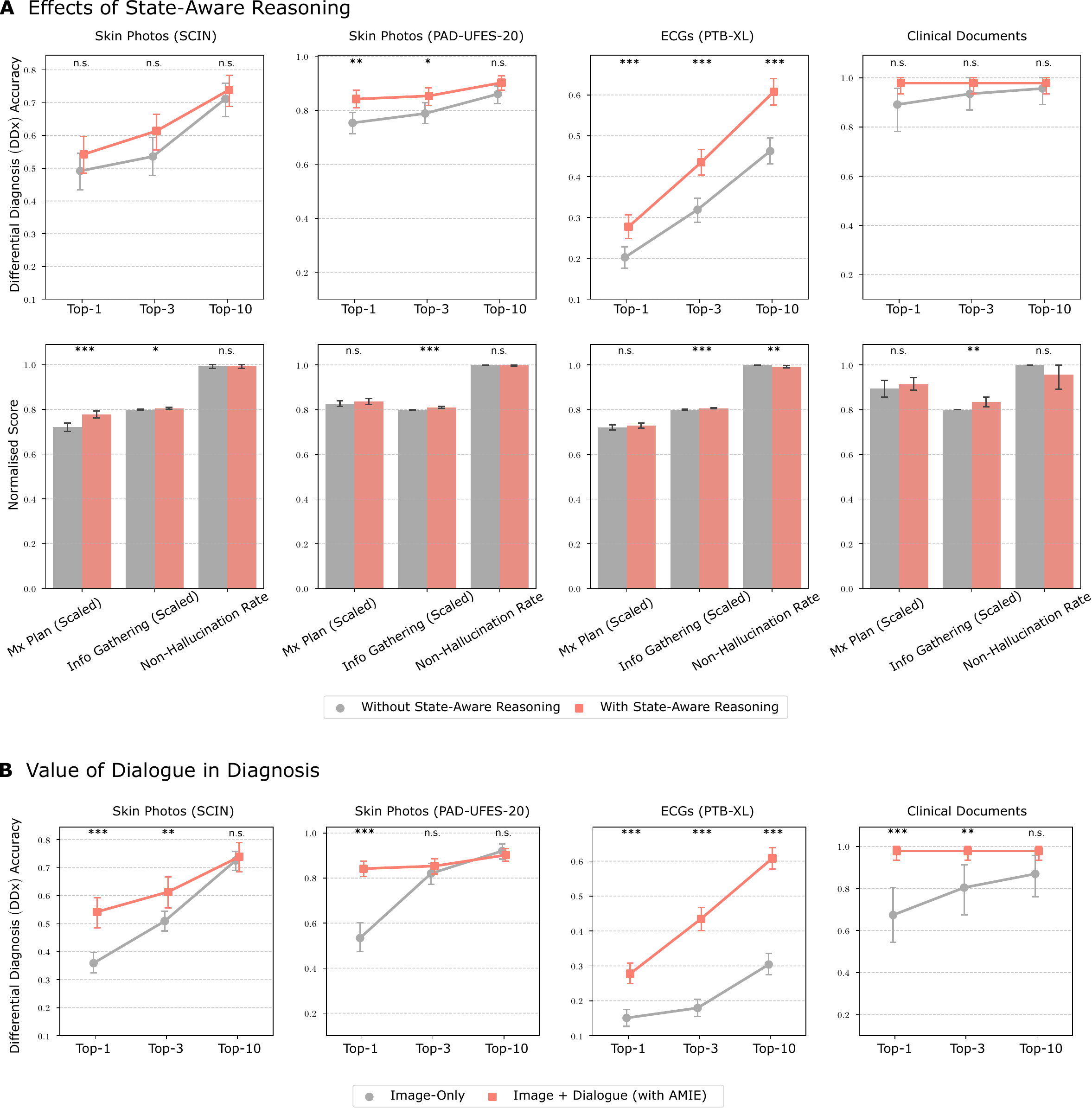}
    \caption{\footnotesize \textbf{Auto-rater evaluation to quantify the value of state-aware reasoning and dialogue-based interaction.}  \textbf{(A) State-aware reasoning ablation} compares the performance of multimodal AMIE with its state-aware reasoning against a baseline without it, using the dialogue simulation environment across four datasets: skin photos (SCIN), skin photos (PAD-UFES-20), ECG traces (PTB-XL), and clinical documents (created by our clinical providers). The metrics include Top-k differential diagnosis (DDx) accuracy, management (Mx) plan appropriateness (scaled), quality of information gathering (scaled), and non-hallucination rate (the proportion of cases without any hallucination). State-aware reasoning consistently improves performance across metrics and datasets. \textbf{(B) Dialogue ablation} aims to quantify the value of dialogue-based interaction by comparing the top-k DDx accuracy of the baseline LLM that infers the differentials directly from the image artifacts (denoted by ``Images-only'') against that of the AMIE system which leverages the combination of provided images and conversational history (``Images + Dialogue (with AMIE)''). Incorporating dialogue significantly boosts diagnostic accuracy relying on static image input alone. Error bars represent $95\%$ confidence intervals derived from $10^4$ bootstrap samples. Significance markers denote p-values obtained from a two-sided Mann-Whitney U test comparing the two conditions within each panel/metric. Asterisks represent statistical significance ($*:p<0.05$, $**:p<0.01$, $***:p<0.001$, $n.s.: $ not significant).
    }
    \label{fig:ablation_autoeval}
\end{figure}

\paragraph{State-aware reasoning improves dialogue quality}

One of the key elements of our system is the state-aware dialogue phase transition framework (detailed in Section \ref{sec:reasoning_method}). In order to assess its contribution, we compared its diagnostic performance and dialogue quality against a baseline model lacking this explicit reasoning structure. We hypothesized that the dynamic phase transitions and uncertainty-driven questioning employed by our framework would lead to more accurate diagnoses compared to a simpler approach.

Using our simulation environment and auto-rater (Section \ref{sec:simulation_method}), we evaluated conversations generated for synthetic scenarios derived from PAD-UFES-20, SCIN, PTB-XL, and ClinicalDoc-QA datasets. We compared two conditions: (1) multimodal AMIE with the full test-time reasoning framework, and (2) a ``vanilla'' baseline using the same Gemini 2.0 Flash model but without the explicit state-aware reasoning framework – relying only on domain-specific instructions and lacking the dynamic phase transitions and uncertainty-based questioning components.

Results (Figure \ref{fig:ablation_autoeval}.A, Table \ref{tab:reasoning_ablation} in Appendix \ref{app:reasoning_ablation}) strongly support our hypothesis. Across all datasets (PAD-UFES-20, SCIN, Clinical Documents, PTB-XL), AMIE with reasoning consistently outperformed the vanilla baseline on key diagnostic metrics.
Specifically, Table \ref{tab:reasoning_ablation} in Appendix \ref{app:reasoning_ablation} shows AMIE with reasoning achieved higher DDx accuracy. Notable Top-1 accuracy gains occurred on Clinical Documents (0.89 to 0.98), PAD-UFES-20 (0.75 to 0.84), and PTB-XL (0.20 to 0.28). Reasoning also improved Information Gathering scores (e.g., Clinical Documents: 4.00 to 4.17). Hallucination rates remained negligible in both settings, although is sometimes higher for state-aware reasoning. Management Plan (Mx Plan) appropriateness also improved (e.g., SCIN: 3.69 to 3.89), indicating more clinically relevant output with reasoning.

\paragraph{History taking improves diagnostic accuracy}
To understand the additional performance gained with history taking versus analyzing multimodal artifacts alone, we conducted an ablation comparing two settings: (1) ``Images-only'', where the base model diagnosed from images without dialogue, omitting history taking; and (2) ``Images + Dialogue'', where multimodal AMIE used images, full dialogues, and its state-aware reasoning at inference.

Results (see Figure \ref{fig:ablation_autoeval}.B) show history taking is crucial. The ``Images-only'' setting yielded markedly lower accuracy; for instance, Top-1 accuracy on PAD-UFES-20 dropped significantly without dialogue. Conversely, the ``Images + Dialogue'' setting consistently improved performance across all datasets (SCIN, PAD-UFES-20, PTB-XL, Clinical Documents). It is important to note that this is not merely using two pieces of independent evidence: AMIE is integrating the information from both images and dialogue by cross-referencing them with each other.

\paragraph{Robust diagnostic conversation with patient scenario augmentations}

To evaluate system robustness against variations in patient presentation, we used LLM-driven augmentation to introduce variations to the original patient scenarios across three axes: personality styles (affecting interaction dynamics), demographics (testing fairness), and semantic changes to background/symptoms (evaluating resilience to minor history variations). Assessing performance under these conditions is important for understanding real-world applicability. The results, detailed in Figure \ref{fig:ablation_robustness} (Appendix \ref{app:patient_augmentations}), demonstrate that AMIE's performance across key auto-rater metrics—including diagnostic accuracy, information gathering, hallucination rate, and management plan appropriateness remained highly consistent between the original and augmented scenarios. This stability indicates AMIE's robustness to these non-clinically significant variations in how patients might present.

%% file: related.tex
\section{Related Work}

To contextualize multimodal AMIE and highlight its contributions, this section reviews prior research in the key areas of diagnostic conversational agents, multimodal AI in healthcare, the evaluation of such systems, and the platforms enabling modern telemedicine.

\paragraph{Diagnostic conversational agents}

Early diagnostic dialogue systems such as MYCIN \cite{excerpt_mycin} and INTERNIST-I \cite{INTERNIST_1982} relied on rules-based approaches. These systems used predefined rules and knowledge bases, limiting their generalizability and conversational fluidity compared to modern LLM-based agents. AMIE \cite{tu2025towards}, optimized for diagnostic dialogue via self-play, outperformed primary care physicians (PCPs) in simulated text-based consultations across clinically relevant metrics like history-taking and diagnostic accuracy, highlighting the potential of LLMs to capture the nuances of medical conversations. Similar technologies are beginning to see deployment, for instance, ``Mo'' \cite{li2024conversational}, a medical advice chat service with physician safety oversight, demonstrating increased patient satisfaction and information clarity in a real-world setting. Beyond general capabilities, research is also focused on modeling clinicians' diagnostic reasoning within these agents. The Emulation Framework \cite{xu2024reasoninglikedoctorimproving} seeks to align medical dialogue systems with clinicians' reasoning by incorporating abductive/deductive analysis, while the IADDx framework \cite{xu2024medicaldialoguegenerationintuitivethenanalytical} explicitly models the differential diagnosis process. Recognizing the impact of cognitive biases, multi-agent frameworks based on GPT-4 \cite{ke2024enhancingdiagnosticaccuracymultiagent} and constellation architectures like Polaris \cite{mukherjee2024polaris} simulate clinical team dynamics or specialist checks to improve diagnostic accuracy by identifying and correcting misconceptions. Further research shows that dialogue-based fine-tuning can improve LLM reasoning capabilities and robustness \cite{liu2025dialoguebettermonologueinstructing}. Despite progress, existing agents primarily operate in text-only settings, facing limitations in replicating the full spectrum of clinical reasoning which often involves interpreting multimodal data.

\paragraph{Multimodal AI in healthcare}
The integration of diverse modalities including images, text, audio, and physiological signals transforms the potential for AI applications in healthcare, potentially enabling tools that could mirror how clinicians synthesize information for a more comprehensive patient understanding \cite{artsi2024advancing}. Studies show multimodal models can outperform text-only counterparts, achieving higher accuracy and a more holistic view \cite{schouten2024navigatinglandscapemultimodalai, Xiao_2025} while also permitting new use-cases. For example Vision-Language Models (VLMs), combining LLMs with vision encoders, show significant potential in radiology report generation and enable informative conversations or visual querying of medical images \cite{Yildirim_2024, simon2024future}, a previously inaccessible task for AI systems in medical imaging. Multimodal approaches are being explored across various diagnostic tasks, such as COVID-19 detection using chest X-rays, text data, and cough detection \cite{islam2025multimodalmarvelsdeeplearning}, and interpreting ECG signals to generate clinician-like reports \cite{bleich2024automatedmedicalreportgeneration}. 

However, significant challenges remain for ongoing research and development, including the optimal management of data heterogeneity, effective fusion strategies between modalities, and ethical considerations like bias, fairness, privacy, and interpretability \cite{schouten2024navigatinglandscapemultimodalai, artsi2024advancing}.

\paragraph{Evaluation of dialogue and multimodal artifacts}

Rigorous, clinically-relevant evaluation is paramount for the safe deployment of AI in healthcare. Objective Structured Clinical Examinations (OSCEs) \cite{Harden1975-ft}, a standard for assessing clinical skills in medical education, provide a helpful framework for the evaluation of conversational AI systems in medicine. Our work adapts the OSCE model, using validated patient actors and clinically relevant multimodal scenarios derived from real clinical practice. While multi-agent conversational evaluation frameworks offer scalable assessment methods, they may lack the clinical fidelity of human evaluations. Benchmarks like GMAI-MMBench \cite{chen2024gmaimmbenchcomprehensivemultimodalevaluation} assess multimodal capabilities on specific tasks, and simulation environments like AgentClinic \cite{schmidgall2024agentclinic} evaluate agents in hypothesized clinical settings. However, a need persists for evaluation metrics specifically tailored to multimodal diagnostic dialogue, which motivated our contribution of assessment rubric components specifically intended for assessing  multimodal data handling and reasoning.

\paragraph{Conversational AI with multimodality}

While multimodal models are advancing in many domains including healthcare, integrating multimodal perception capabilities effectively into diagnostic conversations remains an active research area. VLMs like LLaVa-Med \cite{LLaVaMed_2023} and Med-Flamingo \cite{MedFlamingo_2023} demonstrated capabilities in medical visual question answering alongside the potential for dialogue about images. Domain-specific models, such as SkinGPT-4 \cite{zhou2024pre}, focus on specific modalities like dermatology images. These models provide crucial building blocks but often focus on answering questions or conducting conversations about a specific modality, rather than on fully-integrated multimodal perception. This broader integration is necessary for an AI system to effectively manage and conduct a high-quality clinical consultation. Frameworks like CRAFT-MD \cite{johri2025evaluation} explore approaches for evaluating LLMs using conversational interactions but may not systematically focus on the agent's uncertainty-directed acquisition of multimodal information, or the extent to which onward appropriate multimodal reasoning informs the consultation and diagnostic process. Earlier approaches like MINT \cite{freyberg2024mintwrappermakemultimodal} and A2MT \cite{kossen2022active} explored multimodal intent recognition or sequential data acquisition, but our work aims for a more flexible, dynamic approach where the agent's state and diagnostic uncertainty guide multimodal information gathering, mirroring clinical practice more closely.

\paragraph{Use of multimodal chat messaging platforms in clinical care}

The delivery of remote care has frequently been reported utilizing multimodal chat messaging in widely accessible instant messaging applications, especially in low and middle income countries (LMICs) where traditional electronic health record  portals are less common \cite{SudanWhatsApp_2021, TamilNaduWhatsApp_2024}. These common consumer platforms facilitate asynchronous communication between clinicians and patients and enable text consultations alongside the sharing of multimodal artifacts like photos (e.g., skin lesions \cite{wong2018real}), documents (e.g., lab reports), and potentially even specific investigations (which could for example include ECGs) \cite{giordano2017whatsapp}. While practical, their use has raised concerns regarding patient confidentiality, data governance, and integration with formal medical records \cite{UKWhatsAppPolicy_FT_2022}. Policy has adapted but remains local in nature; with bodies like NHS England recently issuing guidance permitting their use under specific circumstances (e.g., during the pandemic) where benefits outweighed risks \cite{UKWhatsAppPolicyRef_2024}. Messaging is also used for inter-clinician communication \cite{john2022smartphone} and has established utility in simpler interventions like promoting medication adherence \cite{Thakkar2016Mobile} or delivering psychological support \cite{Buddhika2019Review, PisaniSMSPrevention}. Our work on multimodal conversational agents extends the previous analysis of AI systems for conversational diagnostic purposes in text-only settings, to explore performance within this more realistic context.  

%% file: discussion.tex
\section{Discussion}

In this work, we sought to address a key unmet need for medical AI systems to be able to understand, reason about, and appropriately utilize information from multimodal medical data in the course of diagnostic conversations with patients. 

\paragraph{Integrating multimodal perception into clinical dialogue}
Our study demonstrates that integrating multimodal perception and reasoning into AMIE maintained the high standards of diagnostic accuracy and conversational quality previously established in text-only settings~\cite{tu2024towards}. Across the multimodal OSCE evaluations, multimodal AMIE achieved diagnostic accuracy (Figure \ref{fig:main_result_figure_1}) and core conversational metrics—such as empathy, clarity, and history-taking (Figures \ref{fig:osce_patient_centred_metrics_all} and \ref{fig:specialist-eval-all})—that were superior to PCPs, even with more challenging and realistic patient scenarios that require interpretation of images or documents.

While foundation models like Gemini 2.0 provide powerful multimodal capabilities (Section \ref{sec:results_perception}, Figure \ref{fig:perception_test}), the critical and challenging step is effectively weaving artifact interpretation into the dynamic flow of a clinical dialogue without degrading the interaction. AMIE achieved this via our state-aware reasoning framework as detailed in Section \ref{sec:reasoning_method}. This enabled the system to handle multimodal data mid-interaction while preserving clinically sound and empathetic dialogue. This capability to effectively process and integrate information beyond text is essential for developing AI systems intended for real-world telehealth applications where such multimodal data exchange is routine~\cite{Zeltzer2023-iu,Campanozzi2023-xl}. The specific modalities tested in this study reflect the broad and expanding breadth of modalities that may be present in telehealth settings. For instance, clinical photographs captured by smartphones and photographs of test results or reports are readily accessible by the everyday consumer. Additionally, ECG is the most common heart medical exam ordered in primary care \cite{ashley2000prevalence}. While the encounter of this modality may remain infrequent in telemedicine, it is increasingly becoming more available on various consumer-grade wearable devices, which enable the recording of both single-lead ECGs, and with some guidance on positioning, multi-lead configurations~\citep{li2022using}.

Notably, AMIE demonstrated proficiency in handling and reasoning about the multimodal artifacts provided by the patients during the consultations. Assessed in a blinded fashion using our dedicated Multimodal Understanding \& Handling (MUH) rubric (Table \ref{tab:mm_osce_rubric}), specialist physicians gave AMIE strikingly high scores compared to PCPs on an array of axes encompassing artifact engagement, interpretation in isolation, and artifact-grounded reasoning (Figure \ref{fig:specialist-eval-all}).

Moreover, patient actors rated AMIE significantly higher than PCPs in its ability to address the intent of their questions regarding artifacts and to explain the clinical findings derived from this visual evidence (Figure \ref{fig:osce_patient_centred_metrics_all}). While caution is warranted when generalising these findings to broader remote care delivery—given that clinician performance might have been constrained by the text-chat modality as opposed to richer video-call interactions—it remains notable that AMIE was perceived as superior in these specific communicative tasks. This result suggests patients in this setting valued AMIE's systematic and explicit approach to verbalizing visual findings, potentially more so than the implicitly understood assessments sometimes offered by clinicians. One such case is shown in Figure \ref{fig:amie_derm_dialogue_1_page1}. After the patient sends images of their arm, AMIE summarises the image and relates the findings to things mentioned by the patient earlier in the conversation.  In contrast, the PCP (Figure \ref{fig:pcp_derm_dialogue_1_page_1}) in that case did not directly address the image artifacts shared, even though they incorporate the images in their reasoning. Explicitly addressing the images, as AMIE did, may have provided reassurance that the images were transmitted correctly, that the doctor was looking at the correct part of the image, and reached a conclusion that was helpful for further assessment and diagnosis. We regard this communication attribute as an important aspect of a multimodal interaction, as patients may expect explanations that refer to the visual evidence they provide.

The scenarios in this study were designed to ensure that successful diagnosis and management required appropriate reasoning and utilization of multimodal information. Subgroup analyses were supportive of the validity of this experimental design. Both AMIE/PCP diagnostic accuracy dropped in cases where the image quality was believed to be inferior; if the system or clinicians performed well in these tests irrespective of image quality or the integrity of their multimodal reasoning, it might suggest that high-quality consultation was not dependent on information available only in the visual upload to the chat. Instead, the parallel drop in performance (seen when reasoning over the picture was flawed or image quality was poor) confirms that the multimodal inputs were indeed critical to the diagnostic process for both AMIE and PCPs, validating these scenarios as effective tests of multimodal reasoning. Further support for the validity of the scenarios was seen in how diagnostic accuracy improved for both AMIE and PCPs where they were rated as having appropriately used the visual information in their reasoning. These findings lend greater credence to the promising indications of AMIE's greater robustness to lower-quality images and its ability to overcome intermediate misinterpretations of multimodal data.

\paragraph{Robustness, reliability, and equity considerations}

Ensuring both reliability and equity is essential for the responsible development of diagnostic AI systems, although achieving this requires extensive dedicated further study and validation. Our evaluations undertook initial steps to explore these aspects through robustness testing. The automated evaluations using LLM-driven augmentations of patient scenarios (Appendix \ref{app:patient_augmentations}, Figure~\ref{fig:ablation_robustness}) shows that the core performance metrics of AMIE remained largely stable with respect to variations in simulated patient personality, demographics, and minor semantic changes to clinical scenarios. Specialist evaluations in the OSCE study (Figure~\ref{fig:main_result_figure_1}.B) further examined factors influencing diagnostic accuracy.
Notably, AMIE demonstrated greater robustness in diagnostic accuracy against (specialist-rated) image quality than PCPs. While these preliminary findings regarding stability against multiple factors of variations in patient scenarios and image artifacts are encouraging, further research and real-world validation are necessary to ensure reliable and equitable performance across diverse populations and conditions.

\paragraph{The role of model specialization} 

Adapting general foundation models like Gemini~\cite{geminiteam2024gemini15unlockingmultimodal} for specific domains such as medicine can involve various strategies. These range from training-time adaptations like supervised fine-tuning (SFT) and reinforcement learning (RL)~\cite{ouyang2022training, bai2022constitutional}, to inference-time approaches~\cite{brown2020language, wei2022chain, wang2022self}. Training-time specialization holds the potential to significantly boost performance on specific tasks or modalities (e.g., ECG analysis) that might be underrepresented in broad pre-training datasets, as suggested by prior work~\cite{saab2024capabilities}. However, this approach not only requires significant computational resources but also carries the risk of catastrophic forgetting of the existing general capabilities, which could for instance degrade instruction following performance or general conversational competence. To investigate the benefits of domain-specific SFT, we compared the fine-tuned Gemini 2.0 Flash model with the original one using the dialogue simulation environment described in Figure~\ref{fig:simulation}. While the experiments displayed potential gains on specific tasks such as ECG analysis, they also highlighted performance degradation on other important aspects (e.g., management plan appropriateness) relevant to a holistic consultation (Appendix~\ref{app:sft_ablation}).

In this work, we focused on leveraging a strong, general-purpose base model enhanced with a domain-specific inference-time strategy. We selected Gemini 2.0 Flash as the base model due to its overall superior out-of-the-box performance in multimodal understanding, reasoning, and conversational fluency compared to the other variants available and evaluated during development (including the specialized Gemini 2.0 Flash variants, Appendix \ref{app:sft_ablation}). This strong baseline, likely benefiting from architectural advancements and extensive pre-training, provided a robust foundation for the state-aware reasoning framework detailed in Section \ref{sec:reasoning_method}. While training-time specialization remains an important area for future investigation, our findings in this study highlight the significant effectiveness achieved by focusing on inference-time enhancements, a strategy also proven successful for the distinct challenge of management reasoning in related AMIE work on disease management~\citep{Palepu2025-de}.

The state-based approach is also built on rich insights from human-centered research, which has revealed that user-clinician conversations typically unfold in phases. Early interactions are often characterized by the agent gathering information, while later turns focus on the agent providing information to the user~\citep{li2024conversational}. This explicit state-based strategy aligns with qualitative feedback from clinicians, who noted a preference for thorough information gathering, even if an LLM quickly reached a correct conclusion, to avoid premature assessments. However, a potential limitation of this structured approach is its rigidity when new critical information, such as an allergy or contraindication, emerges after a plan has been conveyed, suggesting a need for more fluid state transitions in such scenarios. 

\paragraph{Limitations of chat-based interactions}
Our evaluation utilized a synchronous chat interface, mirroring the text and static image exchange common on ubiquitous instant messaging platforms, whose use has also been reported for medical conversation, particularly in LMICs~\cite{giordano2017whatsapp, SudanWhatsApp_2021}. While this modality offers accessibility and convenience for certain interactions, it presents significant limitations compared to richer communication channels like video consultations or in-person visits. Chat inherently restricts the exchange of non-verbal cues, limits the clinician's ability to perform dynamic visual assessments beyond submitted artifacts, and entirely precludes guided physical examinations, all of which can provide crucial diagnostic information. Furthermore, the richer, more interactive nature of video calls may foster stronger rapport and trust between the clinician and patient, elements that are vital for effective communication~\cite{ritunga2024challenges, donaghy2019acceptability}. Consequently, the scope of conditions and the depth of assessment possible within this chat-based format are inherently constrained. Additionally, specific to our comparative evaluation, the potential for unblinding due to systematic stylistic differences between AMIE and PCPs poses a challenge for comparative studies that is difficult to fully control. These collective limitations must be considered when interpreting the study's findings and contemplating real-world applications.

\paragraph{Importance of real-world validation and future directions}
While our findings demonstrate multimodal AMIE's potential in simulated consultations, significant further research is essential before considering real-world clinical translation. Future studies must rigorously evaluate AMIE's performance, safety, and reliability under the complexities and constraints of actual healthcare delivery. This includes understanding its potential impact on clinical workflows, patient outcomes, and health equity across diverse populations and conditions. We emphasize that AMIE remains an evolving research system, not intended for clinical use at this stage. 
Continued validation and responsible development are paramount as we explore the capabilities required for AI systems that might safely and effectively assist in healthcare.

%% file: conclusion.tex
\section{Conclusion}
This work addresses a key limitation in conversational diagnostic AI by advancing the Articulate Medical Intelligence Explorer (AMIE) to dynamically integrate multimodal reasoning within diagnostic conversations, enabling the system to strategically request, interpret, and reason about medical artifacts based on conversational context and evolving clinical understanding. This was achieved through a novel state-aware inference-time reasoning technique leveraging the advanced multimodal and language capabilities of Gemini 2.0 Flash. In a rigorous evaluation using a multimodal text-based OSCE framework, AMIE's performance matched or surpassed that of primary care physicians across crucial clinical dimensions, demonstrating particular proficiency in multimodal data handling and reasoning, diagnostic accuracy, and overall consultation quality.
Although further research is essential before real-world clinical translation, the findings underscore a substantial step towards developing AI systems capable of more comprehensive and realistic clinical interactions.

%% file: appendix.tex
\section*{Appendix Overview} 

\begin{itemize}
    \item \textbf{Appendix \ref{sec:appendix-questionnaires}: Supplementary Details on Multimodal OSCE Study} \\
    Describes the materials and setup for the OSCE study, focusing on the evaluation criteria and post-consultation questionnaires.

    \item \textbf{Appendix \ref{sec:appendix-osce}: Supplementary Results from the Multimodal OSCE Study} \\
    Presents additional quantitative and qualitative results from the human evaluation (OSCE study), covering patient and specialist evaluations.

    \item \textbf{Appendix \ref{app:autorater_results}: Supplementary Automated Evaluation Details and Results} \\
    Details the methodology and presents results for the automated evaluation framework, including perception tests, auto-rater criteria, ablations, and robustness analysis.

    \item \textbf{Appendix \ref{sec:appendix-qualitative}: Qualitative Results from the Multimodal OSCE} \\
    Provides illustrative example dialogues from the Multimodal OSCE study involving both the AI system and human physicians.

\end{itemize}

\newpage

\section{Supplementary Details on Multimodal OSCE Study}

\label{sec:appendix-questionnaires}
\subsection{Patient actor questionnaire}

Following completion of the dialogue, patient actors completed a questionnaire about their experience. This questionnaire consists of the GMCPQ, and patient-centered MUH developed for this study. For full details of the questions asked in GMCPQ, please see \citet{tu2025towards}, appendix. For the questions asked about patient-centered MUH, see Table \ref{tab:mm_osce_rubric}.

\subsection{Specialist questionnaire}
Similarly, the conversation is graded by specialists using a questionnaire. This questionnaire consisted of Practical Assessment of Clinical Examination Skills (PACES), Patient-Centered Communication Best Practice (PCCBP), and specialist MUH developed by us. For details of the questions asked in PACES and PCCBP, please see \citet{tu2025towards}, appendix. For questions asked about MUH, see Table \ref{tab:mm_osce_rubric}.

\subsection{Dialogue interface used by patient actors}

Patient actors and doctors conducted conversations inside the Google Data Compute environment. While the interface matches previous studies \cite{tu2024towards, Palepu2025-de}, we extended it to enable image upload based on URLs and their rendering. 

\begin{figure}[H]
    \centering
    \includegraphics[width=0.95\textwidth]{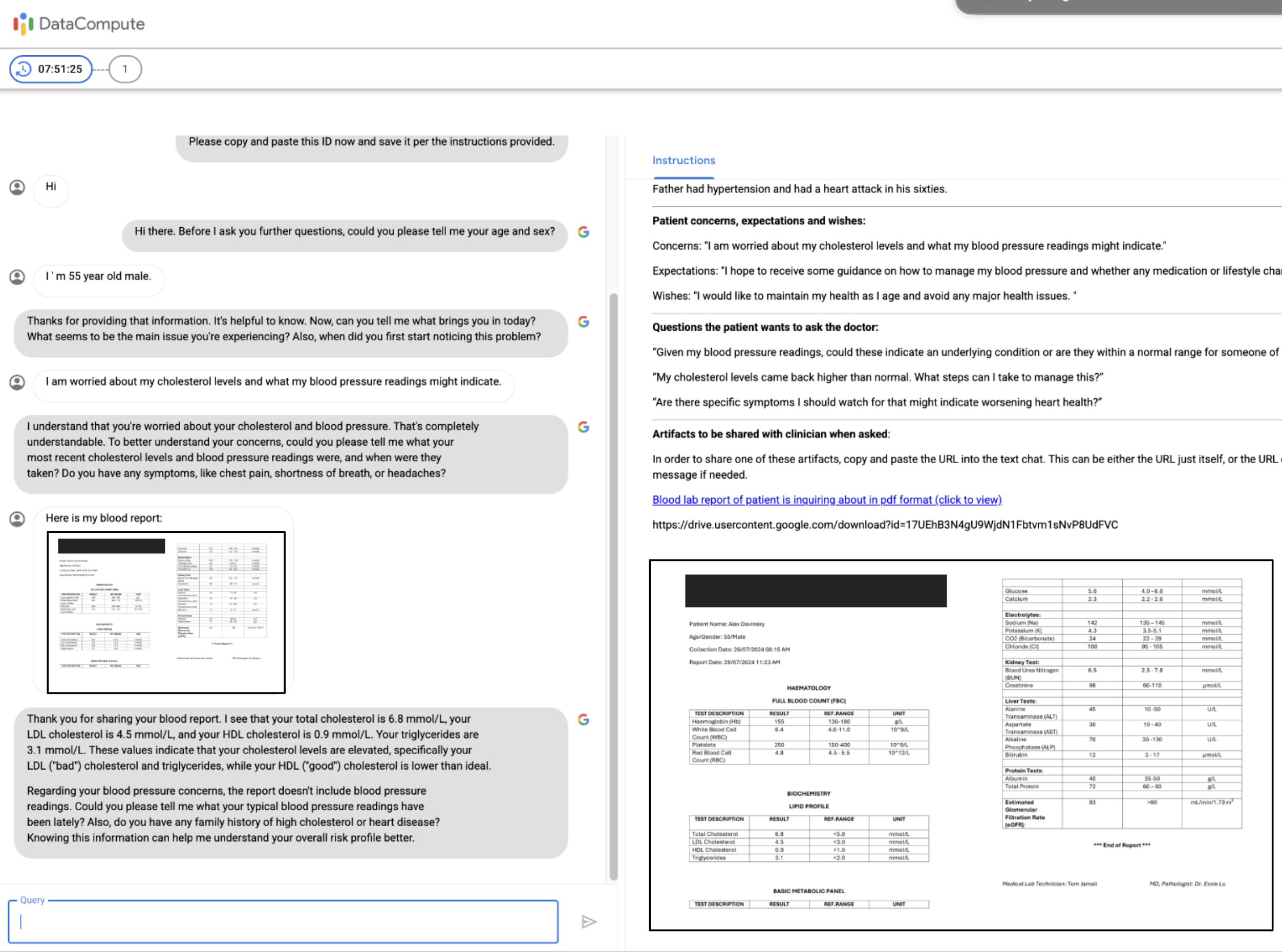}
    \caption{\footnotesize \textbf{The interface for patient actors.} Instructions and the scenario are shown on the right hand side. The patient actor interacts with either a clinician or AMIE via the text interface on the right. Images can be shared by pasting an image URL into the text field as illustrated by the clinical document example above.}
    \label{fig:patient_interface}
\end{figure}

\subsection{Interface used by specialists}

Specialists were presented with a split-screen UI in which the dialogue and the scenario are shown on the left, and questions are shown on the right. To see images, specialists could click on links to open them in new tabs.

\begin{figure}[H]
    \centering
    \includegraphics[width=\textwidth]{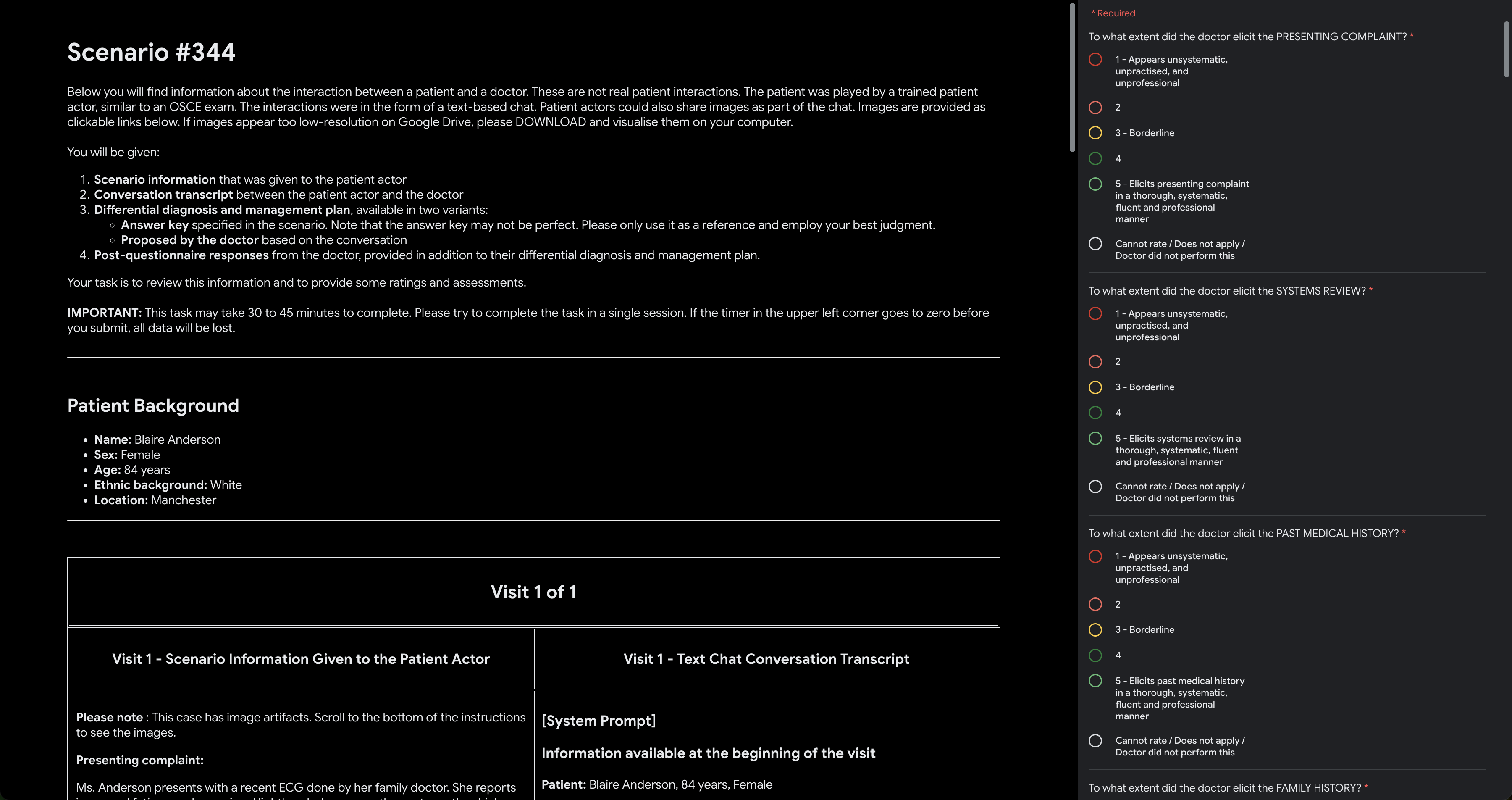}
    \caption{\footnotesize \textbf{The interface for specialists.} the dialogue and the scenario are shown on the left, and the questions are shown on the right. Specialists could scroll each side independently and click on image links to expand them.}
    \label{fig:specialist_interface}
\end{figure}

\subsection{Evaluation Rubric}

To evaluate the interaction with multimodal data, we designed the Multimodal Understanding \& Handling (MUH) rubric. This was crucial for assessing how both AMIE and the PCPs acknowledged, interpreted, and utilized the provided artifacts (skin photos, ECGs, clinical documents) within the consultation flow. Table \ref{tab:mm_osce_rubric} below provides the complete details of this MUH rubric. It breaks down each criterion, the precise questions asked of the specialist and patient actor evaluators, the scoring scales used, and the basis for each assessment. For the details of the other OSCE rubrics including PACES, PCCBP and Diagnosis \& Management, we refer the readers to Extended Data Table 1, 2 and 3 in \citet{tu2025towards}. 

\begin{table}
    \centering
    \caption{\footnotesize \textbf{Multimodal OSCE rubric details}. The table outlines the criteria, questions, scales, and assessment bases used to evaluate numerous aspects of multimodal understanding and handling capabilities of AMIE or PCPs. Specialist metrics represent the set of axes by which the specialist physicians evaluate the quality of consultations and post-questionnaires from PCPs and AMIE that summarise the key clinical findings such as differential diagnosis (DDx), management and treatment plans as well as noteworthy image findings. Patient-centric metrics on the other hand represent the questions that are answered by patient actors upon conclusion of the consultations. }
    \includegraphics[width=\textwidth,height=\textheight,keepaspectratio]{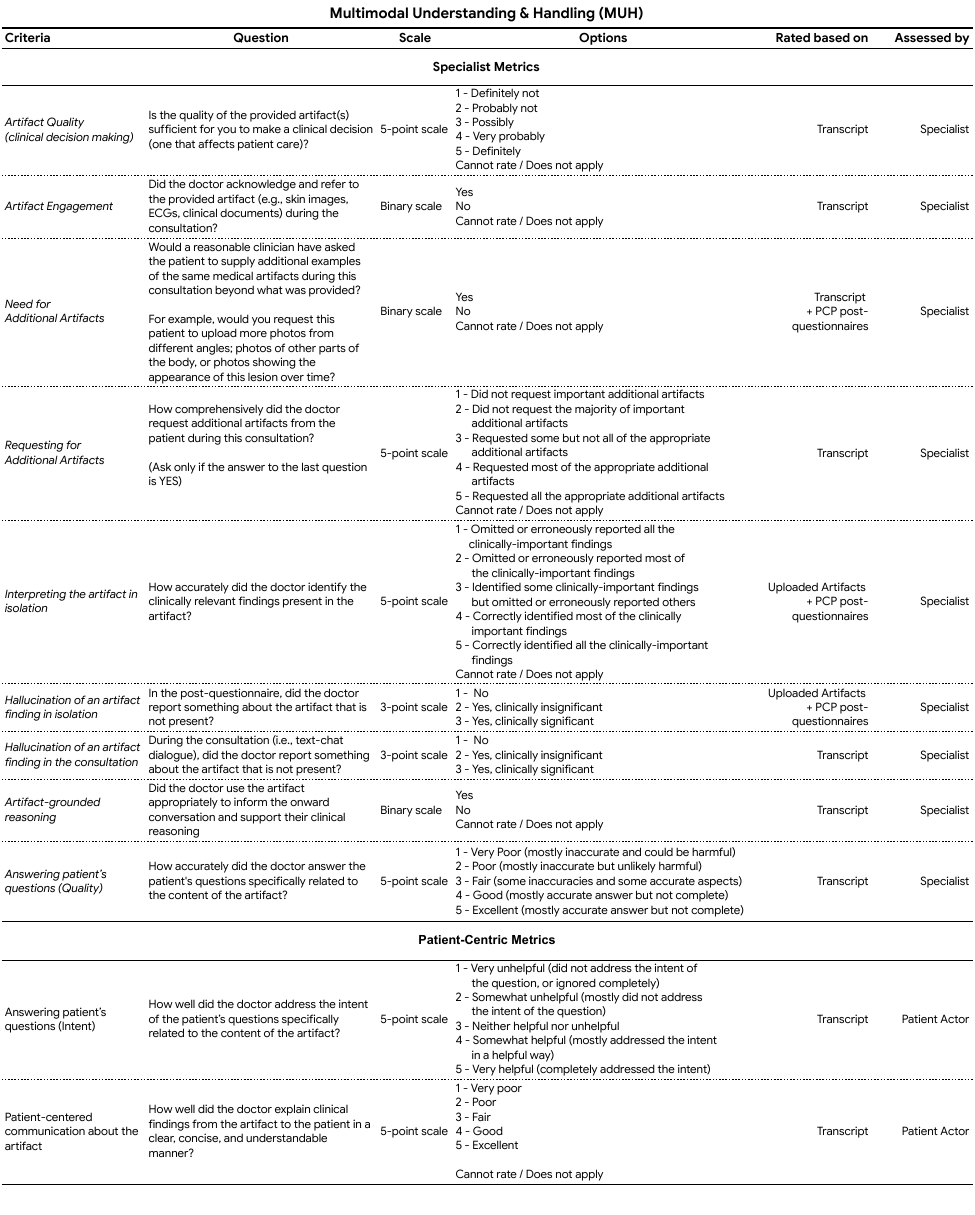}
    \label{tab:mm_osce_rubric}
\end{table}

\clearpage
\section{Supplementary Results from the Multimodal OSCE Study} \label{sec:appendix-osce}

In this section, we provide the additional analysis of the virtual OSCE study precluded from the main body of the work. 

\subsection{Rating score distributions of Figure \ref{fig:main_result_figure_1}\textbf{.C}}
\begin{figure}[H]
    \includegraphics[width=\textwidth]{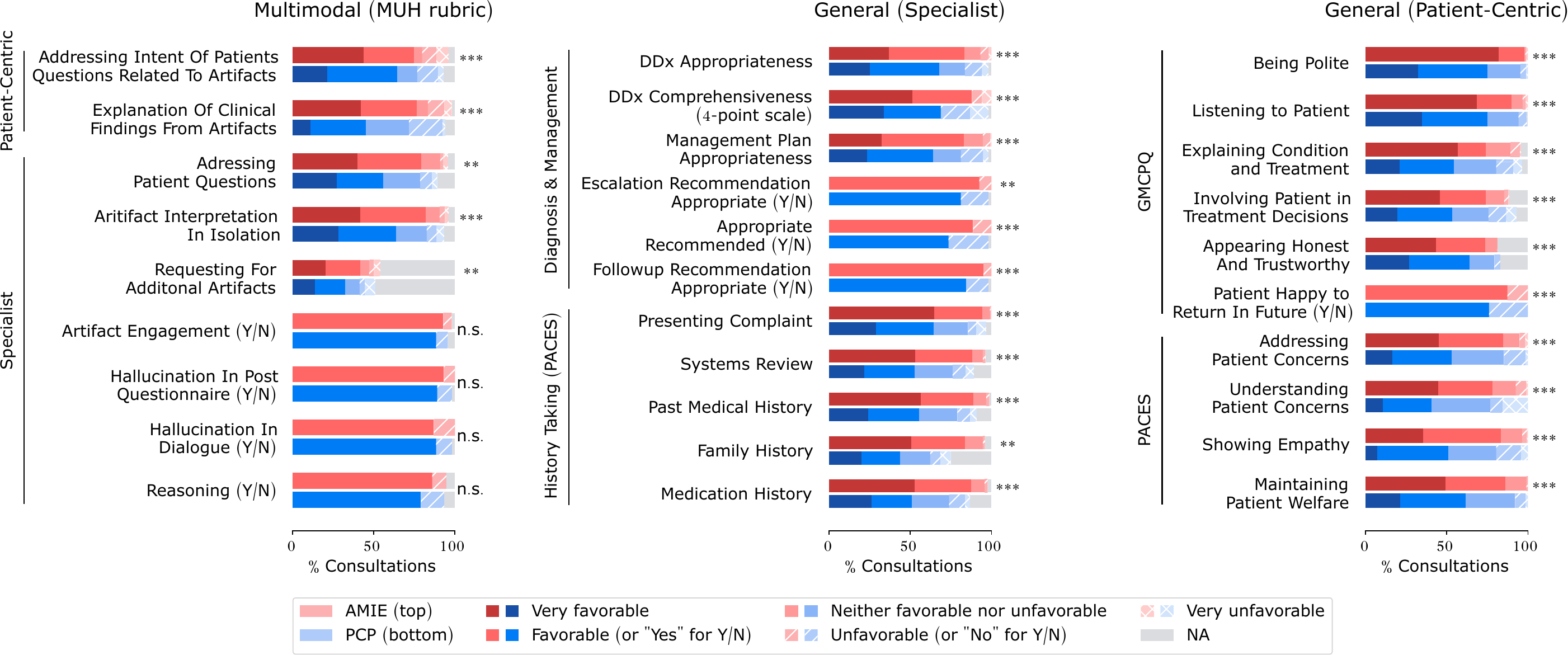}
    \caption{\footnotesize
      \textbf{Comparison of OSCE rating distributions between AMIE and PCPs}. The quality of dialogues were assessed offline by patient actors and specialist physicians, and the distributions of ratings are displayed separately for the sessions with PCPs and AMIE. 
      For illustration purposes, all responses from five-point rating scales were mapped to a generic five-point scale ranging from `Very favorable' to `Very unfavorable'. For Yes/No questions, a (positive) `Yes' response was mapped to the same color as `Favorable' and a (negative) `No' response to the same color as `Unfavorable'.
      Each arm (AMIE or PCP) comes with 105 consultations. 
      \textbf{Left. Multimodal Understanding and Handling.} Various aspects of multimodal capabilities relevant in consultations were assessed according to the MUH criteria by both specialist physicians and patient actors. 
      \textbf{Middle. Specialist metrics.} Expert physicians' ratings on clinically meaningful axes are reported with a focus on attributes related to diagnosis quality, management reasoning and history taking - see Figure~\ref{fig:specialist-eval-all} for the full version. 
      \textbf{Right. Patient-centric metrics} Conversation qualities are assessed by patient actors upon conclusion of the consultation. Only a key subset of the rubric is presented here - see Figure~\ref{fig:osce_patient_centred_metrics_all} for the full version. 
    Asterisks represent statistical significance ($*:p<0.05$, $**:p<0.01$, $***:p<0.001$, $n.s.: $ not significant).
    }
    \label{fig:osce_summary_results_distribution}
\end{figure}

\subsection{Full patient-centered evaluation}
\label{sec:appendix-patient-eval}

\begin{figure}[H]
    \centering
    \includegraphics[width=1\textwidth,height=\textheight,keepaspectratio]{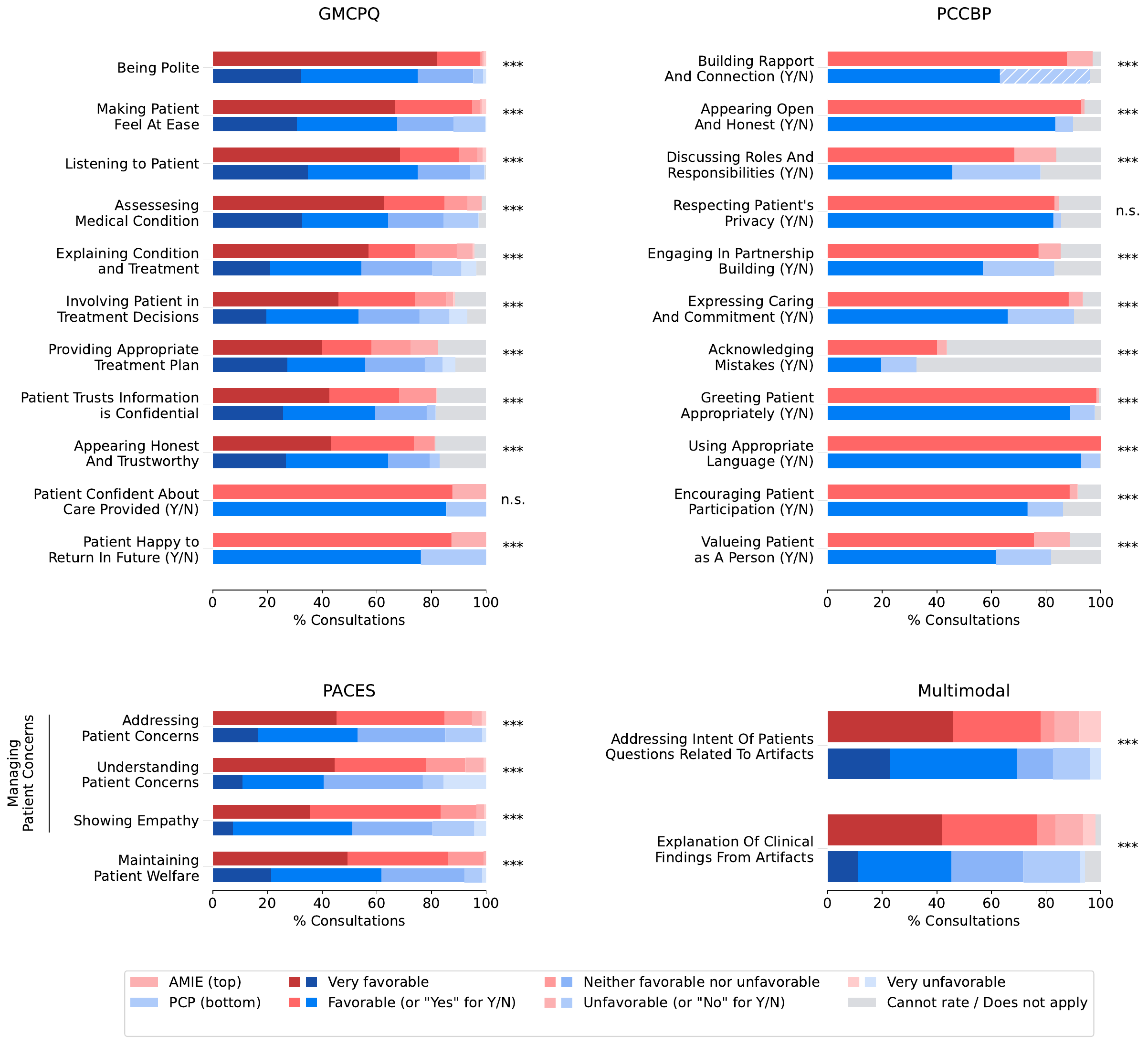}
    \caption{\footnotesize
    \textbf{OSCE patient-centered metrics (full results)}. Conversation qualities as assessed by patient actors upon conclusion of the consultation. For illustration purposes, all responses from five-point rating scales were mapped to a generic five-point scale ranging from `Very favorable' to `Very unfavorable'. For Yes/No questions, a (positive) `Yes' response was mapped to the same color as `Favorable' and a (negative) `No' response to the same color as `Unfavorable'. Rating scales were adapted from the General Medical Council Patient Questionnaire (GMCPQ), the Practical Assessment of Clinical Examination Skills (PACES), and a narrative review about Patient-Centered Communication Best Practice (PCCBP).
    Asterisks represent statistical significance ($*:p<0.05$, $**:p<0.01$, $***:p<0.001$, $n.s.: $ not significant).}
    \label{fig:osce_patient_centred_metrics_all}
\end{figure}

\subsection{Full specialist evaluation}
\label{sec:appendix-specialist-eval}

\begin{figure}[H]
    \centering
    \includegraphics[width=\textwidth, keepaspectratio]{
    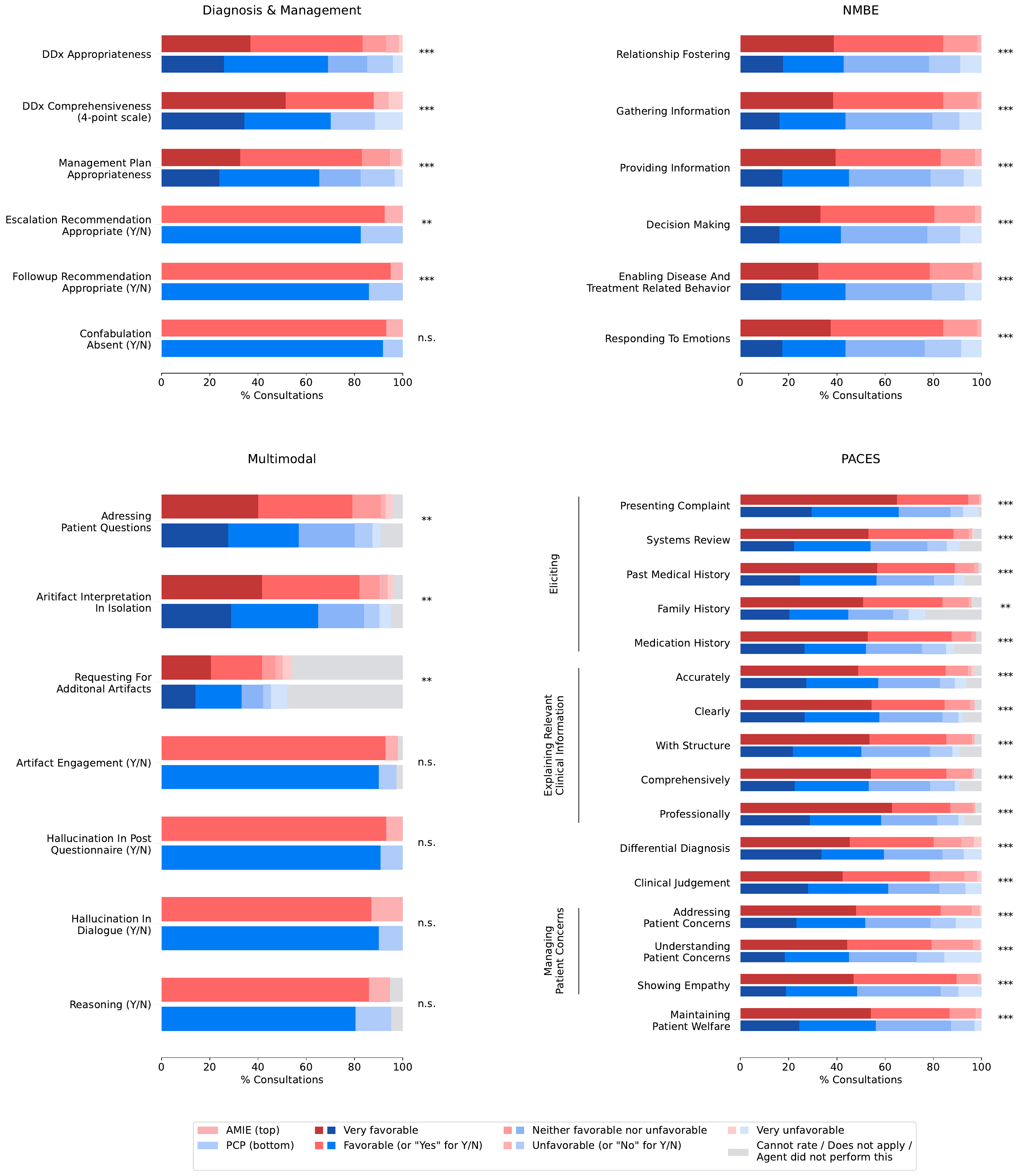}
    \caption{\footnotesize
    \textbf{OSCE specialist evaluation (full results).} Multimodal conversation and reasoning qualities as assessed by specialist physicians. Specialists rated conversations on ordinal scales. For illustration purposes, all responses from five-point rating scales were mapped to a generic five-point scale ranging from `Very favorable' to `Very unfavorable'. The only four-point scale (DDx Comprehensiveness) was mapped to the same scale, ignoring the `Neither favorable nor unfavorable' option. For Yes/No questions, a (positive) `Yes' response was mapped to the same color as `Favorable' and a (negative) `No' response to the same color as `Unfavorable'. Rating scales were adapted from the Practical Assessment of Clinical Examination Skills (PACES), a narrative review about Patient-Centered Communication Best Practice (PCCBP), and other sources. Asterisks represent statistical significance ($*:p<0.05$, $**:p<0.01$, $***:p<0.001$, $n.s.: $ not significant).
    }
    \label{fig:specialist-eval-all}
\end{figure}

\section{Supplementary Automated Evaluation Details and Results} \label{app:autorater_results}

This appendix provides supplementary details and quantitative results from our automated evaluation framework, as discussed in Section \ref{sec:results_ablations}. We first detail the perception tests on medical artifacts (Section \ref{app:perception_tests}). We then describe the automated evaluation framework involving simulated dialogues and an auto-rater (detailed in Section \ref{sec:methods_automated_evals}). This includes the specific auto-rater criteria (Appendix \ref{app:autorater_critera}, Table \ref{table:auto_rating}) and its calibration against human expert judgment (Section \ref{app:auto_rater_calibration}). Using this framework, we present the detailed findings from the ablation study assessing the impact of the state-aware reasoning framework (Section \ref{app:reasoning_ablation}) and the results from the evaluation of system robustness against variations in patient scenario presentation (Section \ref{app:patient_augmentations}). Finally, we compare the performance impact of supervised fine-tuning (Section \ref{app:sft_ablation}).

\subsection{Perception Tests} \label{app:perception_tests}

To gain meaningful insights into the medical artifact understanding capabilities of language models, we evaluate their performance on relevant medical perception benchmark datasets. While the base model for our multimodal AMIE system was selected based on a combination of factors including reasoning, conversational fluency, and efficiency, evaluating performance on these perception tasks is essential to verify its ability to handle the specific medical modalities central to this work. 
This work focuses on benchmark datasets on three key areas: SCIN (skin images), ECG-QA (ECG tracings), PTB-XL (ECG tracings), and ClinicalDoc-QA (clinical documents), which are closely related to the OSCE study scenarios detailed in Section~\ref{sec:osce_study}. We aim to assess the performance of different Gemini model variants~\footnote{\url{https://ai.google.dev/gemini-api/docs/models}}, including Gemini 1.5 Flash, Gemini 1.5 Pro, and Gemini 2.0 Flash, across these selected benchmarks.

\subsubsection{Perception Test Datasets}

\paragraph{Skin Condition Image Network (SCIN)}

\begin{itemize}
    \item \textbf{Dataset:} We utilize the Skin Condition Image Network (SCIN) dataset \cite{ward2024creating}, an open-access resource specifically designed to address limitations in existing dermatology image collections. SCIN uniquely supplements traditional clinical datasets by incorporating images crowdsourced directly from individuals in the US via a consented Google Search contribution pathway. This approach captures a broad spectrum of common dermatological concerns (e.g., rashes, allergies, infections) frequently encountered in online health searches and early disease stages, which are often underrepresented in clinical settings \cite{ward2024creating}. The dataset comprises over 10,000 images from more than 5,000 contributions. A key strength of SCIN is its diversity, featuring images across a wide range of self-reported and dermatologist-estimated Fitzpatrick Skin Types (eFST) and layperson labeler estimated Monk Skin Tones (eMST), contributing to a more balanced representation compared to many existing datasets. Furthermore, SCIN includes rich metadata associated with contributions: self-reported demographics, symptoms, condition history, and dermatologist-provided differential diagnoses with confidence scores.

    \item \textbf{Evaluation:} We evaluated the capability of Gemini models to provide probable diagnoses by incorporating patient context (e.g., demographics, symptoms) and interpreting skin condition images from SCIN dataset. Performance on this challenging, open-ended diagnostic task was evaluated using top-1, top-3, and top-10 accuracy against the ground truth skin conditions. To address the inherent variability in clinical language, including synonyms and spelling variations, a grader LLM was utilized to determine if the model's top-k predicted diagnoses were semantically equivalent to the ground truth.

\end{itemize}

\paragraph{PTB-XL}
\begin{itemize}
    \item \textbf{Dataset:} PTB-XL is a large public dataset containing approximately 22,000 10-second, 12-lead ECG recordings from 18,000 patients with cardiologist-validated annotations adhering to the SCP-ECG and AHA coding standards. Each annotation include multiple diagnostic labels which consist of 4 coarse superclasses and 24 subclasses. The dataset also includes demographic information about the patients, such as age, sex, weight, and height. This allows for more comprehensive analysis and the development of models that can account for these patient contexts. Additionally, ECG waveforms are provided in a standardized digital format that can be easily processed and used with a parser library supported in Python, C, MATLAB, and Java. 
    
    \item \textbf{ECG Image Generation (ECG-Image-Kit):} Standard clinical practice, particularly in settings like Objective Structured Clinical Examinations (OSCEs), often involves interpreting visual ECG printouts. To align our evaluation with this context, we transformed the raw 12-lead ECG time-series signals from the selected ECG-QA subset into static images. This conversion was achieved using ECG-Image-Kit \cite{shivashankara2024ecg}, a library designed to produce realistic ECG visualizations from signal data. We configured the generation process to mimic standard clinical 12-lead ECG layouts, adjusting parameters such as lead arrangement, simulated paper, and background grid display to ensure clinical fidelity. The visual quality and diagnostic faithfulness of the resulting images were verified by a clinical reviewer, ensuring that our multimodal model evaluation assesses interpretation capabilities rather than artifacts introduced during image synthesis.
    
    \item \textbf{Evaluation:} The PTB-XL dataset was used to evaluate the ECG analysis capabilities of Gemini models. To align with clinical practice and the OSCE study format, we formulated an open-ended task requiring the model to provide up to ten probable diagnoses based on patient context and an ECG trace image. Our evaluation setting differs from the standard PTB-XL benchmark task in two key ways that increase its difficulty. Firstly, we used synthetically generated ECG trace images rather than raw ECG signals, simulating scenarios where patients share visual ECG reports. Analyzing ECGs from images, without access to subtle raw signal details, is inherently more challenging. Secondly, asking the model to generate diagnoses without a predefined list of options is more demanding than typical classification or closed-ended multiple-choice tasks. The model's ability to analyze and interpret the ECG image to provide a diagnosis was measured using top-1, top-3, and top-10 accuracy, comparing its responses against PTB-XL's fine-grained diagnostic labels (i.e., diagnostic subclasses).
    
\end{itemize}

\paragraph{ECG-QA}

\begin{itemize}
     \item \textbf{Dataset:} We utilized the ECG-QA benchmark \cite{oh2023ecgqa}, a sensor-text multimodal dataset specifically designed for evaluating question answering performance based on electrocardiogram (ECG) analysis. ECG-QA leverages signals from established clinical ECG databases. The original version is built upon PTB-XL \cite{wagner2020ptb}, a large public dataset containing approximately 22,000 10-second, 12-lead ECG recordings with multi-label, cardiologist-validated annotations adhering to the SCP-ECG standard. It is important to note that PTB-XL annotations within ECG-QA are cardiologist-validated. The benchmark features expert-validated questions covering diverse, clinically relevant ECG interpretation topics, including SCP codes, noise characteristics, infarction stage, heart axis, and quantitative numeric features, and diverse question types such as yes/no, multiple choice, and open-ended questions. Our work focuses on the subset requiring analysis of a single ECG, which provides 159,306 training and 41,093 test samples (question-answer-ECG triplets). Similar to PTB-XL, generated ECG images are used instead of raw ECG waveform to align the evaluation with clinical practice.

    \item \textbf{Evaluation:} We evaluated Gemini models on the ECG-QA benchmark using synthetically generated ECG images and their corresponding questions. As with the PTB-XL evaluation, employing ECG images instead of raw signals increases the task's difficulty due to the absence of raw signal granularity. Performance was measured using a standard question-answering metric such as exact match accuracy.

\end{itemize}

\paragraph{Clinical Document QA (ClinicalDoc-QA)}

\begin{itemize}
    
\item \textbf{Dataset:} To evaluate multimodal AMIE's ability to process and understand visually presented clinical documents, we created a question answering dataset on clinical documents (ClinicalDoc-QA). This is derived from a private dataset consists of ~21k diverse, de-identified clinical notes and patient records sourced from various medical specialties (e.g., Gastroenterology, Internal Medicine, Cardiology). ClinicalDoc-QA focuses specifically on enabling evaluation of medical visual document understanding capabilities.

    \item \textbf{ClinicalDoc-QA Creation Process:} The ClinicalDoc-QA dataset was constructed via the following steps, focusing on generating a targeted evaluation set for visual document QA:
        \begin{enumerate}
            \item \textbf{Sampling and Formatting:} We selected 81 de-identified clinical notes from the de-identified 21k clinical notes and documents collection and represented them in PDF format.
            \item \textbf{Q\&A Generation:} We synthetically generated initial question-answer pairs based on the content of these PDF documents. Questions targeted specific information extraction (e.g., "What is the patient's chief complaint?", "What is the reported Haemoglobin (Hb) result and reference range?").
            \item \textbf{Manual Refinement:} The automatically generated Q\&A pairs were manually reviewed, corrected, and refined by clinical team members to improve clinical accuracy, relevance, and clarity.
            \item \textbf{Data Splitting:} The curated Q\&A pairs were manually partitioned into training and test sets.
            \item \textbf{Multimodal Dataset Preparation:} Subsequent processing prepared the data for model consumption. This involved: (a) Converting each page of the source PDF documents into individual PNG image files. (b) Structuring the curated Q\&A pairs along with references to their corresponding document page images into a format suitable for multimodal model input.
        \end{enumerate}
    This process resulted in a dataset where questions are posed about the content of clinical notes presented as images.

    \item \textbf{Evaluation:} We assessed Gemini models' performance on the ClinicalDoc-QA test set by providing the model with the clinical document images and the corresponding questions. The model's ability to accurately extract information to reason and answer questions based on the visual document content was quantified using exact match accuracy.

\end{itemize}

\subsubsection{Perception Test Results} \label{app:perception_test_results}

We evaluated several Gemini model variants on the aforementioned datasets to quantify their ability to accurately process and interpret medical images. While the selection of the base model for the multimodal AMIE system considered a combination of factors (including general reasoning, conversational fluency, and efficiency), verifying this foundational perceptual capability was a primary objective of these specific tests. The objective was not to achieve state-of-the-art performance on isolated visual question-answering tasks. We also acknowledge that our evaluation setting — characterized by differences in task type, input modality, and dataset splits compared to published baselines — makes it difficult to do direct state-of-the-art comparisons.  Instead, these tests aimed to compare the baseline perceptual performance and tradeoffs of different Gemini model variants, informing one crucial aspect of the multi-faceted selection process for the multimodal AMIE base model.

\begin{figure}[H]
    \centering
    \includegraphics[width=1\textwidth,height=\textheight,keepaspectratio]{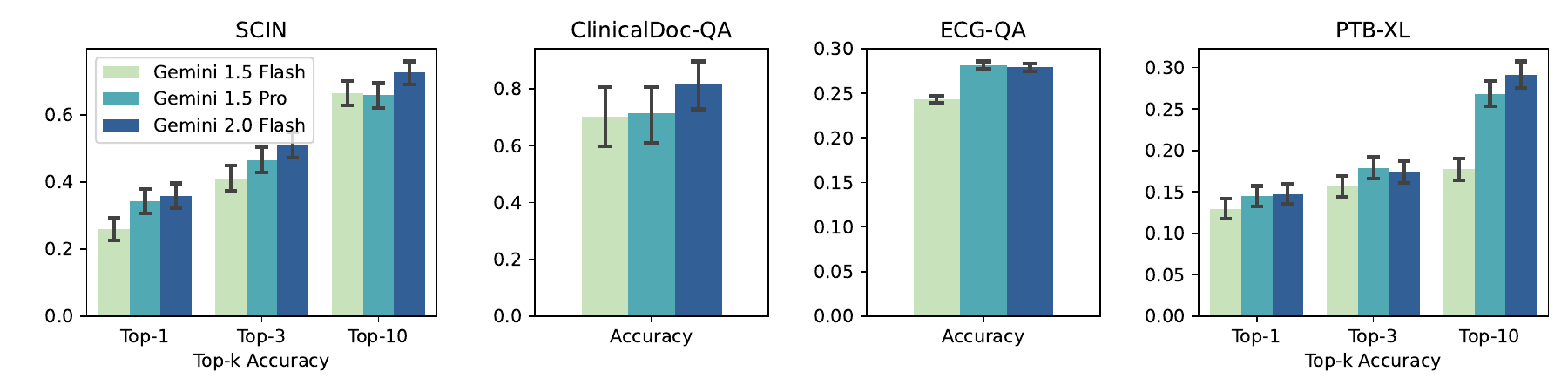}
    \caption{\footnotesize \textbf{Perception test results}. We assess foundational visual understanding across diverse medical data types: SCIN (Dermatology images), PTB-XL (Electrocardiogram tracings), ECG-QA (Electrocardiogram Question Answering), and ClinicalDoc-QA (Clinical Document Question Answering). We evaluate top-k accuracy (top-1, top-3, top-10) for PTB-XL and SCIN, reflecting diagnostic classification ability. For ECG-QA and ClinicalDoc-QA, we report exact match accuracy, indicating the model's correctness on question answering. Three multimodal LLMs are evaluated: Gemini 1.5 Flash, Gemini 1.5 Pro, and Gemini 2.0 Flash. Results indicate that Gemini 2.0 Flash generally exhibits robust perceptual capabilities.}
    \label{fig:perception_test}
\end{figure}

The results of this evaluation are illustrated in Figure~\ref{fig:perception_test}, detailing the top-k accuracy for SCIN and PTB-XL and the exact match accuracy for ClinicalDoc-QA and ECG-QA. We evaluated Gemini 1.5 Flash, Gemini 1.5 Pro, and Gemini 2.0 Flash, with Flash representing a smaller model compared to Pro. Our results show that Gemini 1.5 Pro, a larger model, outperforms Gemini 1.5 Flash across all datasets, highlighting the benefits of increased model capacity. However, notably, Gemini 2.0 Flash, despite its smaller size, achieves performance comparable to or exceeding that of Gemini 1.5 Pro across all benchmarks. We believe Gemini 2.0 Flash's performance is from architectural advancements and extensive pre-training. Considering its efficient inference speed, promising perception test results, and its established strengths in conversational fluency and reasoning, Gemini 2.0 Flash was determined to be the most suitable selection as the base model for the multimodal AMIE system.

\subsection{Automated Dialogue Evaluation: Criteria and Scoring Rubric} \label{app:autorater_critera}

This appendix details the specific criteria and scoring rubric used by the auto-rater (step (3) in Section~\ref{sec:simulation_method}, Figure~\ref{fig:simulation}) for the automated evaluation of simulated dialogues. Table~\ref{table:auto_rating} outlines each criterion—including diagnostic accuracy, information gathering, management appropriateness, and hallucination detection—along with its description and scoring scale. Results obtained using this rubric are presented in Section~\ref{sec:results_automated_evals} and Appendix~\ref{app:autorater_results}.

\begin{table}[H]
\centering
\resizebox{\textwidth}{!}{
\begin{tabular}{clc}
\toprule
\footnotesize
\textbf{Criterion} & \multicolumn{1}{c}{\textbf{Description}} & \textbf{Scoring} \\
\midrule
Top-1 accuracy & \begin{tabular}[c]{@{}l@{}}Presence of the correct diagnosis in the first position of\\ the differential diagnosis list given in the post-questionnaire.\\ Considers the patient's condition given in the patient profile,\\ and take into account synonyms for diagnoses, not only \\ exact matches.\end{tabular} & Binary \\
\midrule
Top-3 accuracy & \begin{tabular}[c]{@{}l@{}}Similar to top-1 accuracy but checks the presence of the \\ correct diagnosis within the first-3 positions of the differential \\ diagnosis list.\end{tabular} & Binary \\
\midrule
Top-10 accuracy & \begin{tabular}[c]{@{}l@{}}Similar to top-1 accuracy but checks the presence of the \\ correct diagnosis within the first-10 positions of the differential\\ diagnosis list.\end{tabular} & Binary \\
\midrule
Gathering information & \begin{tabular}[c]{@{}l@{}}Assessment of the doctor's information-gathering skills, including\\ open-ended questions, active listening, and eliciting patient perspectives.\\ Considers the model attempt to understand the patient's needs for the \\ encounter, elicits full description of major reason for visit from biologic \\ and psychosocial perspectives, checks if the model asks open-ended \\ questions, and allows patient to complete responses and listen actively. \\ Also elicits patient's full set of concerns, patient's perspective on the \\ problem/illness, explores full effect of the illness, and clarifies and \\ summarize information. Enquires about additional concerns if needed.\end{tabular} & 5-point Likert \\
\midrule
\begin{tabular}[c]{@{}c@{}}Management plan \\ appropriateness\end{tabular} & \begin{tabular}[c]{@{}l@{}}Evaluation of the appropriateness of the doctor's management plan, \\ considering best practices and patient safety, including recommending\\ emergency or red-flag presentations to go to the emergency department.\end{tabular} & 5-point Likert \\
\midrule
Hallucination & \begin{tabular}[c]{@{}l@{}}Detection of fabricated information by the doctor agent. Specifically, \\ we consider the following patterns:\\ (1) Mentions information not provided or experienced by the patient.\\ (2) Makes mistaken assumptions or recalls information incorrectly.\\ (3) Claims access to patient databases or services they shouldn't have.\\ (4) Mentions medications that don't exist.\end{tabular} & Binary \\
\bottomrule
\end{tabular}
}
\caption{\footnotesize \textbf{Automated dialogue evaluation: criteria and scoring rubric.} This table details the specific criteria, descriptions, and scoring methods (Binary or 5-point Likert) used by the auto-rater for evaluating simulated diagnostic dialogues.}
\label{table:auto_rating}
\end{table}

\subsection{Verifying Calibration of Auto-Rater Scores} \label{app:auto_rater_calibration}

While the auto-rater described in Section \ref{app:autorater_critera} enables rapid and scalable evaluation of simulated dialogues, crucial for iterative development and ablation studies, it is essential to verify that its assessments align with human expert judgment. Absolute scores from automated systems may not always be directly comparable to human ratings. Therefore, we focus on calibration: ensuring that the relative ranking of dialogues provided by the auto-rater on key qualitative dimensions (like information gathering or management appropriateness) corresponds to the ranking that would be assigned by expert clinicians. Establishing this rank-order agreement provides confidence that improvements detected by the auto-rater reflect meaningful enhancements in dialogue quality when comparing different models or system configurations.

\begin{figure}[H]
    \centering
    \includegraphics[width=0.95\textwidth,height=\textheight,keepaspectratio]{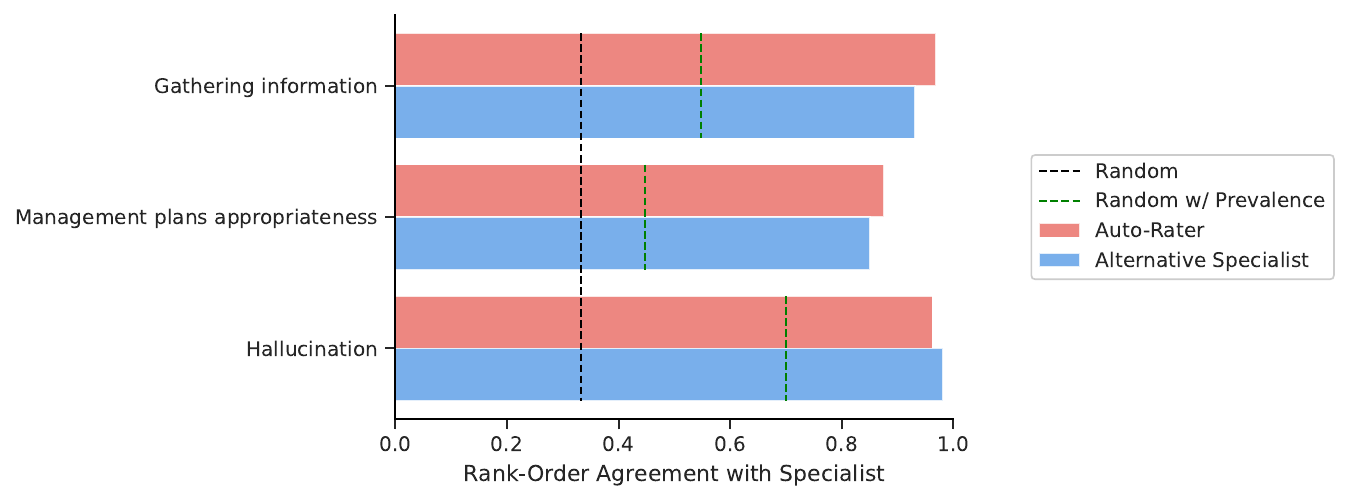}
    \caption{\footnotesize \textbf{Calibration analysis of auto-rater scores}. This figure assesses the calibration of the auto-rater used for evaluating simulated dialogues (Table \ref{table:auto_rating}) by comparing its rank-order agreement with human specialists. Using data from a previous text-only OSCE study~\citep{tu2024towards}, we measure the agreement of the auto-rater (``Auto-Evaluation'') with an ``anchor'' specialist's ranking against the baseline agreement between an ``alternative'' specialist and the anchor specialist ('Alternative Specialist') for three key metrics: Gathering Information, Management Plan Appropriateness, and Hallucination. Dashed lines represent random chance and prevalence-weighted random chance baselines. The results demonstrate that the auto-rater achieves rank-order agreement comparable to inter-specialist agreement and significantly above chance, indicating good calibration for assessing relative dialogue quality on these dimensions.}
    \label{fig:autorater_calibration}
\end{figure}

To perform this calibration, we leveraged data from a previous large-scale, text-only OSCE study comparing AMIE and PCPs~\citep{tu2024towards}, where dialogues were rated by multiple specialist physicians on analogous criteria. For the qualitative metrics assessed by our current auto-rater (Gathering Information, Management Plan Appropriateness, and Hallucination detection), we evaluated the rank-order agreement. For each dialogue assessed by multiple specialists, we randomly selected an ``anchor'' specialist rating and an ``alternative'' specialist rating. We then measured the agreement between the auto-rater's score and the anchor specialist's ranking, comparing it against the baseline agreement between the alternative specialist and the anchor specialist. Figure \ref{fig:autorater_calibration} illustrates this analysis, including baselines for random chance and prevalence-weighted random chance (dashed lines). The results show that the auto-rater achieves rank-order agreement comparable to that of an alternative human specialist, indicating it is well-calibrated for assessing relative performance on these key dialogue quality dimensions.

\subsection{Improvement with State-Aware Reasoning} \label{app:reasoning_ablation}

This section provides the detailed results from the ablation study designed to quantify the contribution of the state-aware reasoning framework (described in Section~\ref{sec:reasoning_method}) to multimodal AMIE's performance. We compared the full system, utilizing the state-aware dialogue phase transitions and uncertainty-driven questioning, against a `Vanilla' baseline. This baseline used the same Gemini 2.0 base model but lacked the explicit reasoning framework, relying solely on domain-specific instructions. The comparison was conducted using our automated evaluation pipeline (Section~\ref{sec:simulation_method}) involving simulated dialogues across four distinct medical domains/datasets: PAD-UFES-20 (Dermatology), SCIN (Dermatology), Clinical Documents, and PTB-XL (Electrocardiography).

Table~\ref{tab:reasoning_ablation} presents the quantitative results from the auto-rater, comparing the two inference methods (``Vanilla'' vs. AMIE with reasoning) across key performance metrics. As the results demonstrate, incorporating the state-aware reasoning framework consistently enhances performance across datasets, notably improving diagnostic accuracy and information gathering scores while maintaining negligible hallucination rates, indicating the value of this structured approach.

\begin{table}[H]
\footnotesize
\centering
\caption{\footnotesize \textbf{Auto-rater results for state-aware reasoning at inference}. We report performance across four datasets representing different medical domains: PAD-UFES-20 and SCIN (Dermatology), PTB-XL (Electrocardiography), and Clinical Documents. Metrics evaluated include Differential Diagnosis Accuracy (at Top-1, Top-3, and Top-10), Information Gathering, Hallucination Rate, and Appropriateness of Management Plan (Mx Plan). Two inference methods are compared: Vanilla (Gemini 2.0 out-of-the-box with domain-specific instructions) and AMIE (Gemini 2.0 with domain-specific state-aware reasoning at inference).}
\label{tab:reasoning_ablation}
\begin{tabular}{@{}c|l|p{1.4cm}p{1.4cm}p{1.4cm}|p{1.8cm}p{1.8cm}p{1.5cm}@{}}
\toprule
\multicolumn{1}{c|}{\textbf{Inference}} & \textbf{Dataset} & \multicolumn{3}{c|}{\textbf{Differential Diagnosis Accuracy}} & \textbf{Information Gathering} & \textbf{Hallucination Rate} & \textbf{Mx Plan} \\
\cmidrule(l){3-5}
 &  & Top-1 & Top-3 & Top-10 &  &  &  \\ \midrule
Vanilla & PAD-UFES-20 (Derm) & 0.75 & 0.79 & 0.86 & 4.00 & 0.00 & 4.14 \\
AMIE & PAD-UFES-20 (Derm) & 0.84 & 0.85 & 0.90 & 4.05 & 0.00 & 4.18 \\
Vanilla & SCIN (Derm) & 0.49 & 0.54 & 0.71 & 3.99 & 0.01 & 3.60 \\
AMIE & SCIN (Derm) & 0.54 & 0.61 & 0.74 & 4.02 & 0.01 & 3.89 \\
Vanilla & Clinical Docs & 0.89 & 0.93 & 0.96 & 4.00 & 0.00 & 4.48 \\
AMIE & Clinical Docs & 0.98 & 0.98 & 0.98 & 4.17 & 0.04 & 4.57 \\
Vanilla & PTB-XL (ECG) & 0.20 & 0.32 & 0.46 & 4.00 & 0.00 & 3.60 \\
AMIE & PTB-XL (ECG) & 0.28 & 0.44 & 0.61 & 4.03 & 0.01 & 3.64 \\ \bottomrule
\end{tabular}
\end{table}

\subsection{Robustness to Patient Scenario Augmentations} \label{app:patient_augmentations}

To assess the robustness of multimodal AMIE's diagnostic conversations against realistic variations in patient presentation that do not alter the core clinical facts, we performed an analysis using augmented patient scenarios. This evaluation helps gauge the system's stability and potential for reliable performance across diverse interaction styles and patient backgrounds.

We employed LLM-driven augmentation techniques to modify the original synthetic patient scenarios along three distinct axes:

\begin{enumerate}
    \item \textbf{Personality variations:} We simulated variations in patient personality styles based on the "Big 5" personality traits model~\citep{roccas2002big} – Openness, Conscientiousness, Extraversion, Agreeableness, and Neuroticism. LLM prompts were used to instruct the patient agent simulator to adopt characteristics associated with random low/high combinations of these traits (e.g., more talkative/extraverted, more anxious/neurotic, less agreeable). These variations primarily affect the style, tone, and potentially the amount or order of information shared during the dialogue, allowing us to test AMIE's ability to manage different interaction dynamics. The core medical facts of the scenario remained unchanged.

    \item \textbf{Demographic variations:} LLM prompts were utilized to perturb demographic details within the patient scenarios, such as sex, gender, and race. These modifications were carefully filtered using another LLM to ensure the resulting demographic profile remained logically and clinically consistent with the underlying medical condition presented in the scenario. This axis of augmentation is crucial for evaluating the fairness of the system and ensuring its performance is not negatively impacted by variations in patient demographics, guarding against potential biases.

    \item \textbf{Semantic background/symptom variations:} We introduced minor semantic changes to the patient's background medical history or the description of their symptoms. This was achieved using LLM prompts to add plausible but non-critical details, such as slightly different wording for symptoms, minor related social or family history elements, or variations in the timeline description. Again, an LLM filter verified that these semantic perturbations maintained logical and clinical consistency with the scenario's primary condition. This tests AMIE's resilience to the natural variability and occasional minor inconsistencies found in how patients recall and report their medical information.
\end{enumerate}

\paragraph{Evaluation and Results}

Dialogues generated using these augmented scenarios were evaluated using the same auto-rater framework and metrics described in Appendix~\ref{app:autorater_critera}: Differential Diagnosis Accuracy (Top-1, Top-3, Top-10), Information Gathering score, Hallucination Rate, and Management Plan (Mx Plan) appropriateness score.

\begin{figure}[H]
    \centering
    \includegraphics[width=0.9\textwidth,height=\textheight,keepaspectratio]{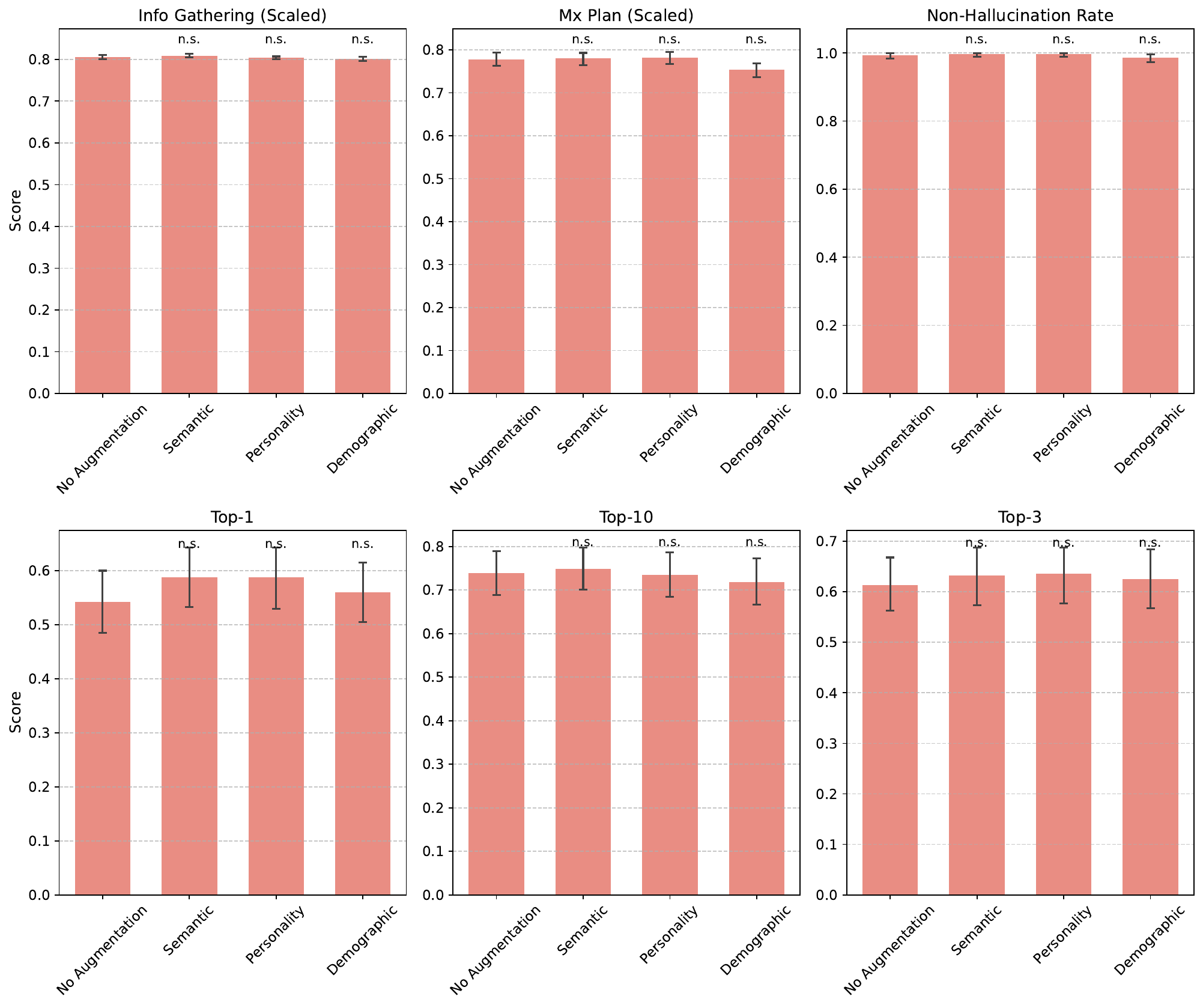}
    \caption{\footnotesize \textbf{Robustness evaluation of AMIE to patient scenario augmentations}. Auto-rater performance of AMIE compared between original synthetic patient scenarios and scenarios with augmentations. Augmentations tested AMIE's robustness across Demographic, Personality, and Semantic axes, while maintaining clinical conclusions.  Error bars represent 95\% confidence intervals (bootstrapped). Significance markers denote p-values obtained from a two-sided Mann-Whitney U test comparing each augmentation condition (Semantic, Personality, Demographic) against the baseline ('No Augmentation') within each metric. Asterisks represent statistical significance ($*:p<0.05$, $**:p<0.01$, $***:p<0.001$, $n.s.: $ not significant). Consistent performance across augmentations demonstrates AMIE's robustness to non-clinically significant variations in patient presentations.}
    \label{fig:ablation_robustness}
\end{figure}

Figure~\ref{fig:ablation_robustness} presents a comparison of AMIE's performance on the original synthetic scenarios versus the scenarios augmented along these three axes. The bar plots show the average scores for each metric across the different scenario types. As illustrated in the figure, AMIE's performance remains highly consistent across all evaluated metrics when interacting with the augmented scenarios compared to the original ones. There are no significant drops in diagnostic accuracy, information gathering effectiveness, or management plan quality, and hallucination rates remain negligible.

The stability of performance observed across these diverse, LLM-driven augmentations, indicates that multimodal AMIE is robust to non-clinically significant variations in patient personality, demographics, and background/symptom descriptions. This suggests a degree of resilience relevant for potential real-world application where patient presentations naturally vary.

\subsection{Impact of Supervised Fine-Tuning on Diagnostic Accuracy and Conversation Quality}
\label{app:sft_ablation}

In line with prior work~\cite{saab2024capabilities} highlighting the value of supervised fine-tuning (SFT) for medical tasks, we investigated its potential to enhance the multimodal conversational abilities of the Gemini 2.0 Flash model. This involved fine-tuning the model to predict clinician turns within dialogue histories containing both text and image data (specifically, images of skin lesions, ECGs, and clinical documents), where dialogues are synthetically generated as described in Section~\ref{sec:methods_automated_evals} (Steps 1 and 2). The SFT training data also include a mix of modality-specific datasets – SCIN for analyzing dermatological images, ECG-QA for interpreting electrocardiograms, and ClinicalDoc-QA for reasoning over clinical documents – all formatted for instruction-based or conversational question answering.

\begin{figure}[H]
    \centering
    \includegraphics[width=1\textwidth,height=\textheight,keepaspectratio]{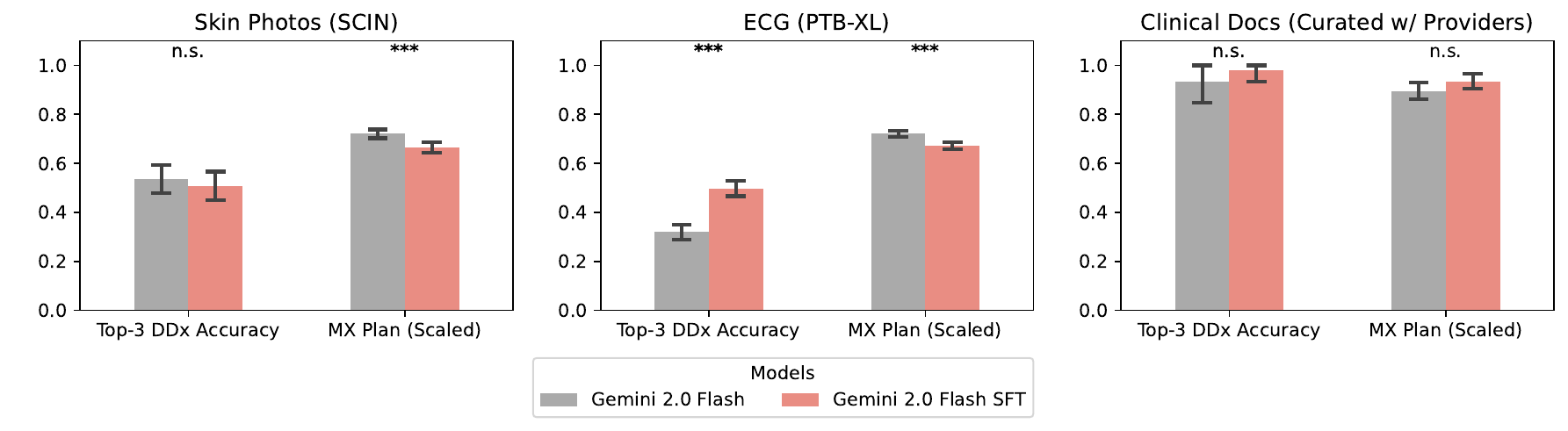}
    \caption{\footnotesize \textbf{Gemini 2.0 vs. SFT performance on top-3 diagnostic accuracy and management plan appropriateness.} This figure compares Gemini 2.0 and a supervised fine-tuned version (SFT) on Top-3 differential diagnosis (DDx) accuracy and Mx plan appropriateness across SCIN (dermatology), PTB-XL (ECG), and Clinical Documents tasks using synthetic dialogues. Fine-tuning notably enhances performance on the PTB-XL ECG task for both Top-3 accuracy and plan appropriateness. SFT also yields improvements, particularly in plan appropriateness, on the Clinical Documents task, while showing modest gains on SCIN. Evaluation uses agent-based simulations, distinct from human-actor evaluations. Error bars represent 95\% confidence intervals (bootstrapped). Significance markers denote p-values obtained from a two-sided Mann-Whitney U test comparing the two conditions within each panel/metric. Asterisks represent statistical significance ($*:p<0.05$, $**:p<0.01$, $***:p<0.001$, $n.s.: $ not significant).}
    \label{fig:sft_synthetic_ddx}
\end{figure}

To determine the effectiveness of SFT, we compared the fine-tuned Gemini 2.0 Flash model with the original base model on a set of synthetic conversations reserved for auto-rating (Section~\ref{sec:methods_automated_evals}). Figure \ref{fig:sft_synthetic_ddx} displays the top-3 diagnostic accuracy and the appropriateness of the management plan, which are key metrics used in our auto-rating. Our evaluation showed that SFT led to statistically significant improvements in diagnostic accuracy for ECGs, but did not provide similar benefits for skin images or clinical documents. Interestingly, we observed a decrease in the quality of the generated conversations, specifically in terms of management plan appropriateness for skin images and ECGs. We believe this degradation may be attributed to the fine-tuning process on a relatively small dataset, potentially impacting the model's core language capabilities and instruction-following abilities. Given that SFT did not offer a substantial advantage and considering the benefits of utilizing a readily available off-the-shelf base model, we decided to proceed with the base Gemini 2.0 Flash model for the AMIE system's base model. This approach allows us to leverage the multimodal capabilities of the latest large language models and maintain the adaptability to enhance the model with inference-time reasoning techniques like our state-aware reasoning (Section~\ref{sec:reasoning_method}).

\clearpage
\section{Qualitative Results from the Multimodal OSCE} \label{sec:appendix-qualitative}

To complement the quantitative results presented in the main paper, this appendix provides qualitative examples drawn from the multimodal OSCE study. These examples offer a closer look at AMIE's conversational and reasoning processes during simulated consultations, particularly highlighting its ability to integrate multimodal information.

\subsection{Examination of Multimodal State-Aware Reasoning}

To demonstrate AMIE's multimodal state-aware reasoning capabilities during the OSCE, we highlight illustrative examples that juxtapose the conversation (left panel) with AMIE's internal reasoning state (right panel, hidden from the user). These examples specifically demonstrate how AMIE's internal state identifies knowledge gaps, prompts targeted questions to address these uncertainties, processes received multimodal artifacts (such as patient-provided images or documents) to fill these gaps, and subsequently updates or confirms its differential diagnosis based on the new information. This iterative process may also reveal new knowledge gaps, which AMIE then assesses for further inquiry within the consultation flow.

\begin{figure}[!htbp]
    \caption{\footnotesize
        \textbf{Example illustrating AMIE's multimodal state-aware reasoning with skin pictures. } The left panel shows the user-facing dialogue, while the right panel shows AMIE's internal state, including identified knowledge gaps and updates after receiving multimodal information (images of a rash in this case). Patient names are fictitious.
    }
    \label{fig:reasoning_derm_320}
    \centering
    \includegraphics[width=\textwidth,keepaspectratio]{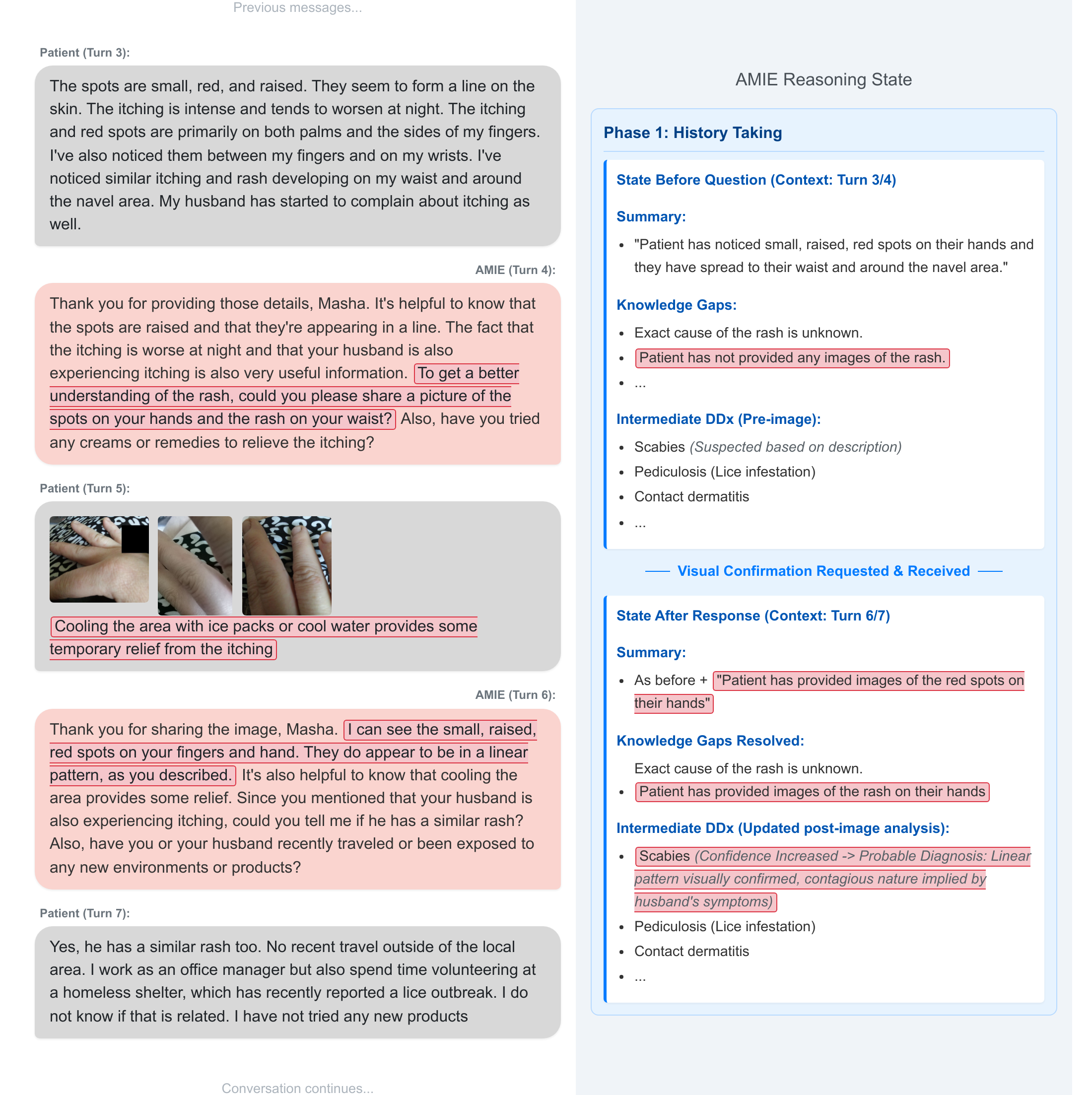}
\end{figure}

For instance, the example in \Cref{fig:reasoning_derm_320} shows this process during Phase 1: History Taking. The intermediate Agent state (right panel) reveals "Knowledge Gaps" identified by AMIE, in this case, the need for visual confirmation of the patient's rash. Next, AMIE requests images from the patient (Turn 4). Upon receiving the images (Turn 5), AMIE processes the visual information and resolves the knowledge gap (shown in the updated State After Response (Context: Turn 6/7)), increasing confidence in the intermediate differential diagnosis (Scabies). Having addressed the visual gap, AMIE proceeds with targeted history taking to assess other knowledge gaps, including deeper investigation into the contagious nature implied by the husband's symptoms (Turn 6).

\begin{figure}[!htbp]
    \caption{\footnotesize
        \textbf{Example illustrating AMIE's multimodal state-aware reasoning with clinical documents.} The left panel shows the user-facing dialogue, while the right panel shows AMIE's internal state. Highlights AMIE identifying a knowledge gap regarding specific values (Thyroid function tests). Upon the patient's request, AMIE processes the document, and updates the differential diagnosis based on information from the clinical document. Patient names are fictitious.
    }
    \label{fig:reasoning_docs_108}
    \centering
    \includegraphics[width=\textwidth,keepaspectratio]{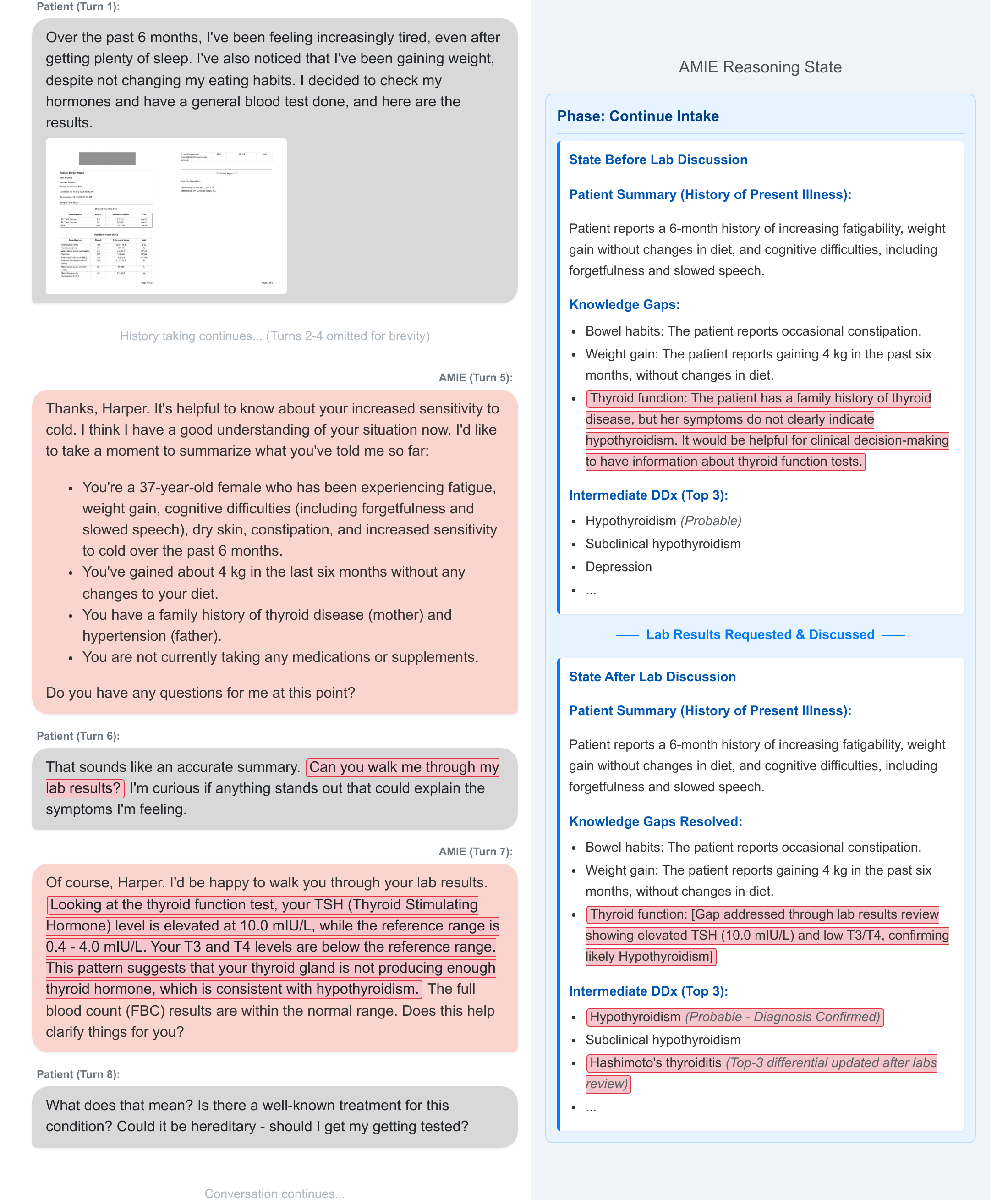}
\end{figure}

Similarly, the example in \Cref{fig:reasoning_docs_108} involves a clinical document. The initial Agent state reveals a specific "Knowledge Gap" identified by AMIE: the need for thyroid function test results to clarify the diagnosis, despite symptoms and family history suggesting potential hypothyroidism. Although the patient provided a link to the results earlier (Turn 1), the key interaction occurs when the patient explicitly requests interpretation (Turn 6). AMIE then processes the information derived from the clinical document, discussing the specific lab values (TSH, T3, T4) with the patient (Turn 7). This action directly resolves the previously identified knowledge gap, as shown in the updated internal state. Consequently, AMIE's confidence in the leading diagnosis increases (Hypothyroidism), and the top-3 differential diagnosis list is updated based on processing this critical information from the clinical document.

Following this focused examination of the reasoning mechanism, we provide examples of full dialogues from the Multimodal OSCE in the subsequent section without internal reasoning states.

\subsection{Example Dialogue Transcripts}
\label{sec:appendix-qualitative-examples}
\begin{figure}[H]
    \centering
    \caption{\footnotesize \textbf{Example of AMIE-patient dialogue for a dermatology scenario}}
    \label{fig:amie_derm_dialogue_1_page1}
    \includegraphics[trim= 2.5cm 2.5cm 2.5cm 3.5cm, width=1\textwidth, height=\textheight,clip,keepaspectratio]{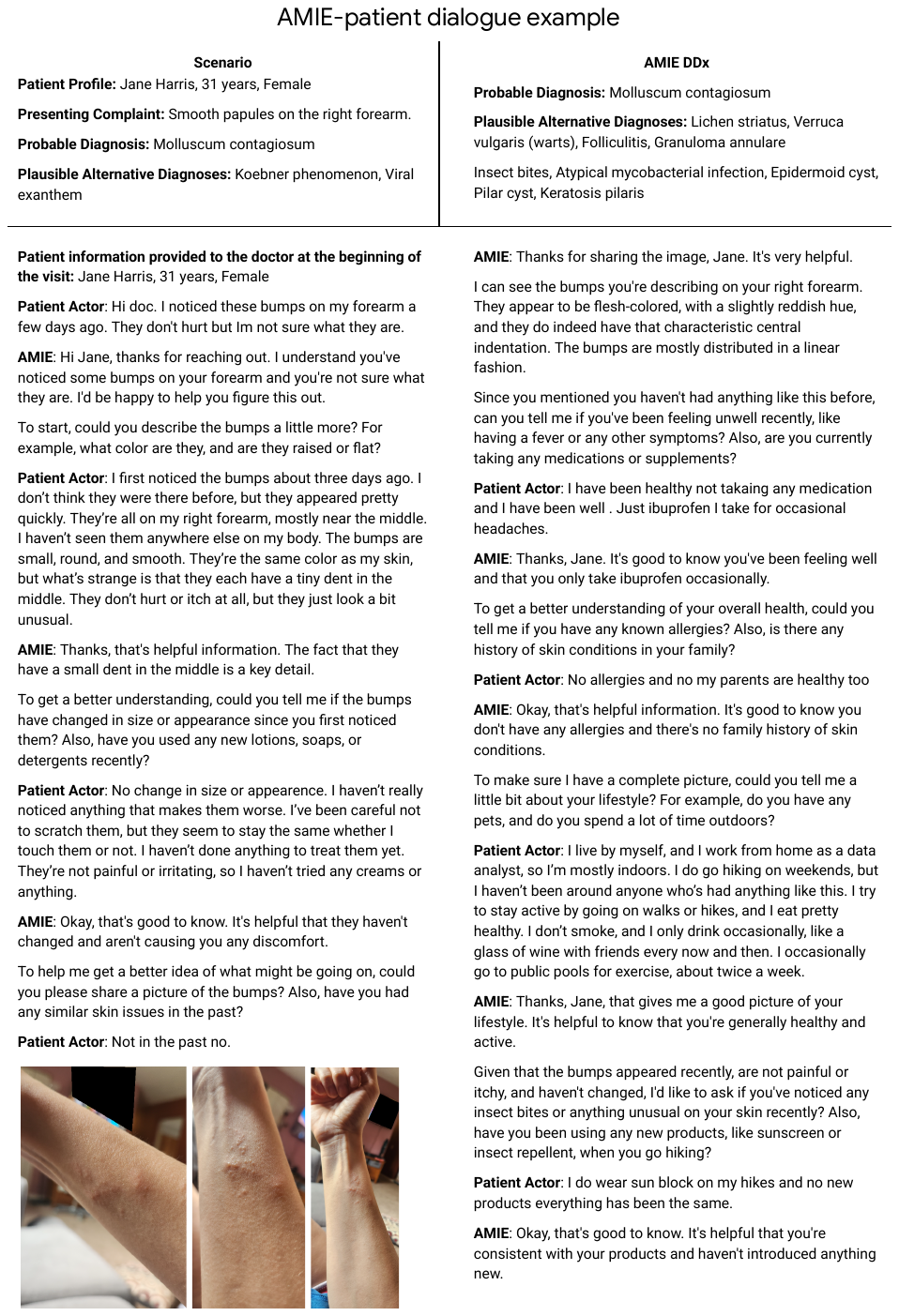}
    \label{fig:dialogue_1}
\end{figure}

\begin{figure}[H]
    \centering
    \caption{\footnotesize \textbf{Example of PCP-patient dialogue for a dermatology scenario}}
    \label{fig:pcp_derm_dialogue_1_page_1}
    \includegraphics[trim= 2.5cm 2.5cm 2.5cm 3.5cm, width=1\textwidth, height=\textheight,clip,keepaspectratio]{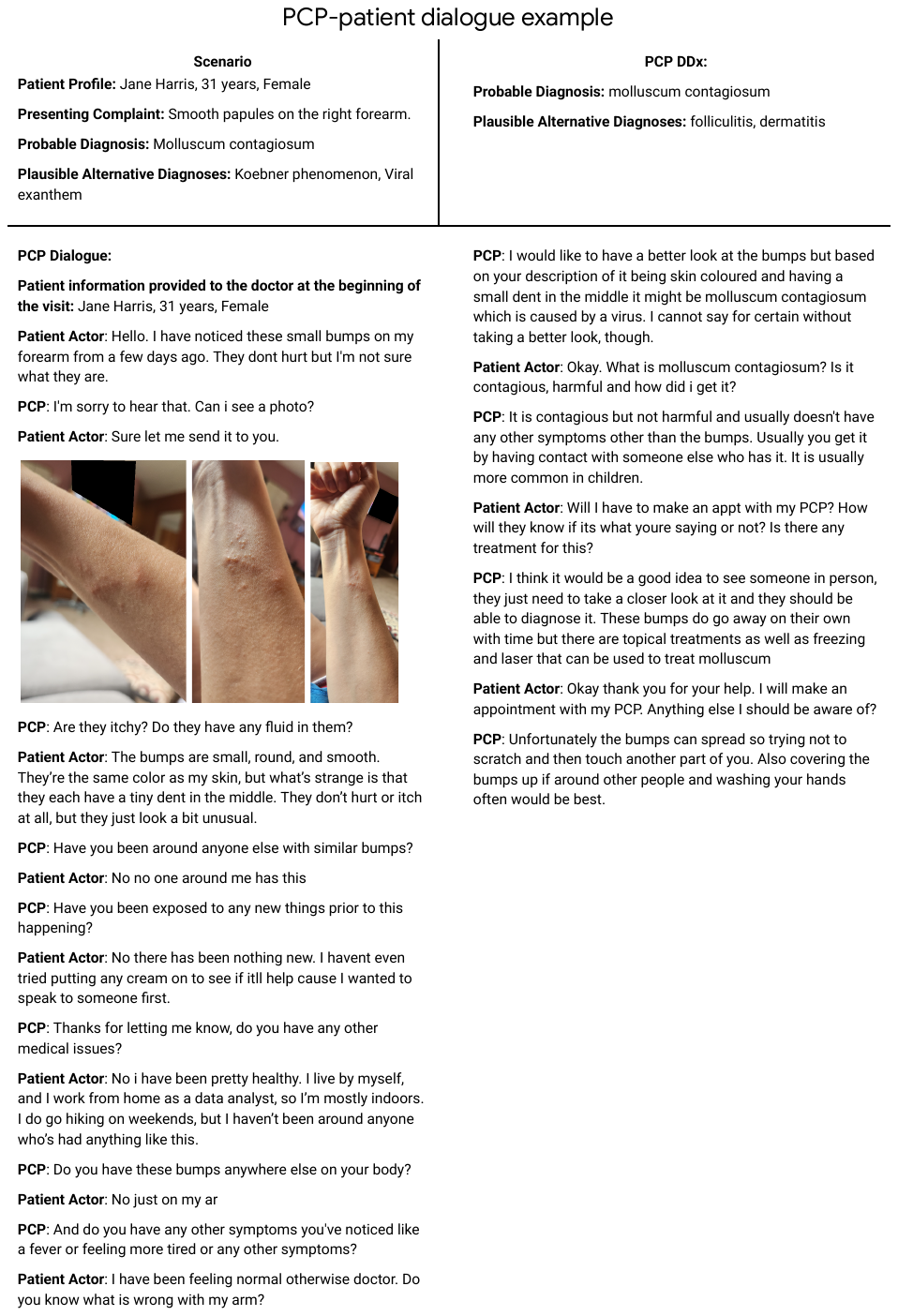}
    \label{fig:dialogue_2}
\end{figure}

\begin{figure}[H]
    \centering
    \caption{\footnotesize \textbf{Example of AMIE-patient dialogue for a clinical document scenario.} Note that the patient data, such as the patient name in the clinical document in this example is not real patient data but was synthesised for the purposes of our study.}
    \label{fig:AMIE_clinical_docs_dialogue_page_1}
    \includegraphics[trim= 2.5cm 2.5cm 2.5cm 3.5cm, width=1\textwidth, height=\textheight,clip,keepaspectratio]{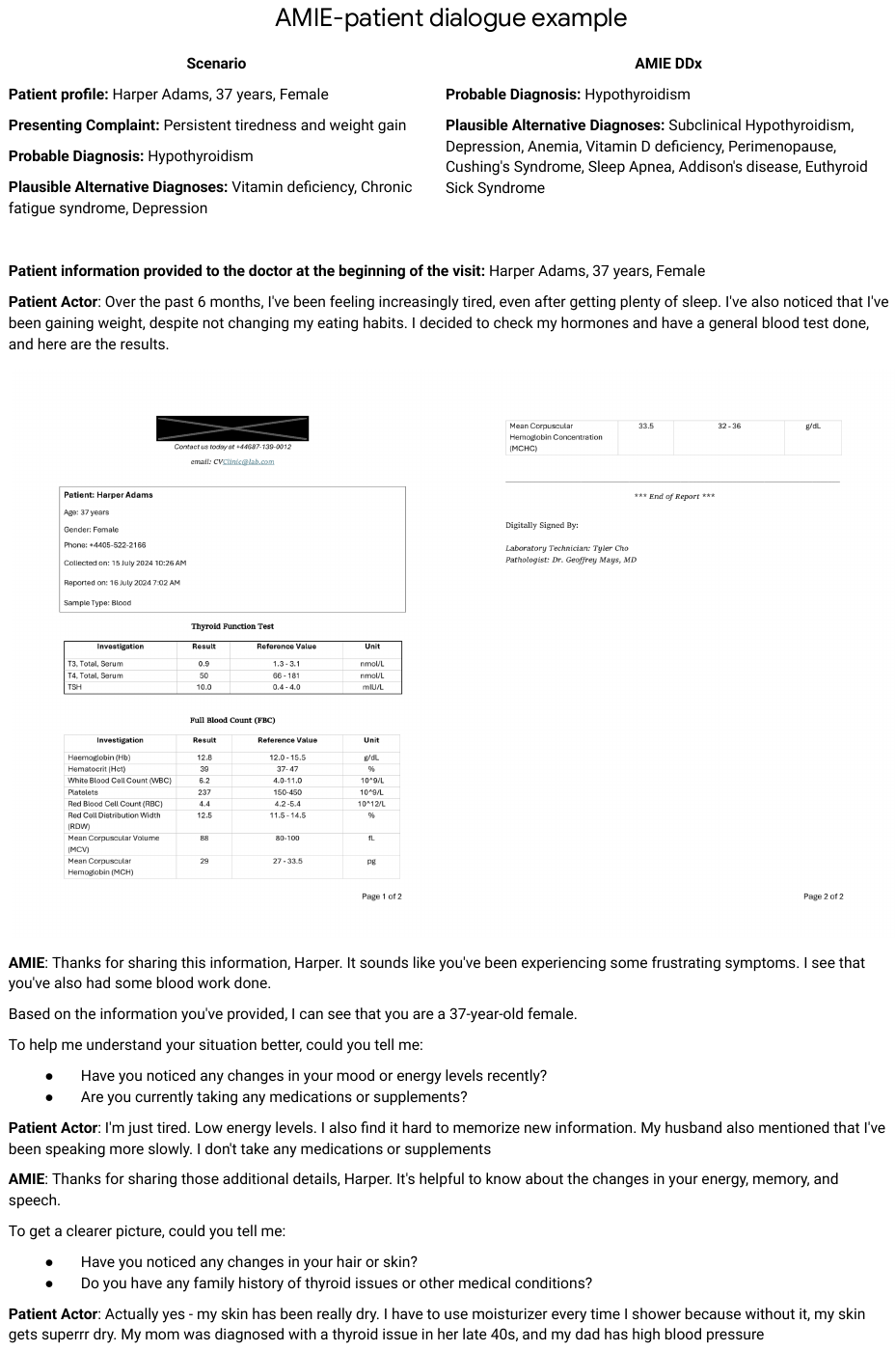}
    \label{fig:dialogue_3}
\end{figure}

\begin{figure}[H]
    \centering
    \caption{\footnotesize \textbf{ Example of PCP-patient dialogue for a clinical document scenario.} Note that the patient data, such as the patient name in the clinical document in this example is not real patient data but was synthesised for the purposes of our study.}
    \label{fig:PCP_clinical_docs_dialogue}
    \includegraphics[trim= 2.5cm 2.5cm 2.5cm 3.5cm, width=0.98\textwidth, height=0.98\textheight,clip,keepaspectratio]{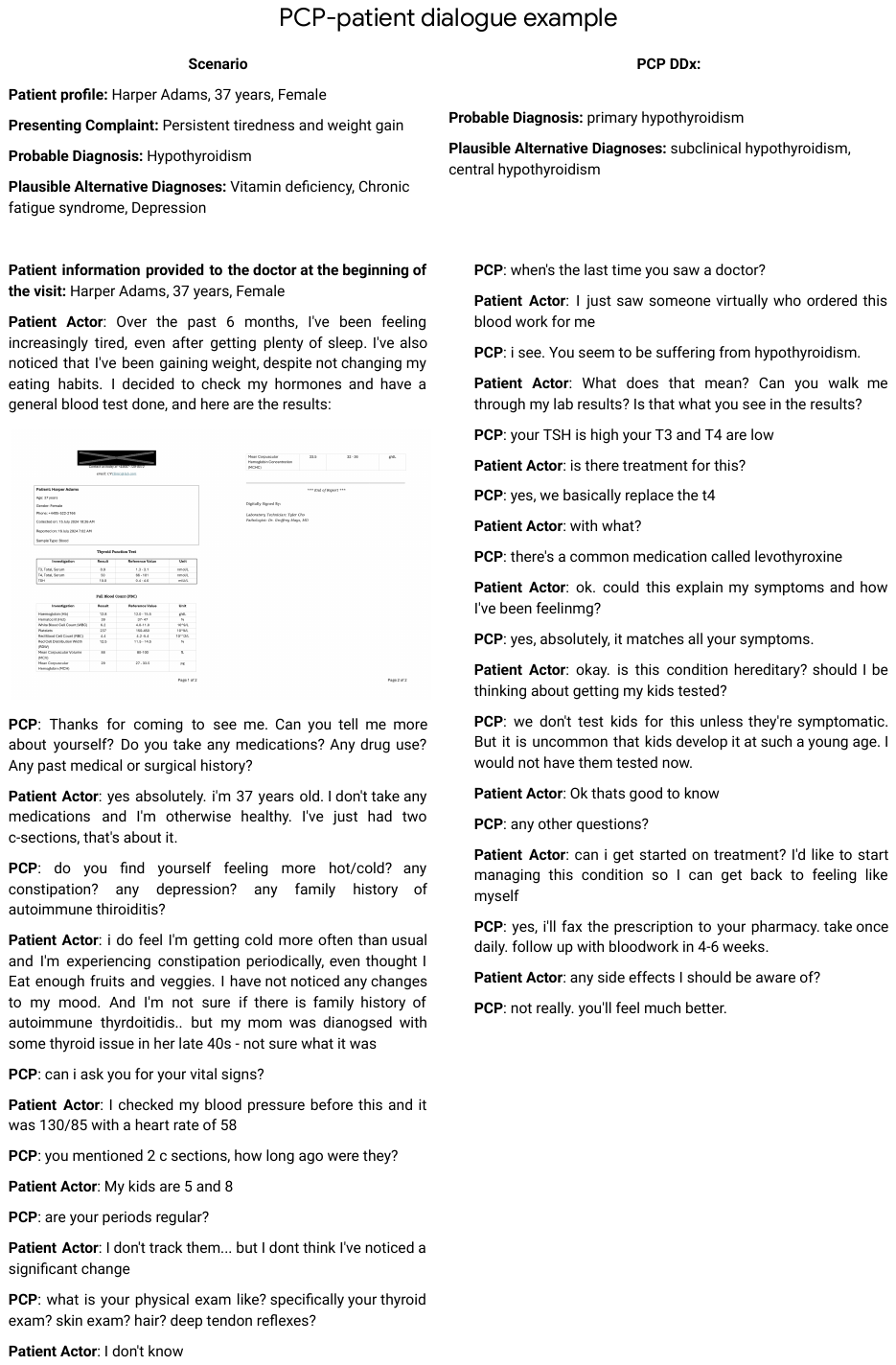}
     \label{fig:dialogue_4}
\end{figure}

\begin{figure}[H]
    \caption{\footnotesize \textbf{Example of AMIE-patient dialogue for a scenario based on ECG measurements.}}
    \label{fig:AMIE_ECG_dialogue}
    \centering
    \includegraphics[trim= 2.5cm 2.5cm 2.5cm 3.5cm, width=1\textwidth, height=\textheight,clip,keepaspectratio]{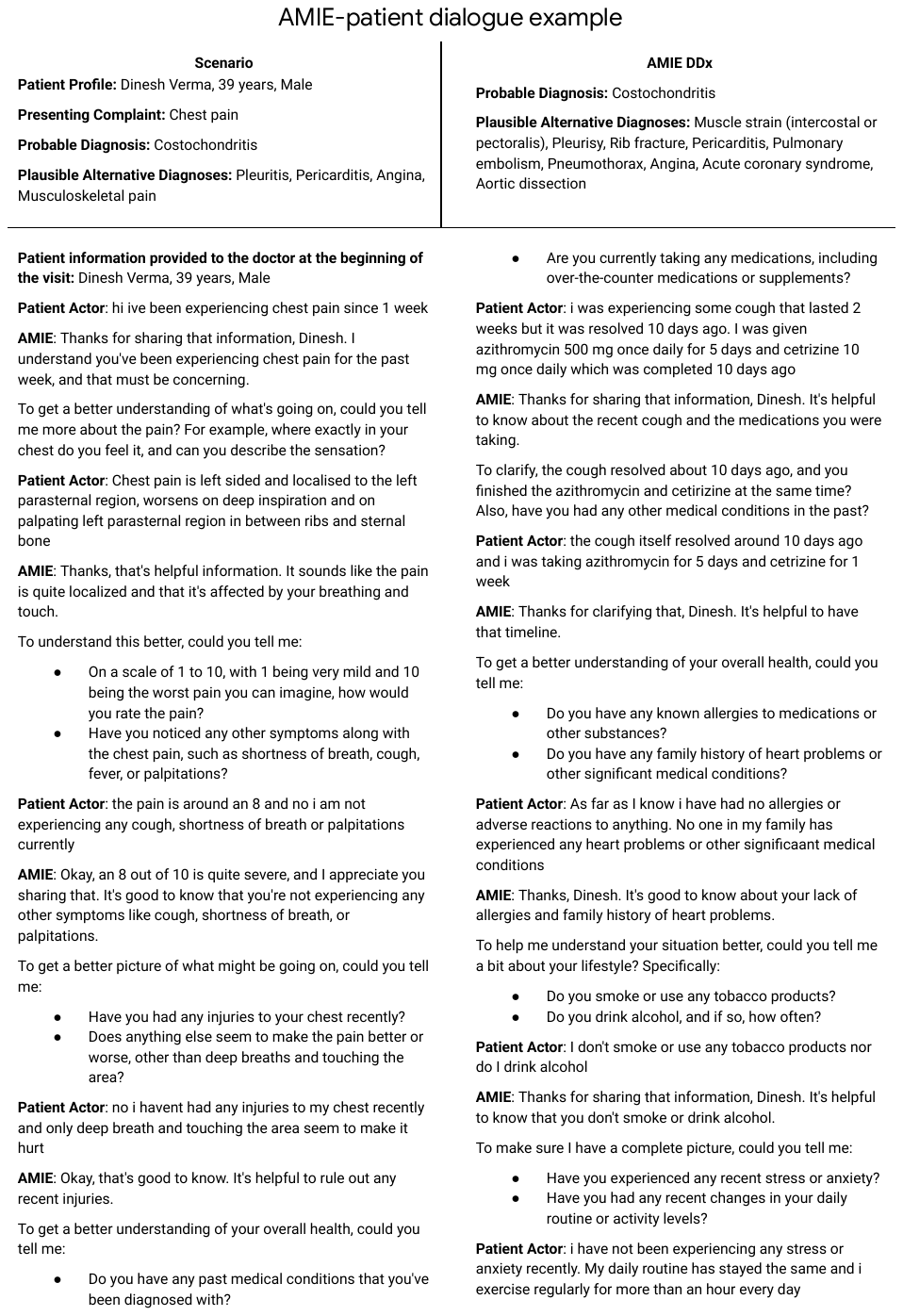}
     \label{fig:dialogue_5}
\end{figure}

\begin{figure}[H]
    \caption{\footnotesize \textbf{Example of PCP-patient dialogue for a scenario based on ECG measurements.}}
    \label{fig:PCP_ECG_dialogue}
    \centering
    \includegraphics[trim= 2.5cm 2.5cm 2.5cm 3.5cm, width=0.98\textwidth, height=0.98\textheight,clip,keepaspectratio]{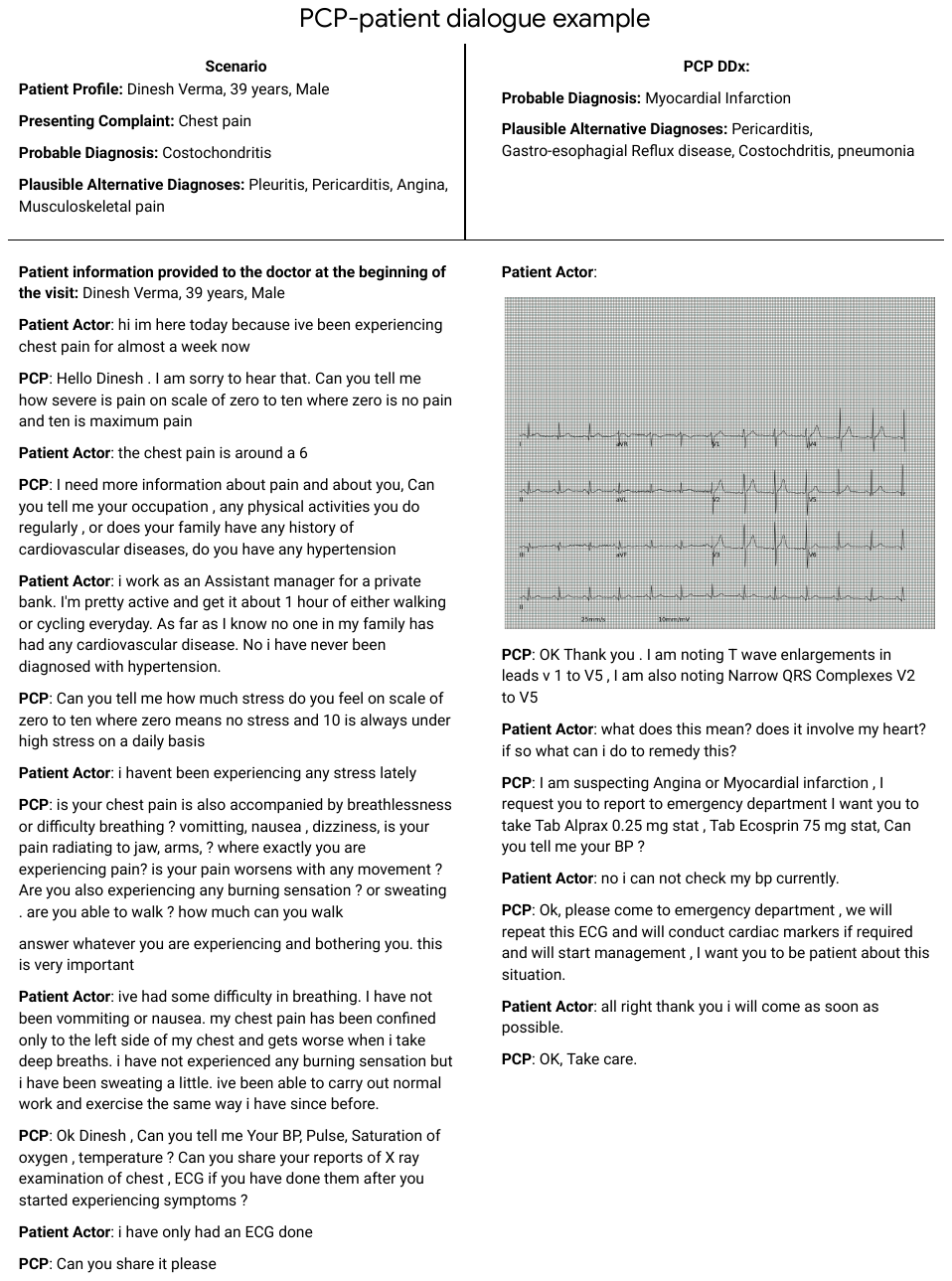}
    \label{fig:dialogue_6}
\end{figure}

Here we include for illustration purposes some examples of dialogue transcripts collected from our virtual OSCE study. For each of three artifact types (skin pictures, clinical documents and ECGs), we provide a pair of AMIE-patient and PCP-patient dialogues. See Figures~\ref{fig:dialogue_1}, \ref{fig:dialogue_2}, \ref{fig:dialogue_3}, \ref{fig:dialogue_4}, \ref{fig:dialogue_5} and \ref{fig:dialogue_6}. 

\begin{figure}[H]
    \centering
    \includegraphics[trim= 2.5cm 2.5cm 2.5cm 2.5cm, width=1\textwidth, height=\textheight,clip,keepaspectratio]{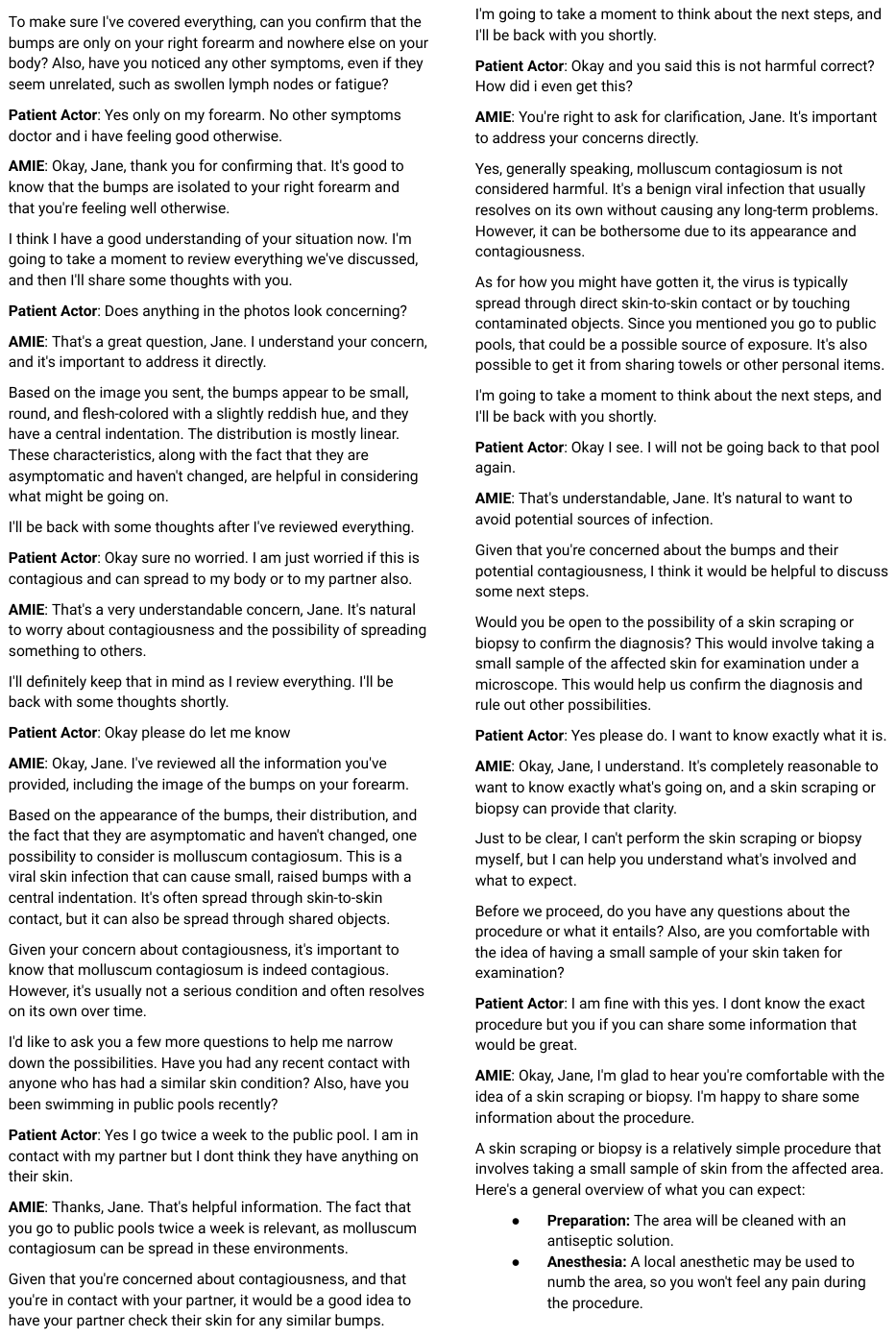}       
\end{figure}

\begin{figure}[H]
    \centering
    \includegraphics[trim= 2.5cm 2.5cm 2.5cm 2.5cm, width=1\textwidth, height=\textheight,clip,keepaspectratio]{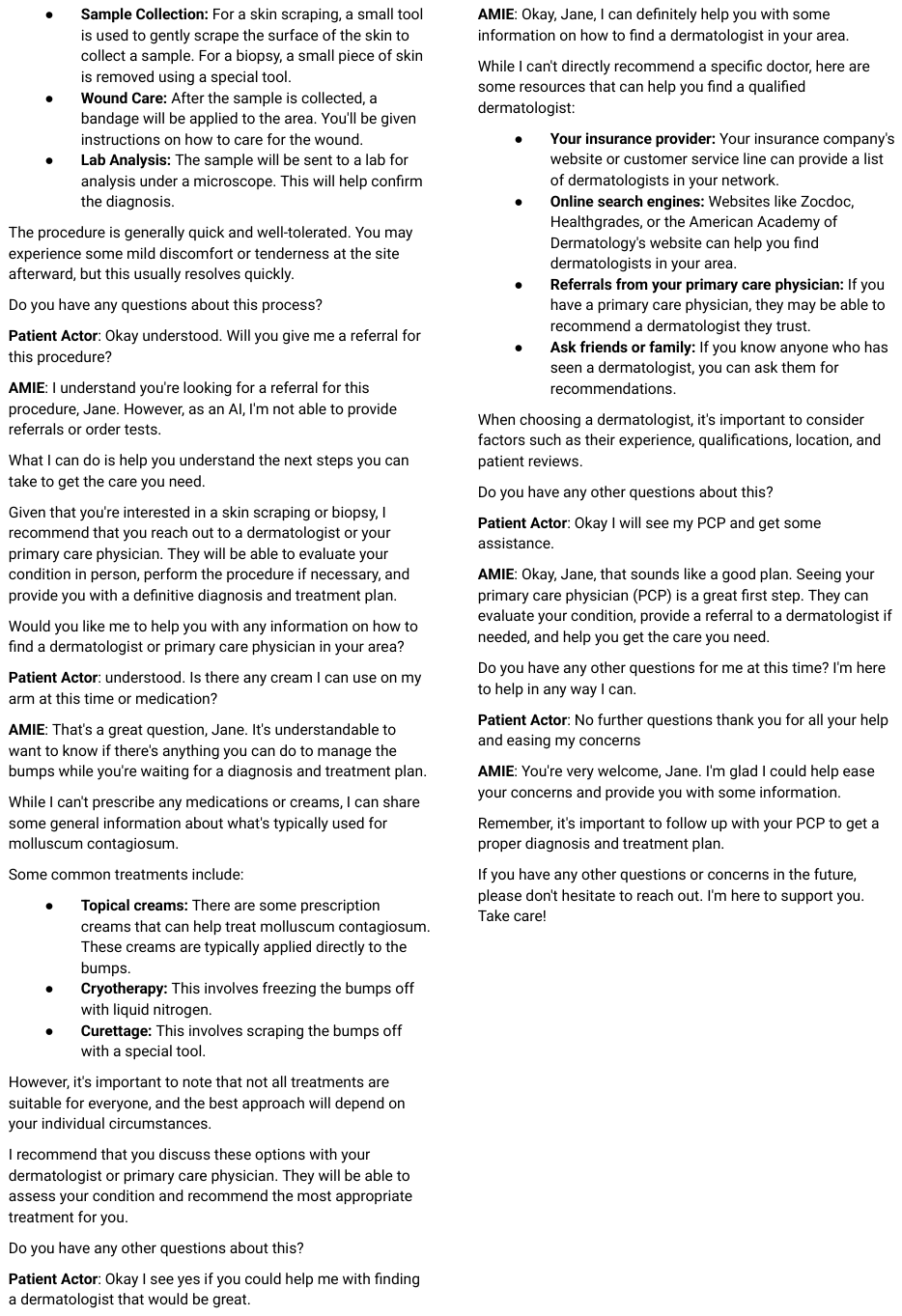}
\end{figure}

\begin{figure}[H]
    \centering
    \includegraphics[trim= 2.5cm 2.5cm 2.5cm 2.5cm, width=1\textwidth, height=\textheight,clip,keepaspectratio]{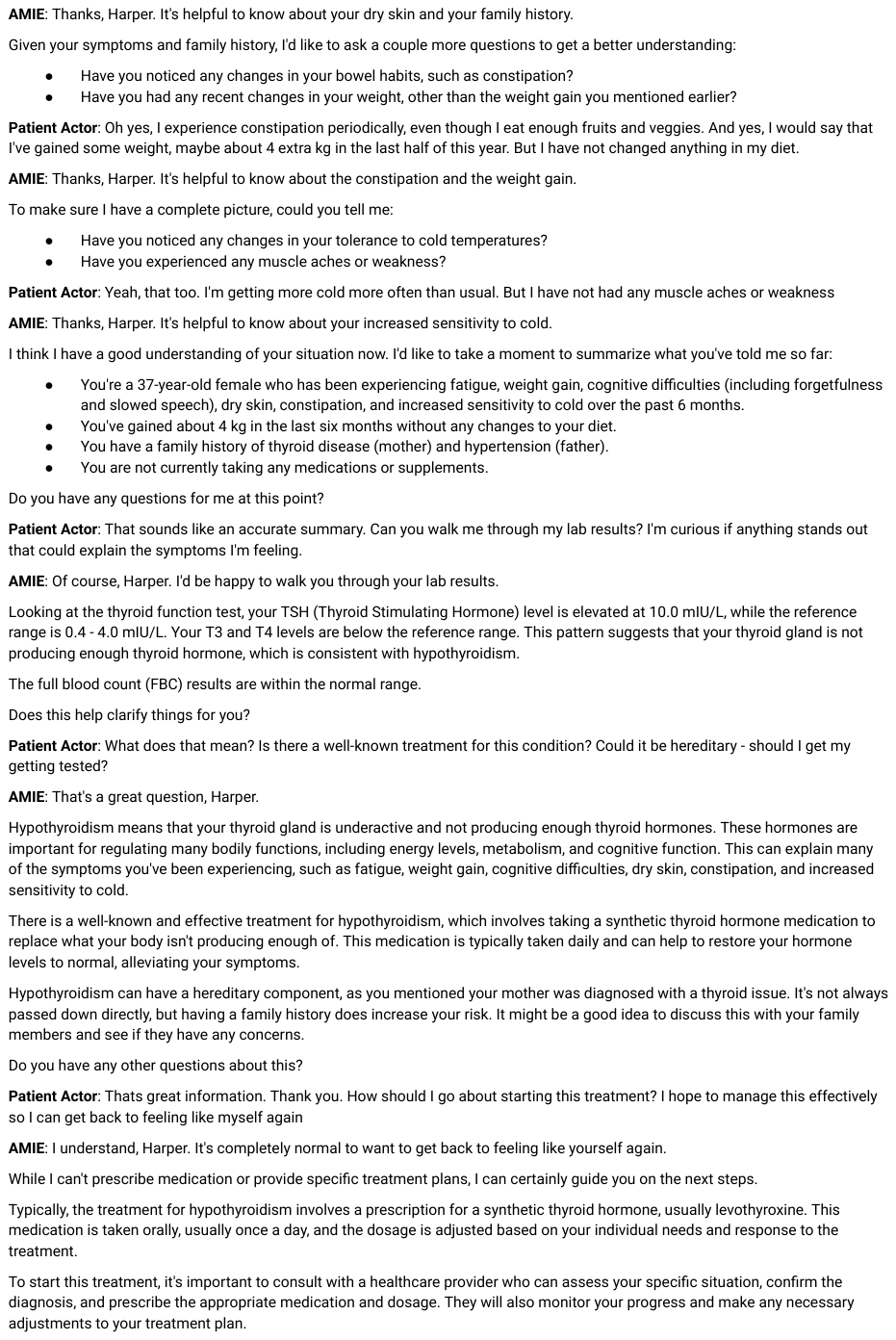}
\end{figure}

\begin{figure}[H]
    \centering
    \includegraphics[trim= 2.5cm 2.5cm 2.5cm 2.5cm, width=1\textwidth, height=\textheight,clip,keepaspectratio]{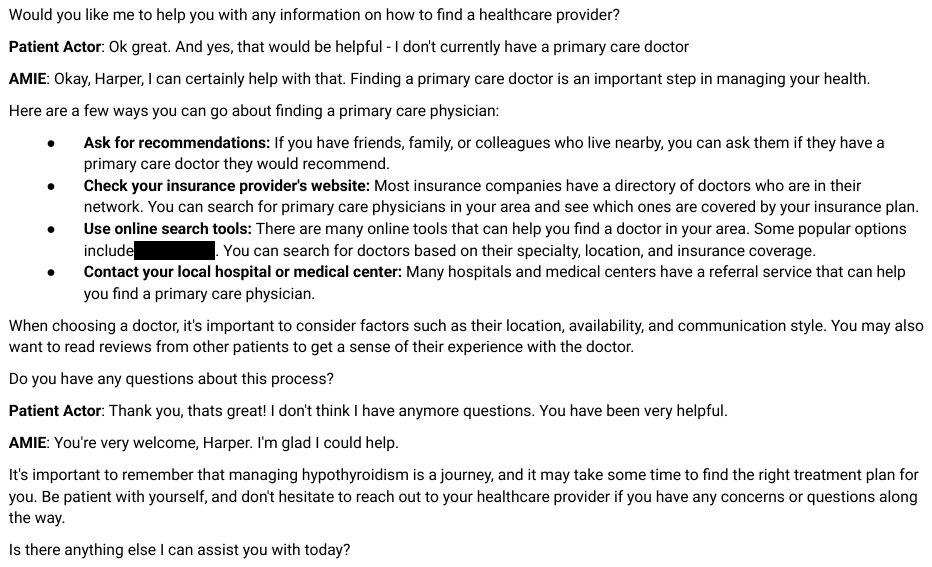}
\end{figure}

\begin{figure}[H]
    \centering
    \includegraphics[trim= 2.5cm 2.5cm 2.5cm 2.5cm, width=1\textwidth, height=\textheight,clip,keepaspectratio]{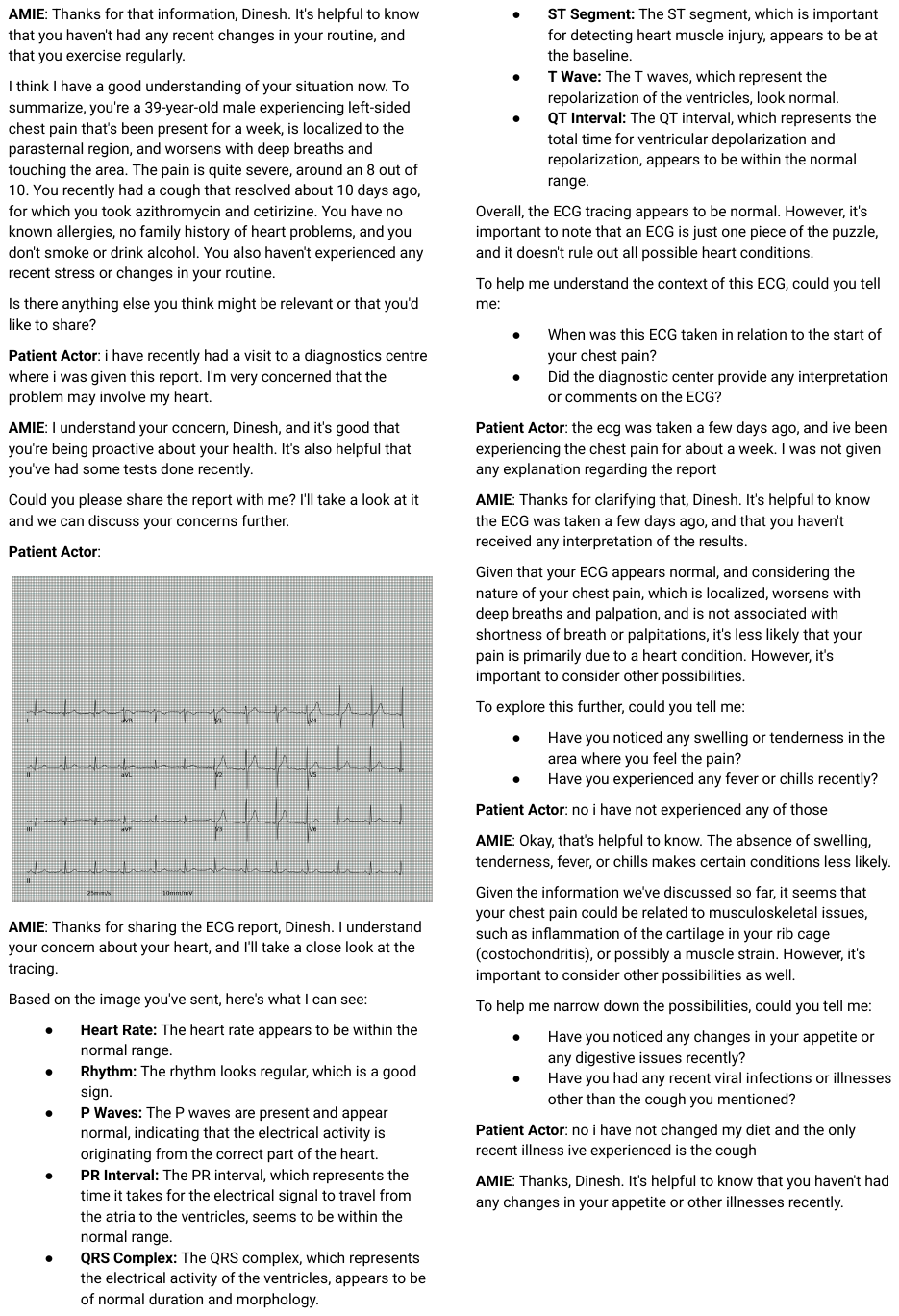}
\end{figure}

\begin{figure}[H]
    \centering
    \includegraphics[trim= 2.5cm 2.5cm 2.5cm 2.5cm, width=1\textwidth, height=\textheight,clip,keepaspectratio]{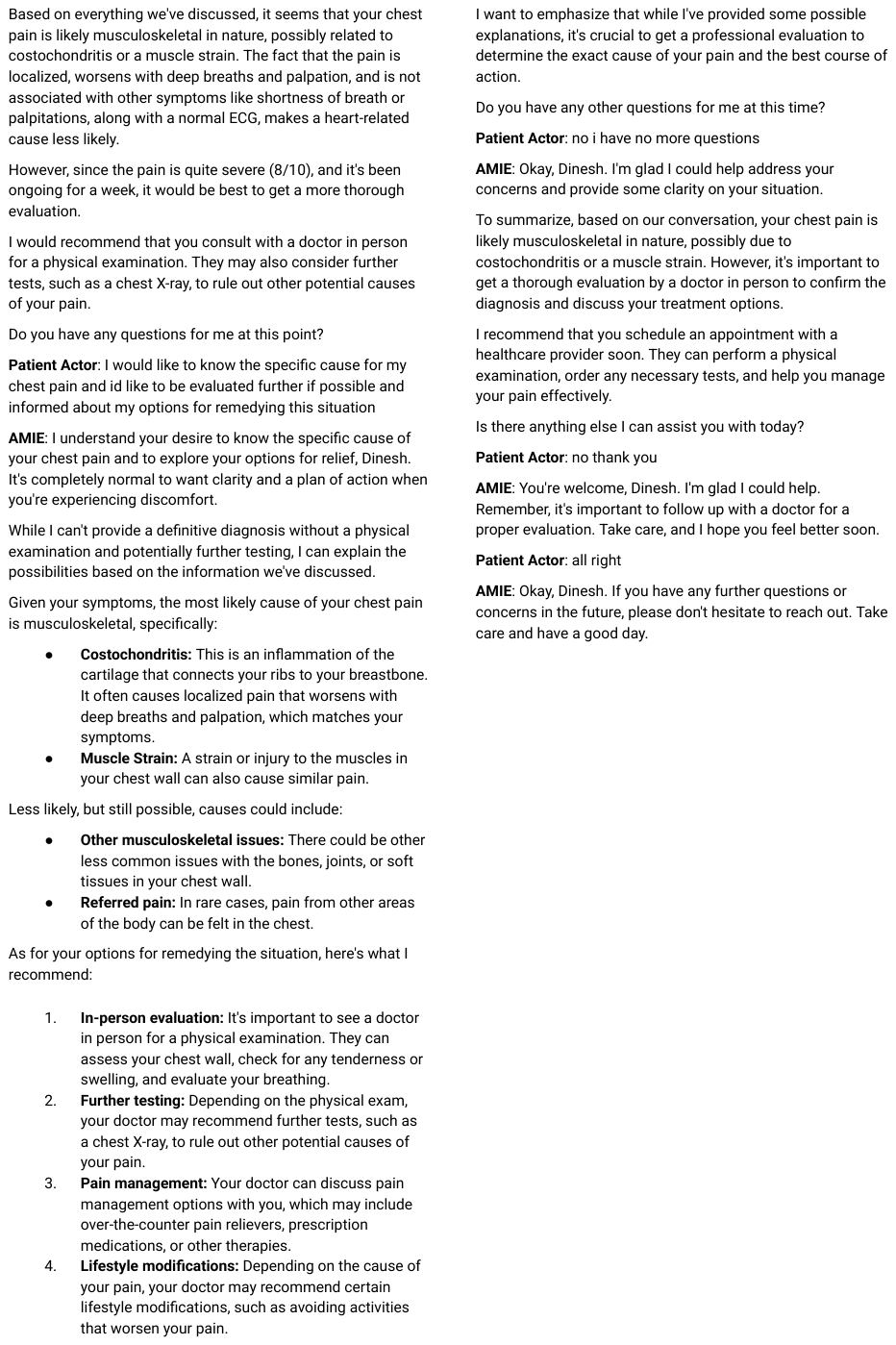}
\end{figure}

%% file: main.bib
@ARTICLE{Zeltzer2025-ks,
  title    = "Comparison of initial artificial intelligence ({AI}) and final
              physician recommendations in {AI}-assisted virtual urgent care
              visits",
  author   = "Zeltzer, Dan and Kugler, Zehavi and Hayat, Lior and Brufman, Tamar
              and Ilan Ber, Ran and Leibovich, Keren and Beer, Tom and Frank,
              Ilan and Shaul, Ran and Goldzweig, Caroline and Pevnick, Joshua",
  journal  = "Ann. Intern. Med.",
  month    =  apr,
  year     =  2025,
  language = "en"
}

@article{roccas2002big,
  title={The big five personality factors and personal values},
  author={Roccas, Sonia and Sagiv, Lilach and Schwartz, Shalom H and Knafo, Ariel},
  journal={Personality and social psychology bulletin},
  volume={28},
  number={6},
  pages={789--801},
  year={2002},
  publisher={Sage Publications Sage CA: Thousand Oaks, CA}
}

@article{tu2025towards,
  title={Towards conversational diagnostic artificial intelligence},
  author={Tu, Tao and Schaekermann, Mike and Palepu, Anil and Saab, Khaled and Freyberg, Jan and Tanno, Ryutaro and Wang, Amy and Li, Brenna and Amin, Mohamed and Cheng, Yong and others},
  journal={Nature},
  pages={1--9},
  year={2025},
  publisher={Nature Publishing Group UK London}
}

@misc{geminiteam2024gemini15unlockingmultimodal,
      title={Gemini 1.5: Unlocking multimodal understanding across millions of tokens of context}, 
      author={Gemini Team and Petko Georgiev and Ving Ian Lei and Ryan Burnell and Libin Bai and Anmol Gulati and Garrett Tanzer and Damien Vincent and Zhufeng Pan and Shibo Wang and Soroosh Mariooryad and Yifan Ding and Xinyang Geng and Fred Alcober and Roy Frostig and Mark Omernick and Lexi Walker and Cosmin Paduraru and Christina Sorokin and Andrea Tacchetti and Colin Gaffney and Samira Daruki and Olcan Sercinoglu and Zach Gleicher and Juliette Love and Paul Voigtlaender and Rohan Jain and Gabriela Surita and Kareem Mohamed and Rory Blevins and Junwhan Ahn and Tao Zhu and Kornraphop Kawintiranon and Orhan Firat and Yiming Gu and Yujing Zhang and Matthew Rahtz and Manaal Faruqui and Natalie Clay and Justin Gilmer and JD Co-Reyes and Ivo Penchev and Rui Zhu and Nobuyuki Morioka and Kevin Hui and Krishna Haridasan and Victor Campos and Mahdis Mahdieh and Mandy Guo and Samer Hassan and Kevin Kilgour and Arpi Vezer and Heng-Tze Cheng and Raoul de Liedekerke and Siddharth Goyal and Paul Barham and DJ Strouse and Seb Noury and Jonas Adler and Mukund Sundararajan and Sharad Vikram and Dmitry Lepikhin and Michela Paganini and Xavier Garcia and Fan Yang and Dasha Valter and Maja Trebacz and Kiran Vodrahalli and Chulayuth Asawaroengchai and Roman Ring and Norbert Kalb and Livio Baldini Soares and Siddhartha Brahma and David Steiner and Tianhe Yu and Fabian Mentzer and Antoine He and Lucas Gonzalez and Bibo Xu and Raphael Lopez Kaufman and Laurent El Shafey and Junhyuk Oh and Tom Hennigan and George van den Driessche and Seth Odoom and Mario Lucic and Becca Roelofs and Sid Lall and Amit Marathe and Betty Chan and Santiago Ontanon and Luheng He and Denis Teplyashin and Jonathan Lai and Phil Crone and Bogdan Damoc and Lewis Ho and Sebastian Riedel and Karel Lenc and Chih-Kuan Yeh and Aakanksha Chowdhery and Yang Xu and Mehran Kazemi and Ehsan Amid and Anastasia Petrushkina and Kevin Swersky and Ali Khodaei and Gowoon Chen and Chris Larkin and Mario Pinto and Geng Yan and Adria Puigdomenech Badia and Piyush Patil and Steven Hansen and Dave Orr and Sebastien M. R. Arnold and Jordan Grimstad and Andrew Dai and Sholto Douglas and Rishika Sinha and Vikas Yadav and Xi Chen and Elena Gribovskaya and Jacob Austin and Jeffrey Zhao and Kaushal Patel and Paul Komarek and Sophia Austin and Sebastian Borgeaud and Linda Friso and Abhimanyu Goyal and Ben Caine and Kris Cao and Da-Woon Chung and Matthew Lamm and Gabe Barth-Maron and Thais Kagohara and Kate Olszewska and Mia Chen and Kaushik Shivakumar and Rishabh Agarwal and Harshal Godhia and Ravi Rajwar and Javier Snaider and Xerxes Dotiwalla and Yuan Liu and Aditya Barua and Victor Ungureanu and Yuan Zhang and Bat-Orgil Batsaikhan and Mateo Wirth and James Qin and Ivo Danihelka and Tulsee Doshi and Martin Chadwick and Jilin Chen and Sanil Jain and Quoc Le and Arjun Kar and Madhu Gurumurthy and Cheng Li and Ruoxin Sang and Fangyu Liu and Lampros Lamprou and Rich Munoz and Nathan Lintz and Harsh Mehta and Heidi Howard and Malcolm Reynolds and Lora Aroyo and Quan Wang and Lorenzo Blanco and Albin Cassirer and Jordan Griffith and Dipanjan Das and Stephan Lee and Jakub Sygnowski and Zach Fisher and James Besley and Richard Powell and Zafarali Ahmed and Dominik Paulus and David Reitter and Zalan Borsos and Rishabh Joshi and Aedan Pope and Steven Hand and Vittorio Selo and Vihan Jain and Nikhil Sethi and Megha Goel and Takaki Makino and Rhys May and Zhen Yang and Johan Schalkwyk and Christina Butterfield and Anja Hauth and Alex Goldin and Will Hawkins and Evan Senter and Sergey Brin and Oliver Woodman and Marvin Ritter and Eric Noland and Minh Giang and Vijay Bolina and Lisa Lee and Tim Blyth and Ian Mackinnon and Machel Reid and Obaid Sarvana and David Silver and Alexander Chen and Lily Wang and Loren Maggiore and Oscar Chang and Nithya Attaluri and Gregory Thornton and Chung-Cheng Chiu and Oskar Bunyan and Nir Levine and Timothy Chung and Evgenii Eltyshev and Xiance Si and Timothy Lillicrap and Demetra Brady and Vaibhav Aggarwal and Boxi Wu and Yuanzhong Xu and Ross McIlroy and Kartikeya Badola and Paramjit Sandhu and Erica Moreira and Wojciech Stokowiec and Ross Hemsley and Dong Li and Alex Tudor and Pranav Shyam and Elahe Rahimtoroghi and Salem Haykal and Pablo Sprechmann and Xiang Zhou and Diana Mincu and Yujia Li and Ravi Addanki and Kalpesh Krishna and Xiao Wu and Alexandre Frechette and Matan Eyal and Allan Dafoe and Dave Lacey and Jay Whang and Thi Avrahami and Ye Zhang and Emanuel Taropa and Hanzhao Lin and Daniel Toyama and Eliza Rutherford and Motoki Sano and HyunJeong Choe and Alex Tomala and Chalence Safranek-Shrader and Nora Kassner and Mantas Pajarskas and Matt Harvey and Sean Sechrist and Meire Fortunato and Christina Lyu and Gamaleldin Elsayed and Chenkai Kuang and James Lottes and Eric Chu and Chao Jia and Chih-Wei Chen and Peter Humphreys and Kate Baumli and Connie Tao and Rajkumar Samuel and Cicero Nogueira dos Santos and Anders Andreassen and Nemanja Rakićević and Dominik Grewe and Aviral Kumar and Stephanie Winkler and Jonathan Caton and Andrew Brock and Sid Dalmia and Hannah Sheahan and Iain Barr and Yingjie Miao and Paul Natsev and Jacob Devlin and Feryal Behbahani and Flavien Prost and Yanhua Sun and Artiom Myaskovsky and Thanumalayan Sankaranarayana Pillai and Dan Hurt and Angeliki Lazaridou and Xi Xiong and Ce Zheng and Fabio Pardo and Xiaowei Li and Dan Horgan and Joe Stanton and Moran Ambar and Fei Xia and Alejandro Lince and Mingqiu Wang and Basil Mustafa and Albert Webson and Hyo Lee and Rohan Anil and Martin Wicke and Timothy Dozat and Abhishek Sinha and Enrique Piqueras and Elahe Dabir and Shyam Upadhyay and Anudhyan Boral and Lisa Anne Hendricks and Corey Fry and Josip Djolonga and Yi Su and Jake Walker and Jane Labanowski and Ronny Huang and Vedant Misra and Jeremy Chen and RJ Skerry-Ryan and Avi Singh and Shruti Rijhwani and Dian Yu and Alex Castro-Ros and Beer Changpinyo and Romina Datta and Sumit Bagri and Arnar Mar Hrafnkelsson and Marcello Maggioni and Daniel Zheng and Yury Sulsky and Shaobo Hou and Tom Le Paine and Antoine Yang and Jason Riesa and Dominika Rogozinska and Dror Marcus and Dalia El Badawy and Qiao Zhang and Luyu Wang and Helen Miller and Jeremy Greer and Lars Lowe Sjos and Azade Nova and Heiga Zen and Rahma Chaabouni and Mihaela Rosca and Jiepu Jiang and Charlie Chen and Ruibo Liu and Tara Sainath and Maxim Krikun and Alex Polozov and Jean-Baptiste Lespiau and Josh Newlan and Zeyncep Cankara and Soo Kwak and Yunhan Xu and Phil Chen and Andy Coenen and Clemens Meyer and Katerina Tsihlas and Ada Ma and Juraj Gottweis and Jinwei Xing and Chenjie Gu and Jin Miao and Christian Frank and Zeynep Cankara and Sanjay Ganapathy and Ishita Dasgupta and Steph Hughes-Fitt and Heng Chen and David Reid and Keran Rong and Hongmin Fan and Joost van Amersfoort and Vincent Zhuang and Aaron Cohen and Shixiang Shane Gu and Anhad Mohananey and Anastasija Ilic and Taylor Tobin and John Wieting and Anna Bortsova and Phoebe Thacker and Emma Wang and Emily Caveness and Justin Chiu and Eren Sezener and Alex Kaskasoli and Steven Baker and Katie Millican and Mohamed Elhawaty and Kostas Aisopos and Carl Lebsack and Nathan Byrd and Hanjun Dai and Wenhao Jia and Matthew Wiethoff and Elnaz Davoodi and Albert Weston and Lakshman Yagati and Arun Ahuja and Isabel Gao and Golan Pundak and Susan Zhang and Michael Azzam and Khe Chai Sim and Sergi Caelles and James Keeling and Abhanshu Sharma and Andy Swing and YaGuang Li and Chenxi Liu and Carrie Grimes Bostock and Yamini Bansal and Zachary Nado and Ankesh Anand and Josh Lipschultz and Abhijit Karmarkar and Lev Proleev and Abe Ittycheriah and Soheil Hassas Yeganeh and George Polovets and Aleksandra Faust and Jiao Sun and Alban Rrustemi and Pen Li and Rakesh Shivanna and Jeremiah Liu and Chris Welty and Federico Lebron and Anirudh Baddepudi and Sebastian Krause and Emilio Parisotto and Radu Soricut and Zheng Xu and Dawn Bloxwich and Melvin Johnson and Behnam Neyshabur and Justin Mao-Jones and Renshen Wang and Vinay Ramasesh and Zaheer Abbas and Arthur Guez and Constant Segal and Duc Dung Nguyen and James Svensson and Le Hou and Sarah York and Kieran Milan and Sophie Bridgers and Wiktor Gworek and Marco Tagliasacchi and James Lee-Thorp and Michael Chang and Alexey Guseynov and Ale Jakse Hartman and Michael Kwong and Ruizhe Zhao and Sheleem Kashem and Elizabeth Cole and Antoine Miech and Richard Tanburn and Mary Phuong and Filip Pavetic and Sebastien Cevey and Ramona Comanescu and Richard Ives and Sherry Yang and Cosmo Du and Bo Li and Zizhao Zhang and Mariko Iinuma and Clara Huiyi Hu and Aurko Roy and Shaan Bijwadia and Zhenkai Zhu and Danilo Martins and Rachel Saputro and Anita Gergely and Steven Zheng and Dawei Jia and Ioannis Antonoglou and Adam Sadovsky and Shane Gu and Yingying Bi and Alek Andreev and Sina Samangooei and Mina Khan and Tomas Kocisky and Angelos Filos and Chintu Kumar and Colton Bishop and Adams Yu and Sarah Hodkinson and Sid Mittal and Premal Shah and Alexandre Moufarek and Yong Cheng and Adam Bloniarz and Jaehoon Lee and Pedram Pejman and Paul Michel and Stephen Spencer and Vladimir Feinberg and Xuehan Xiong and Nikolay Savinov and Charlotte Smith and Siamak Shakeri and Dustin Tran and Mary Chesus and Bernd Bohnet and George Tucker and Tamara von Glehn and Carrie Muir and Yiran Mao and Hideto Kazawa and Ambrose Slone and Kedar Soparkar and Disha Shrivastava and James Cobon-Kerr and Michael Sharman and Jay Pavagadhi and Carlos Araya and Karolis Misiunas and Nimesh Ghelani and Michael Laskin and David Barker and Qiujia Li and Anton Briukhov and Neil Houlsby and Mia Glaese and Balaji Lakshminarayanan and Nathan Schucher and Yunhao Tang and Eli Collins and Hyeontaek Lim and Fangxiaoyu Feng and Adria Recasens and Guangda Lai and Alberto Magni and Nicola De Cao and Aditya Siddhant and Zoe Ashwood and Jordi Orbay and Mostafa Dehghani and Jenny Brennan and Yifan He and Kelvin Xu and Yang Gao and Carl Saroufim and James Molloy and Xinyi Wu and Seb Arnold and Solomon Chang and Julian Schrittwieser and Elena Buchatskaya and Soroush Radpour and Martin Polacek and Skye Giordano and Ankur Bapna and Simon Tokumine and Vincent Hellendoorn and Thibault Sottiaux and Sarah Cogan and Aliaksei Severyn and Mohammad Saleh and Shantanu Thakoor and Laurent Shefey and Siyuan Qiao and Meenu Gaba and Shuo-yiin Chang and Craig Swanson and Biao Zhang and Benjamin Lee and Paul Kishan Rubenstein and Gan Song and Tom Kwiatkowski and Anna Koop and Ajay Kannan and David Kao and Parker Schuh and Axel Stjerngren and Golnaz Ghiasi and Gena Gibson and Luke Vilnis and Ye Yuan and Felipe Tiengo Ferreira and Aishwarya Kamath and Ted Klimenko and Ken Franko and Kefan Xiao and Indro Bhattacharya and Miteyan Patel and Rui Wang and Alex Morris and Robin Strudel and Vivek Sharma and Peter Choy and Sayed Hadi Hashemi and Jessica Landon and Mara Finkelstein and Priya Jhakra and Justin Frye and Megan Barnes and Matthew Mauger and Dennis Daun and Khuslen Baatarsukh and Matthew Tung and Wael Farhan and Henryk Michalewski and Fabio Viola and Felix de Chaumont Quitry and Charline Le Lan and Tom Hudson and Qingze Wang and Felix Fischer and Ivy Zheng and Elspeth White and Anca Dragan and Jean-baptiste Alayrac and Eric Ni and Alexander Pritzel and Adam Iwanicki and Michael Isard and Anna Bulanova and Lukas Zilka and Ethan Dyer and Devendra Sachan and Srivatsan Srinivasan and Hannah Muckenhirn and Honglong Cai and Amol Mandhane and Mukarram Tariq and Jack W. Rae and Gary Wang and Kareem Ayoub and Nicholas FitzGerald and Yao Zhao and Woohyun Han and Chris Alberti and Dan Garrette and Kashyap Krishnakumar and Mai Gimenez and Anselm Levskaya and Daniel Sohn and Josip Matak and Inaki Iturrate and Michael B. Chang and Jackie Xiang and Yuan Cao and Nishant Ranka and Geoff Brown and Adrian Hutter and Vahab Mirrokni and Nanxin Chen and Kaisheng Yao and Zoltan Egyed and Francois Galilee and Tyler Liechty and Praveen Kallakuri and Evan Palmer and Sanjay Ghemawat and Jasmine Liu and David Tao and Chloe Thornton and Tim Green and Mimi Jasarevic and Sharon Lin and Victor Cotruta and Yi-Xuan Tan and Noah Fiedel and Hongkun Yu and Ed Chi and Alexander Neitz and Jens Heitkaemper and Anu Sinha and Denny Zhou and Yi Sun and Charbel Kaed and Brice Hulse and Swaroop Mishra and Maria Georgaki and Sneha Kudugunta and Clement Farabet and Izhak Shafran and Daniel Vlasic and Anton Tsitsulin and Rajagopal Ananthanarayanan and Alen Carin and Guolong Su and Pei Sun and Shashank V and Gabriel Carvajal and Josef Broder and Iulia Comsa and Alena Repina and William Wong and Warren Weilun Chen and Peter Hawkins and Egor Filonov and Lucia Loher and Christoph Hirnschall and Weiyi Wang and Jingchen Ye and Andrea Burns and Hardie Cate and Diana Gage Wright and Federico Piccinini and Lei Zhang and Chu-Cheng Lin and Ionel Gog and Yana Kulizhskaya and Ashwin Sreevatsa and Shuang Song and Luis C. Cobo and Anand Iyer and Chetan Tekur and Guillermo Garrido and Zhuyun Xiao and Rupert Kemp and Huaixiu Steven Zheng and Hui Li and Ananth Agarwal and Christel Ngani and Kati Goshvadi and Rebeca Santamaria-Fernandez and Wojciech Fica and Xinyun Chen and Chris Gorgolewski and Sean Sun and Roopal Garg and Xinyu Ye and S. M. Ali Eslami and Nan Hua and Jon Simon and Pratik Joshi and Yelin Kim and Ian Tenney and Sahitya Potluri and Lam Nguyen Thiet and Quan Yuan and Florian Luisier and Alexandra Chronopoulou and Salvatore Scellato and Praveen Srinivasan and Minmin Chen and Vinod Koverkathu and Valentin Dalibard and Yaming Xu and Brennan Saeta and Keith Anderson and Thibault Sellam and Nick Fernando and Fantine Huot and Junehyuk Jung and Mani Varadarajan and Michael Quinn and Amit Raul and Maigo Le and Ruslan Habalov and Jon Clark and Komal Jalan and Kalesha Bullard and Achintya Singhal and Thang Luong and Boyu Wang and Sujeevan Rajayogam and Julian Eisenschlos and Johnson Jia and Daniel Finchelstein and Alex Yakubovich and Daniel Balle and Michael Fink and Sameer Agarwal and Jing Li and Dj Dvijotham and Shalini Pal and Kai Kang and Jaclyn Konzelmann and Jennifer Beattie and Olivier Dousse and Diane Wu and Remi Crocker and Chen Elkind and Siddhartha Reddy Jonnalagadda and Jong Lee and Dan Holtmann-Rice and Krystal Kallarackal and Rosanne Liu and Denis Vnukov and Neera Vats and Luca Invernizzi and Mohsen Jafari and Huanjie Zhou and Lilly Taylor and Jennifer Prendki and Marcus Wu and Tom Eccles and Tianqi Liu and Kavya Kopparapu and Francoise Beaufays and Christof Angermueller and Andreea Marzoca and Shourya Sarcar and Hilal Dib and Jeff Stanway and Frank Perbet and Nejc Trdin and Rachel Sterneck and Andrey Khorlin and Dinghua Li and Xihui Wu and Sonam Goenka and David Madras and Sasha Goldshtein and Willi Gierke and Tong Zhou and Yaxin Liu and Yannie Liang and Anais White and Yunjie Li and Shreya Singh and Sanaz Bahargam and Mark Epstein and Sujoy Basu and Li Lao and Adnan Ozturel and Carl Crous and Alex Zhai and Han Lu and Zora Tung and Neeraj Gaur and Alanna Walton and Lucas Dixon and Ming Zhang and Amir Globerson and Grant Uy and Andrew Bolt and Olivia Wiles and Milad Nasr and Ilia Shumailov and Marco Selvi and Francesco Piccinno and Ricardo Aguilar and Sara McCarthy and Misha Khalman and Mrinal Shukla and Vlado Galic and John Carpenter and Kevin Villela and Haibin Zhang and Harry Richardson and James Martens and Matko Bosnjak and Shreyas Rammohan Belle and Jeff Seibert and Mahmoud Alnahlawi and Brian McWilliams and Sankalp Singh and Annie Louis and Wen Ding and Dan Popovici and Lenin Simicich and Laura Knight and Pulkit Mehta and Nishesh Gupta and Chongyang Shi and Saaber Fatehi and Jovana Mitrovic and Alex Grills and Joseph Pagadora and Dessie Petrova and Danielle Eisenbud and Zhishuai Zhang and Damion Yates and Bhavishya Mittal and Nilesh Tripuraneni and Yannis Assael and Thomas Brovelli and Prateek Jain and Mihajlo Velimirovic and Canfer Akbulut and Jiaqi Mu and Wolfgang Macherey and Ravin Kumar and Jun Xu and Haroon Qureshi and Gheorghe Comanici and Jeremy Wiesner and Zhitao Gong and Anton Ruddock and Matthias Bauer and Nick Felt and Anirudh GP and Anurag Arnab and Dustin Zelle and Jonas Rothfuss and Bill Rosgen and Ashish Shenoy and Bryan Seybold and Xinjian Li and Jayaram Mudigonda and Goker Erdogan and Jiawei Xia and Jiri Simsa and Andrea Michi and Yi Yao and Christopher Yew and Steven Kan and Isaac Caswell and Carey Radebaugh and Andre Elisseeff and Pedro Valenzuela and Kay McKinney and Kim Paterson and Albert Cui and Eri Latorre-Chimoto and Solomon Kim and William Zeng and Ken Durden and Priya Ponnapalli and Tiberiu Sosea and Christopher A. Choquette-Choo and James Manyika and Brona Robenek and Harsha Vashisht and Sebastien Pereira and Hoi Lam and Marko Velic and Denese Owusu-Afriyie and Katherine Lee and Tolga Bolukbasi and Alicia Parrish and Shawn Lu and Jane Park and Balaji Venkatraman and Alice Talbert and Lambert Rosique and Yuchung Cheng and Andrei Sozanschi and Adam Paszke and Praveen Kumar and Jessica Austin and Lu Li and Khalid Salama and Wooyeol Kim and Nandita Dukkipati and Anthony Baryshnikov and Christos Kaplanis and XiangHai Sheng and Yuri Chervonyi and Caglar Unlu and Diego de Las Casas and Harry Askham and Kathryn Tunyasuvunakool and Felix Gimeno and Siim Poder and Chester Kwak and Matt Miecnikowski and Vahab Mirrokni and Alek Dimitriev and Aaron Parisi and Dangyi Liu and Tomy Tsai and Toby Shevlane and Christina Kouridi and Drew Garmon and Adrian Goedeckemeyer and Adam R. Brown and Anitha Vijayakumar and Ali Elqursh and Sadegh Jazayeri and Jin Huang and Sara Mc Carthy and Jay Hoover and Lucy Kim and Sandeep Kumar and Wei Chen and Courtney Biles and Garrett Bingham and Evan Rosen and Lisa Wang and Qijun Tan and David Engel and Francesco Pongetti and Dario de Cesare and Dongseong Hwang and Lily Yu and Jennifer Pullman and Srini Narayanan and Kyle Levin and Siddharth Gopal and Megan Li and Asaf Aharoni and Trieu Trinh and Jessica Lo and Norman Casagrande and Roopali Vij and Loic Matthey and Bramandia Ramadhana and Austin Matthews and CJ Carey and Matthew Johnson and Kremena Goranova and Rohin Shah and Shereen Ashraf and Kingshuk Dasgupta and Rasmus Larsen and Yicheng Wang and Manish Reddy Vuyyuru and Chong Jiang and Joana Ijazi and Kazuki Osawa and Celine Smith and Ramya Sree Boppana and Taylan Bilal and Yuma Koizumi and Ying Xu and Yasemin Altun and Nir Shabat and Ben Bariach and Alex Korchemniy and Kiam Choo and Olaf Ronneberger and Chimezie Iwuanyanwu and Shubin Zhao and David Soergel and Cho-Jui Hsieh and Irene Cai and Shariq Iqbal and Martin Sundermeyer and Zhe Chen and Elie Bursztein and Chaitanya Malaviya and Fadi Biadsy and Prakash Shroff and Inderjit Dhillon and Tejasi Latkar and Chris Dyer and Hannah Forbes and Massimo Nicosia and Vitaly Nikolaev and Somer Greene and Marin Georgiev and Pidong Wang and Nina Martin and Hanie Sedghi and John Zhang and Praseem Banzal and Doug Fritz and Vikram Rao and Xuezhi Wang and Jiageng Zhang and Viorica Patraucean and Dayou Du and Igor Mordatch and Ivan Jurin and Lewis Liu and Ayush Dubey and Abhi Mohan and Janek Nowakowski and Vlad-Doru Ion and Nan Wei and Reiko Tojo and Maria Abi Raad and Drew A. Hudson and Vaishakh Keshava and Shubham Agrawal and Kevin Ramirez and Zhichun Wu and Hoang Nguyen and Ji Liu and Madhavi Sewak and Bryce Petrini and DongHyun Choi and Ivan Philips and Ziyue Wang and Ioana Bica and Ankush Garg and Jarek Wilkiewicz and Priyanka Agrawal and Xiaowei Li and Danhao Guo and Emily Xue and Naseer Shaik and Andrew Leach and Sadh MNM Khan and Julia Wiesinger and Sammy Jerome and Abhishek Chakladar and Alek Wenjiao Wang and Tina Ornduff and Folake Abu and Alireza Ghaffarkhah and Marcus Wainwright and Mario Cortes and Frederick Liu and Joshua Maynez and Slav Petrov and Yonghui Wu and Demis Hassabis and Koray Kavukcuoglu and Jeffrey Dean and Oriol Vinyals},
      year={2024},
      eprint={2403.05530},
      archivePrefix={arXiv},
      primaryClass={cs.CL},
      url={https://arxiv.org/abs/2403.05530}, 
}

@article{Harden1975-ft,
  title     = "Assessment of clinical competence using objective structured examination",
  author    = "RM Harden and M Stevenson and WW Downie and GM Wilson",
  journal   = "BMJ",
  publisher = "BMJ",
  volume    =  1,
  number    =  5955,
  year      =  1975,
  url       =  {https://doi.org/10.1136/bmj.1.5955.447}
}

@article{excerpt_mycin,

title = {Computer-Based Medical Consultations: MYCIN.},
journal = {Annals of Internal Medicine},
volume = {85},
number = {6},
pages = {831-831},
year = {1976},
doi = {10.7326/0003-4819-85-6-831\_1},

URL = { 
    
    
        https://www.acpjournals.org/doi/abs/10.7326/0003-4819-85-6-831_1
    

},
eprint = { 
    
    
        https://www.acpjournals.org/doi/pdf/10.7326/0003-4819-85-6-831_1
    

}
}

@article{brown2020language,
  title={Language models are few-shot learners},
  author={Brown, Tom and Mann, Benjamin and Ryder, Nick and Subbiah, Melanie and Kaplan, Jared D and Dhariwal, Prafulla and Neelakantan, Arvind and Shyam, Pranav and Sastry, Girish and Askell, Amanda and others},
  journal={Advances in neural information processing systems},
  volume={33},
  pages={1877--1901},
  year={2020}
}

@article{saab2024capabilities,
  title={Capabilities of gemini models in medicine},
  author={Saab, Khaled and Tu, Tao and Weng, Wei-Hung and Tanno, Ryutaro and Stutz, David and Wulczyn, Ellery and Zhang, Fan and Strother, Tim and Park, Chunjong and Vedadi, Elahe and others},
  journal={arXiv preprint arXiv:2404.18416},
  year={2024}
}

@article{wang2022self,
  title={Self-consistency improves chain of thought reasoning in language models},
  author={Wang, Xuezhi and Wei, Jason and Schuurmans, Dale and Le, Quoc and Chi, Ed and Zhou, Denny},
  journal={arXiv preprint arXiv:2203.11171},
  year={2022}
}

@article{ouyang2022training,
  title={Training language models to follow instructions with human feedback},
  author={Ouyang, Long and Wu, Jeffrey and Jiang, Xu and Almeida, Diogo and Wainwright, Carroll and Mishkin, Pamela and Zhang, Chong and Agarwal, Sandhini and Slama, Katarina and Ray, Alex and others},
  journal={Advances in neural information processing systems},
  volume={35},
  pages={27730--27744},
  year={2022}
}

@misc{kossen2022active,
      title={Active Acquisition for Multimodal Temporal Data: A Challenging Decision-Making Task}, 
      author={Jannik Kossen and Cătălina Cangea and Eszter Vértes and Andrew Jaegle and Viorica Patraucean and Ira Ktena and Nenad Tomasev and Danielle Belgrave},
      year={2023},
      eprint={2211.05039},
      archivePrefix={arXiv},
      primaryClass={cs.LG},
      url={https://arxiv.org/abs/2211.05039}, 
}

@article{king2013best,
  title={“Best practice” for patient-centered communication: a narrative review},
  author={King, Ann and Hoppe, Ruth B},
  journal={Journal of graduate medical education},
  volume={5},
  number={3},
  pages={385--393},
  year={2013},
  publisher={The Accreditation Council for Graduate Medical Education Suite 2000, 515~…}
}

@article{dacre2003mrcp,
  title={MRCP (UK) PART 2 Clinical Examination (PACES): a review of the first four examination sessions (June 2001--July 2002)},
  author={Dacre, Jane and Besser, Mike and White, Patricia},
  journal={Clinical Medicine},
  volume={3},
  number={5},
  pages={452},
  year={2003},
  publisher={Royal College of Physicians}
}

@article{sloan1995objective,
  title={The Objective Structured Clinical Examination. The new gold standard for evaluating postgraduate clinical performance.},
  author={Sloan, David A and Donnelly, Michael B and Schwartz, Richard W and Strodel, William E},
  journal={Annals of surgery},
  volume={222},
  number={6},
  pages={735},
  year={1995},
  publisher={Lippincott, Williams, and Wilkins}
}

@article{bai2022constitutional,
  title={Constitutional {AI}: Harmlessness from {AI} feedback},
  author={Bai, Yuntao and Kadavath, Saurav and Kundu, Sandipan and Askell, Amanda and Kernion, Jackson and Jones, Andy and Chen, Anna and Goldie, Anna and Mirhoseini, Azalia and McKinnon, Cameron and others},
  journal={arXiv preprint arXiv:2212.08073},
  year={2022}
}

@article{wei2022chain,
  title={Chain-of-thought prompting elicits reasoning in large language models},
  author={Wei, Jason and Wang, Xuezhi and Schuurmans, Dale and Bosma, Maarten and Xia, Fei and Chi, Ed and Le, Quoc V and Zhou, Denny and others},
  journal={Advances in Neural Information Processing Systems},
  volume={35},
  pages={24824--24837},
  year={2022}
}

@misc{mukherjee2024polaris,
      title={Polaris: A Safety-focused LLM Constellation Architecture for Healthcare}, 
      author={Subhabrata Mukherjee and Paul Gamble and Markel Sanz Ausin and Neel Kant and Kriti Aggarwal and Neha Manjunath and Debajyoti Datta and Zhengliang Liu and Jiayuan Ding and Sophia Busacca and Cezanne Bianco and Swapnil Sharma and Rae Lasko and Michelle Voisard and Sanchay Harneja and Darya Filippova and Gerry Meixiong and Kevin Cha and Amir Youssefi and Meyhaa Buvanesh and Howard Weingram and Sebastian Bierman-Lytle and Harpreet Singh Mangat and Kim Parikh and Saad Godil and Alex Miller},
      year={2024},
      eprint={2403.13313},
      archivePrefix={arXiv},
      primaryClass={cs.AI},
      url={https://arxiv.org/abs/2403.13313}, 
}

@ARTICLE{Zeltzer2023-iu,
  title     = "Diagnostic accuracy of artificial intelligence in virtual primary
               care",
  author    = "Zeltzer, Dan and Herzog, Lee and Pickman, Yishai and Steuerman,
               Yael and Ber, Ran Ilan and Kugler, Zehavi and Shaul, Ran and
               Ebbert, Jon O",
  journal   = "Mayo Clinic Proceedings: Digital Health",
  publisher = "Elsevier BV",
  volume    =  1,
  number    =  4,
  pages     = "480--489",
  month     =  dec,
  year      =  2023,
  language  = "en"
}

@ARTICLE{Lizee2024-uy,
  title         = "Conversational medical {AI}: Ready for practice",
  author        = "Lizée, Antoine and Beaucoté, Pierre-Auguste and Whitbeck,
                   James and Doumeingts, Marion and Beaugnon, Anaël and
                   Feldhaus, Isabelle",
  journal       = "arXiv [cs.AI]",
  month         =  nov,
  year          =  2024,
  archivePrefix = "arXiv",
  primaryClass  = "cs.AI",
  language      = "en"
}

@ARTICLE{Shaver2022-yb,
  title     = "The state of telehealth before and after the {COVID}-19 pandemic",
  author    = "Shaver, Julia",
  journal   = "Prim. Care",
  publisher = "Elsevier BV",
  volume    =  49,
  number    =  4,
  pages     = "517--530",
  month     =  dec,
  year      =  2022,
  language  = "en"
}

@article{tu2024towards,
  title={Towards conversational diagnostic ai},
  author={Tu, Tao and Azizi, Shekoofeh and Driess, Danny and Schaekermann, Mike and Amin, Mohamed and Chang, Pi-Chuan and Carroll, Andrew and Lau, Chuck and Tanno, Ryutaro and Ktena, Ira and others},
  journal={arXiv preprint arXiv:2401.05654},
  year={2024},
  url={https://arxiv.org/abs/2401.05654}
}

@misc{li2024conversational,
      title={Conversational Medical AI: Ready for Practice}, 
      author={Antoine Lizée and Pierre-Auguste Beaucoté and James Whitbeck and Marion Doumeingts and Anaël Beaugnon and Isabelle Feldhaus},
      year={2025},
      eprint={2411.12808},
      archivePrefix={arXiv},
      primaryClass={cs.AI},
      url={https://arxiv.org/abs/2411.12808}, 
}

@misc{xu2024reasoninglikedoctorimproving,
      title={Reasoning Like a Doctor: Improving Medical Dialogue Systems via Diagnostic Reasoning Process Alignment}, 
      author={Kaishuai Xu and Yi Cheng and Wenjun Hou and Qiaoyu Tan and Wenjie Li},
      year={2024},
      eprint={2406.13934},
      archivePrefix={arXiv},
      primaryClass={cs.CL},
      url={https://arxiv.org/abs/2406.13934}, 
}

@misc{xu2024medicaldialoguegenerationintuitivethenanalytical,
      title={Medical Dialogue Generation via Intuitive-then-Analytical Differential Diagnosis}, 
      author={Kaishuai Xu and Wenjun Hou and Yi Cheng and Jian Wang and Wenjie Li},
      year={2024},
      eprint={2401.06541},
      archivePrefix={arXiv},
      primaryClass={cs.CL},
      url={https://arxiv.org/abs/2401.06541}, 
}

@misc{ke2024enhancingdiagnosticaccuracymultiagent,
      title={Enhancing Diagnostic Accuracy through Multi-Agent Conversations: Using Large Language Models to Mitigate Cognitive Bias}, 
      author={Yu He Ke and Rui Yang and Sui An Lie and Taylor Xin Yi Lim and Hairil Rizal Abdullah and Daniel Shu Wei Ting and Nan Liu},
      year={2024},
      eprint={2401.14589},
      archivePrefix={arXiv},
      primaryClass={cs.CL},
      url={https://arxiv.org/abs/2401.14589}, 
}

@misc{liu2025dialoguebettermonologueinstructing,
      title={Dialogue is Better Than Monologue: Instructing Medical LLMs via Strategical Conversations}, 
      author={Zijie Liu and Xinyu Zhao and Jie Peng and Zhuangdi Zhu and Qingyu Chen and Xia Hu and Tianlong Chen},
      year={2025},
      eprint={2501.17860},
      archivePrefix={arXiv},
      primaryClass={cs.CL},
      url={https://arxiv.org/abs/2501.17860}, 
}

@misc{artsi2024advancing,
      title={{Advancing Clinical Practice: The Potential of Multimodal Technology in Modern Medicine}}, 
      author={Y Artsi and V Sorin and BS Glicksberg and GN Nadkarni and E Klang},
      year={2024},
      url={https://pmc.ncbi.nlm.nih.gov/articles/PMC11508674/}
}

@misc{schouten2024navigatinglandscapemultimodalai,
      title={Navigating the landscape of multimodal AI in medicine: a scoping review on technical challenges and clinical applications}, 
      author={Daan Schouten and Giulia Nicoletti and Bas Dille and Catherine Chia and Pierpaolo Vendittelli and Megan Schuurmans and Geert Litjens and Nadieh Khalili},
      year={2024},
      eprint={2411.03782},
      archivePrefix={arXiv},
      primaryClass={cs.AI},
      url={https://arxiv.org/abs/2411.03782}, 
}

@article{Xiao_2025,
   title={A comprehensive survey of large language models and multimodal large language models in medicine},
   volume={117},
   ISSN={1566-2535},
   url={http://dx.doi.org/10.1016/j.inffus.2024.102888},
   DOI={10.1016/j.inffus.2024.102888},
   journal={Information Fusion},
   publisher={Elsevier BV},
   author={Xiao, Hanguang and Zhou, Feizhong and Liu, Xingyue and Liu, Tianqi and Li, Zhipeng and Liu, Xin and Huang, Xiaoxuan},
   year={2025},
   month=may, pages={102888} }

@inproceedings{Yildirim_2024, series={CHI ’24},
   title={Multimodal Healthcare AI: Identifying and Designing Clinically Relevant Vision-Language Applications for Radiology},
   url={http://dx.doi.org/10.1145/3613904.3642013},
   DOI={10.1145/3613904.3642013},
   booktitle={Proceedings of the CHI Conference on Human Factors in Computing Systems},
   publisher={ACM},
   author={Yildirim, Nur and Richardson, Hannah and Wetscherek, Maria Teodora and Bajwa, Junaid and Jacob, Joseph and Pinnock, Mark Ames and Harris, Stephen and Coelho De Castro, Daniel and Bannur, Shruthi and Hyland, Stephanie and Ghosh, Pratik and Ranjit, Mercy and Bouzid, Kenza and Schwaighofer, Anton and Pérez-García, Fernando and Sharma, Harshita and Oktay, Ozan and Lungren, Matthew and Alvarez-Valle, Javier and Nori, Aditya and Thieme, Anja},
   year={2024},
   month=may, pages={1–22},
   collection={CHI ’24} }

@misc{simon2024future,
      title={{The future of multimodal artificial intelligence models for integrating imaging and clinical metadata: a narrative review}}, 
      author={BD Simon and KB Ozyoruk and DG Gelikman and SA Harmon and B T{\"u}rkbey},
      year={2024},
      url={https://doi.org/10.4274/dir.2024.242631} 
}

@misc{islam2025multimodalmarvelsdeeplearning,
      title={Multimodal Marvels of Deep Learning in Medical Diagnosis: A Comprehensive Review of COVID-19 Detection}, 
      author={Md Shofiqul Islam and Khondokar Fida Hasan and Hasibul Hossain Shajeeb and Humayan Kabir Rana and Md Saifur Rahmand and Md Munirul Hasan and AKM Azad and Ibrahim Abdullah and Mohammad Ali Moni},
      year={2025},
      eprint={2501.09506},
      archivePrefix={arXiv},
      primaryClass={cs.LG},
      url={https://arxiv.org/abs/2501.09506}, 
}

@misc{bleich2024automatedmedicalreportgeneration,
      title={Automated Medical Report Generation for ECG Data: Bridging Medical Text and Signal Processing with Deep Learning}, 
      author={Amnon Bleich and Antje Linnemann and Bjoern H. Diem and Tim OF Conrad},
      year={2024},
      eprint={2412.04067},
      archivePrefix={arXiv},
      primaryClass={cs.CL},
      url={https://arxiv.org/abs/2412.04067}, 
}

@misc{chen2024gmaimmbenchcomprehensivemultimodalevaluation,
      title={GMAI-MMBench: A Comprehensive Multimodal Evaluation Benchmark Towards General Medical AI}, 
      author={Pengcheng Chen and Jin Ye and Guoan Wang and Yanjun Li and Zhongying Deng and Wei Li and Tianbin Li and Haodong Duan and Ziyan Huang and Yanzhou Su and Benyou Wang and Shaoting Zhang and Bin Fu and Jianfei Cai and Bohan Zhuang and Eric J Seibel and Junjun He and Yu Qiao},
      year={2024},
      eprint={2408.03361},
      archivePrefix={arXiv},
      primaryClass={eess.IV},
      url={https://arxiv.org/abs/2408.03361}, 
}

@article{ward2024creating,
  title={Creating an Empirical Dermatology Dataset Through Crowdsourcing With Web Search Advertisements},
  author={Ward, Abbi and Li, Jimmy and Wang, Julie and Lakshminarasimhan, Sriram and Carrick, Ashley and Campana, Bilson and Hartford, Jay and Sreenivasaiah, Pradeep K and Tiyasirisokchai, Tiya and Virmani, Sunny and others},
  journal={JAMA Network Open},
  volume={7},
  number={11},
  pages={e2446615--e2446615},
  year={2024},
  publisher={American Medical Association}
}

@article{wagner2020ptb,
  title={PTB-XL, a large publicly available electrocardiography dataset},
  author={Wagner, Patrick and Strodthoff, Nils and Bousseljot, Ralf-Dieter and Kreiseler, Dieter and Lunze, Fatima I and Samek, Wojciech and Schaeffter, Tobias},
  journal={Scientific data},
  volume={7},
  number={1},
  pages={1--15},
  year={2020},
  publisher={Nature Publishing Group}
}

@inproceedings{oh2023ecgqa,
  author        = {Oh, Jungwoo and Lee, Gyubok and Bae, Seongsu and Kwon, Joon-myoung and Choi, Edward},
  title         = {{ECG-QA: A Comprehensive Question Answering Dataset Combined With Electrocardiogram}},
  booktitle     = {Advances in Neural Information Processing Systems 36 (NeurIPS 2023) Datasets and Benchmarks Track},
  year          = {2023},
  doi           = {10.48550/arXiv.2306.15681},
  url           = {https://arxiv.org/abs/2306.15681},
  note          = {arXiv preprint arXiv:2306.15681},
  archivePrefix = {arXiv},
  eprint        = {2306.15681},
  primaryClass  = {q-bio.QM}
}

@ARTICLE{Lawson2023-jd,
  title     = "The global primary care crisis",
  author    = "Lawson, Euan",
  journal   = "Br. J. Gen. Pract.",
  publisher = "Royal College of General Practitioners",
  volume    =  73,
  number    =  726,
  pages     =  3,
  month     =  jan,
  year      =  2023,
  language  = "en"
}

@ARTICLE{Russo2023-vq,
  title     = "The layered crisis of the primary care medical workforce in the
               European region: what evidence do we need to identify causes and
               solutions?",
  author    = "Russo, Giuliano and Perelman, Julian and Zapata, Tomas and
               Šantrić-Milićević, Milena",
  journal   = "Hum. Resour. Health",
  publisher = "Springer Science and Business Media LLC",
  volume    =  21,
  number    =  1,
  pages     =  55,
  month     =  jul,
  year      =  2023,
  language  = "en"
}

@ARTICLE{Lee2024-gg,
  title     = "A systematic review exploring the factors that contribute to
               increased primary care physician turnover in socio-economically
               deprived areas",
  author    = "Lee, Jasmine and Kontopantelis, Evangelos",
  journal   = "PLoS One",
  publisher = "Public Library of Science (PLoS)",
  volume    =  19,
  number    =  12,
  pages     = "e0315433",
  month     =  dec,
  year      =  2024,
  language  = "en"
}

@ARTICLE{Liu2024-ud,
  title    = "Primary care practice telehealth use and low-value care services",
  author   = "Liu, Terrence and Zhu, Ziwei and Thompson, Michael P and
              McCullough, Jeffrey S and Hou, Hechuan and Chang, Chiang-Hua and
              Fendrick, A Mark and Ellimoottil, Chad",
  journal  = "JAMA Netw. Open",
  volume   =  7,
  number   =  11,
  pages    = "e2445436",
  month    =  nov,
  year     =  2024,
  language = "en"
}

@ARTICLE{Palepu2025-de,
  title         = "Towards conversational {AI} for disease management",
  author        = "Palepu, Anil and Liévin, Valentin and Weng, Wei-Hung and
                   Saab, Khaled and Stutz, David and Cheng, Yong and Kulkarni,
                   Kavita and Mahdavi, S Sara and Barral, Joëlle and Webster,
                   Dale R and Chou, Katherine and Hassidim, Avinatan and Matias,
                   Yossi and Manyika, James and Tanno, Ryutaro and Natarajan,
                   Vivek and Rodman, Adam and Tu, Tao and Karthikesalingam, Alan
                   and Schaekermann, Mike",
  journal       = "arXiv [cs.CL]",
  month         =  mar,
  year          =  2025,
  archivePrefix = "arXiv",
  primaryClass  = "cs.CL"
}

@ARTICLE{Campanozzi2023-xl,
  title    = "The role of digital literacy in achieving health equity in the
              third millennium society: A literature review",
  author   = "Campanozzi, Laura Leondina and Gibelli, Filippo and Bailo, Paolo
              and Nittari, Giulio and Sirignano, Ascanio and Ricci, Giovanna",
  journal  = "Front. Public Health",
  volume   =  11,
  pages    =  1109323,
  month    =  feb,
  year     =  2023,
  language = "en"
}

@misc{phillips2020chexphoto10000photostransformations,
      title={CheXphoto: 10,000+ Photos and Transformations of Chest X-rays for Benchmarking Deep Learning Robustness}, 
      author={Nick A. Phillips and Pranav Rajpurkar and Mark Sabini and Rayan Krishnan and Sharon Zhou and Anuj Pareek and Nguyet Minh Phu and Chris Wang and Mudit Jain and Nguyen Duong Du and Steven QH Truong and Andrew Y. Ng and Matthew P. Lungren},
      year={2020},
      eprint={2007.06199},
      archivePrefix={arXiv},
      primaryClass={eess.IV},
      url={https://arxiv.org/abs/2007.06199}, 
}

@article{johri2025evaluation,
  title={An evaluation framework for clinical use of large language models in patient interaction tasks},
  author={Johri, Shreya and Jeong, Jaehwan and Tran, Benjamin A and Schlessinger, Daniel I and Wongvibulsin, Shannon and Barnes, Leandra A and Zhou, Hong-Yu and Cai, Zhuo Ran and Van Allen, Eliezer M and Kim, David and others},
  journal={Nature Medicine},
  pages={1--10},
  year={2025},
  publisher={Nature Publishing Group US New York}
}

@article{zhou2024pre,
  title={Pre-trained multimodal large language model enhances dermatological diagnosis using SkinGPT-4},
  author={Zhou, Juexiao and He, Xiaonan and Sun, Liyuan and Xu, Jiannan and Chen, Xiuying and Chu, Yuetan and Zhou, Longxi and Liao, Xingyu and Zhang, Bin and Afvari, Shawn and others},
  journal={Nature Communications},
  volume={15},
  number={1},
  pages={5649},
  year={2024},
  publisher={Nature Publishing Group UK London}
}

@misc{schmidgall2024agentclinic,
      title={AgentClinic: a multimodal agent benchmark to evaluate AI in simulated clinical environments}, 
      author={Samuel Schmidgall and Rojin Ziaei and Carl Harris and Eduardo Reis and Jeffrey Jopling and Michael Moor},
      year={2024},
      eprint={2405.07960},
      archivePrefix={arXiv},
      primaryClass={cs.HC},
      url={https://arxiv.org/abs/2405.07960}, 
}

@article{gruppen2017clinical,
  title={Clinical reasoning: defining it, teaching it, assessing it, studying it},
  author={Gruppen, Larry D},
  journal={Western Journal of Emergency Medicine},
  volume={18},
  number={1},
  pages={4},
  year={2017},
  publisher={California Chapter of the American Academy of Emergency Medicine (Cal/AAEM)}
}

@article{wong2018real,
  title={Real-world use of telemedicine-a picture is worth a thousand words},
  author={Wong, Richard and Dunn, Ken},
  journal={PMFA J},
  volume={5},
  number={2},
  pages={331--342},
  year={2018}
}

@article{newble2004techniques,
  title={Techniques for measuring clinical competence: objective structured clinical examinations},
  author={Newble, David},
  journal={Medical education},
  volume={38},
  number={2},
  pages={199--203},
  year={2004},
  publisher={Wiley Online Library}
}

@article{liu2020deep,
  title={A deep learning system for differential diagnosis of skin diseases},
  author={Liu, Yuan and Jain, Ayush and Eng, Clara and Way, David H and Lee, Kang and Bui, Peggy and Kanada, Kimberly and de Oliveira Marinho, Guilherme and Gallegos, Jessica and Gabriele, Sara and others},
  journal={Nature medicine},
  volume={26},
  number={6},
  pages={900--908},
  year={2020},
  publisher={Nature Publishing Group US New York}
}

@article{shivashankara2024ecg,
  title={ECG-Image-Kit: a synthetic image generation toolbox to facilitate deep learning-based electrocardiogram digitization},
  author={Shivashankara, Kshama Kodthalu and Shervedani, Afagh Mehri and Clifford, Gari D and Reyna, Matthew A and Sameni, Reza and others},
  journal={Physiological measurement},
  volume={45},
  number={5},
  pages={055019},
  year={2024},
  publisher={IOP Publishing}
}

@article{john2022smartphone,
  title={Smartphone technology for communications between clinicians--A scoping review},
  author={John, Bernadette and McCreary, Christine and Roberts, Anthony},
  journal={Journal of Dentistry},
  volume={122},
  pages={104112},
  year={2022},
  publisher={Elsevier}
}

@article{giordano2017whatsapp,
  title={WhatsApp messenger as an adjunctive tool for telemedicine: an overview},
  author={Giordano, Vincenzo and Koch, Hilton and Godoy-Santos, Alexandre and Belangero, William Dias and Pires, Robinson Esteves Santos and Labronici, Pedro and others},
  journal={Interactive journal of medical research},
  volume={6},
  number={2},
  pages={e6214},
  year={2017},
  publisher={JMIR Publications Inc., Toronto, Canada}
}

@article{INTERNIST_1982,
  author    = {Miller, Randolph A and Pople Jr, Harry E and Myers, Jack D},
  title     = {{INTERNIST-1}, an experimental computer-based diagnostic consultant for general internal medicine},
  journal   = {New England Journal of Medicine},
  year      =  1982,
  volume    =  307,
  number    =  8,
  pages     = "468--476",
  publisher = {Mass Medical Soc},
  doi       = {10.1056/NEJM198208193070803}
}

@article{donaghy2019acceptability,
  title={Acceptability, benefits, and challenges of video consulting: a qualitative study in primary care},
  author={Donaghy, Eddie and Atherton, Helen and Hammersley, Victoria and McNeilly, Hannah and Bikker, Annemieke and Robbins, Lucy and Campbell, John and McKinstry, Brian},
  journal={British journal of general practice},
  volume={69},
  number={686},
  pages={e586--e594},
  year={2019},
  publisher={British Journal of General Practice}
}

@article{ritunga2024challenges,
  title={Challenges and recommendations in the implementation of audiovisual telemedicine communication: a systematic review},
  author={Ritunga, Imelda and Claramita, Mora and Widaty, Sandra and Soebono, Hardyanto},
  journal={Korean journal of medical education},
  volume={36},
  number={3},
  pages={315},
  year={2024}
}

@misc{LLaVaMed_2023,
      title={LLaVA-Med: Training a Large Language-and-Vision Assistant for Biomedicine in One Day}, 
      author={Chunyuan Li and Cliff Wong and Sheng Zhang and Naoto Usuyama and Haotian Liu and Jianwei Yang and Tristan Naumann and Hoifung Poon and Jianfeng Gao},
      year={2023},
      eprint={2306.00890},
      archivePrefix={arXiv},
      primaryClass={cs.CV},
      url={https://arxiv.org/abs/2306.00890}, 
}

@preprint{MedFlamingo_2023,
    title = {{Med-Flamingo: a Multimodal Medical Few-shot Learner}},
    author = {Moor, Michael and Huang, Qian and Wu, Shirley and Yasunaga, Michihiro and Zakka, Cyril and Dalmia, Yash and Reis, Eduardo Pontes and Rajpurkar, Pranav and Leskovec, Jure},
    year = {2023},
    eprint = {2307.15189},
    archivePrefix = {arXiv},
    primaryClass = {cs.CL},
    url = {https://arxiv.org/abs/2307.15189}
}

@misc{SudanWhatsApp_2021,
      title={{Utility of WhatsApp in healthcare provision and sharing of medical information with caregivers of children with neurodisabilties: experience from Sudan}}, 
      author={IN Mohamed and MA Elseed},
      year={2021},
      url={https://doi.org/10.24911/SJP.106-1596913564}
}

@misc{TamilNaduWhatsApp_2024,
	author = {Sharma, Kamal and Manickaraj, Saranya and Saini, Navsangeet},
	title = {{F}rom {T}exts to {T}riage: {T}he {W}hats{A}pp {C}linic {E}xperience - {B}{M}{J} {G}lobal {H}ealth blog --- blogs.bmj.com},
	howpublished = {\url{https://blogs.bmj.com/bmjgh/2025/01/14/from-texts-to-triage-the-whatsapp-clinic-experience/}},
	year = {2025},
}

@misc{UKWhatsAppPolicy_FT_2022,
  author = {Hughes, Laura},
  title = {{Hospitals should check staff use of WhatsApp to ensure patient safety, UK watchdog says}},
  year = {2024},
  howpublished = {Financial Times},
  note = {URL: \url{https://www.ft.com/content/f428e4ff-4dd9-4613-8485-dfc77fa154ec}}
}

@article{UKWhatsAppPolicyRef_2024,
  author = {Dobbs, Tom},
  title = {'What's so wrong with Whatsapp?},
  journal = {The Bulletin of the Royal College of Surgeons of England},
  volume = {106},
  number = {3},
  pages = {166-167},
  year = {2024},
  doi = {10.1308/rcsbull.2024.62},
  url={https://doi.org/10.1308/rcsbull.2024.62}
}

@article{Thakkar2016Mobile,
  title={{Mobile telephone text messaging for medication adherence in chronic disease: a meta-analysis}},
  author={Thakkar, Jitesh and Kurup, Rahul and Laba, Tracey-Lea and Santo, Karla and Thiagalingam, Aravinda and Rodgers, Anthony and Woodward, Mark and Redfern, Julie and Chow, Clara K},
  journal={JAMA Internal Medicine},
  volume={176},
  number={3},
  pages={340--349},
  year={2016},
  publisher={American Medical Association},
  doi={10.1001/jamainternmed.2015.7667}
}

@article{Buddhika2019Review,
author = {Buddhika Senanayake and Sumudu I Wickramasinghe and Mark D Chatfield and Julie Hansen and Sisira Edirippulige and Anthony C Smith},
title ={Effectiveness of text messaging interventions for the management of depression: A systematic review and meta-analysis},
journal = {Journal of Telemedicine and Telecare},
volume = {25},
number = {9},
pages = {513-523},
year = {2019},
doi = {10.1177/1357633X19875852},
url = {https://doi.org/10.1177/1357633X19875852},
}

@Article{PisaniSMSPrevention,
author="Pisani, Anthony R
and Wyman, Peter A
and Cero, Ian
and Kelberman, Caroline
and Gurditta, Kunali
and Judd, Emily
and Schmeelk-Cone, Karen
and Mohr, David
and Goldston, David
and Ertefaie, Ashkan",
title="Text Messaging to Extend School-Based Suicide Prevention: Pilot Randomized Controlled Trial",
journal="JMIR Ment Health",
year="2024",
month="Dec",
day="6",
volume="11",
issn="2368-7959",
doi="10.2196/56407",
url="https://mental.jmir.org/2024/1/e56407",
url="https://doi.org/10.2196/56407"
}

@article{PACHECO2020106221,
title = {PAD-UFES-20: A skin lesion dataset composed of patient data and clinical images collected from smartphones},
journal = {Data in Brief},
volume = {32},
pages = {106221},
year = {2020},
issn = {2352-3409},
doi = {https://doi.org/10.1016/j.dib.2020.106221},
url = {https://www.sciencedirect.com/science/article/pii/S235234092031115X},
author = {Andre G.C. Pacheco and Gustavo R. Lima and Amanda S. Salomão and Breno Krohling and Igor P. Biral and Gabriel G. {de Angelo} and Fábio C.R. {Alves Jr} and José G.M. Esgario and Alana C. Simora and Pedro B.C. Castro and Felipe B. Rodrigues and Patricia H.L. Frasson and Renato A. Krohling and Helder Knidel and Maria C.S. Santos and Rachel B. {do Espírito Santo} and Telma L.S.G. Macedo and Tania R.P. Canuto and Luíz F.S. {de Barros}}
}

@misc{freyberg2024mintwrappermakemultimodal,
      title={MINT: A wrapper to make multi-modal and multi-image AI models interactive}, 
      author={Jan Freyberg and Abhijit Guha Roy and Terry Spitz and Beverly Freeman and Mike Schaekermann and Patricia Strachan and Eva Schnider and Renee Wong and Dale R Webster and Alan Karthikesalingam and Yun Liu and Krishnamurthy Dvijotham and Umesh Telang},
      year={2024},
      eprint={2401.12032},
      archivePrefix={arXiv},
      primaryClass={cs.HC},
      url={https://arxiv.org/abs/2401.12032}, 
}

@article{li2022using,
  title={Using the apple watch to record multiple-lead electrocardiograms in detecting myocardial infarction: where are we now?},
  author={Li, Ke and Elgalad, Abdelmotagaly and Cardoso, Cristiano and Perin, Emerson C},
  journal={Texas Heart Institute Journal},
  volume={49},
  number={4},
  pages={e227845},
  year={2022},
  publisher={Texas Heart{\textregistered} Institute, Houston}
}

@misc{googleGemini20FlashGenerative,
	author = {},
	title = {{G}emini 2.0 {F}lash  |  {G}enerative {A}{I} on {V}ertex {A}{I}  |  {G}oogle {C}loud --- cloud.google.com},
	howpublished = {\url{https://cloud.google.com/vertex-ai/generative-ai/docs/models/gemini/2-0-flash}},
	year = {},
	note = {[Accessed 30-04-2025]},
}

@article{ashley2000prevalence,
  title={The prevalence and prognostic significance of electrocardiographic abnormalities},
  author={Ashley, Euan Angus and Raxwal, Vinod Kumar and Froelicher, Victor F},
  journal={Current problems in Cardiology},
  volume={25},
  number={1},
  pages={1--72},
  year={2000},
  publisher={Elsevier}
}
